\RequirePackage{rotating}
\documentclass[acmsmall]{acmart}

\usepackage[utf8]{inputenc} 
\usepackage[T1]{fontenc}    
\usepackage{hyperref}       
\usepackage{url}            
\usepackage{booktabs}       
\usepackage{amsfonts}       
\usepackage{nicefrac}       
\usepackage{microtype}      
\usepackage{amsmath}
\usepackage{xfrac}
\usepackage[ruled,vlined]{algorithm2e}
\usepackage{amsthm}
\theoremstyle{plain}

\theoremstyle{plain}
\newtheorem{definition}{Definition}
\theoremstyle{plain}

\usepackage{bbm}
\usepackage{wrapfig}
\usepackage{xcolor}
\usepackage{enumitem}
\usepackage{subfigure}
\usepackage{array}
\usepackage{multirow}
\usepackage{hhline}
\usepackage{longtable} 
\usepackage{lscape} 
\usepackage{adjustbox}

\usepackage{makecell}
\usepackage{tabularx}
\usepackage{rotating}

\newcommand{\specialcell}[2][c]{%
  \begin{tabular}[#1]{@{}c@{}}#2\end{tabular}}

\usepackage{pdflscape}
\usepackage{fancyhdr}
\usepackage[flushleft]{threeparttable}
\usepackage[nolist,printonlyused]{acronym}

\usepackage{verbatim}

\begin{acronym}
	\acro{AI}{Artificial Intelligence}
	\acro{ALU}{arithmetic logic unit}
	\acro{CNN}{Convolutional Neural Network}
	\acro{DL}{Deep Learning}
	\acro{DNN}{Deep Neural Network}
	\acro{DRAM}{dynamic random-access memory}
	\acro{DSP}{Digital Signal Processor}
	\acro{FFT}{Fast Fourier Transform}
	\acro{FLOP}{floating point operations}
	\acro{FLOPS}{floating point operations per second}
	\acro{FoM}{Figure of Merit}
	\acro{GPU}{Graphics Processing Unit}
	\acro{HPC}{High Performance Computing}
	\acro{IO}{input/output}
	\acro{IPS}{instructions per second}
	\acro{MAC}{multiplier–accumulate}
	\acro{ML}{Machine Learning}
	\acro{NLR}{No Local Reuse}
	\acro{OS}{Output Stationary}
	\acro{PE}{processing engine}
	\acro{QoS}{Quality of Service}
	\acro{RS}{Row Stationary}
	\acro{SIMD}{Single instruction, multiple data}
	\acro{SIMT}{Single instruction, multiple threads}
	\acro{SoC}{System-on-Chip}
	\acro{WS}{Weight Stationary}
\end{acronym}

\title{On-Device Machine Learning: An Algorithms and Learning Theory Perspective}

\author{Sauptik Dhar}
\affiliation{America Research Center, LG Electronics}
\email{sauptik.dhar@lge.com}

\author{Junyao Guo}
\affiliation{America Research Center, LG Electronics}
\email{junyao.guo@lge.com}

\author{Jiayi (Jason) Liu}
\affiliation{America Research Center, LG Electronics}
\email{jason.liu@lge.com}

\author{Samarth Tripathi}
\affiliation{America Research Center, LG Electronics}
\email{samarth.tripathi@lge.com}

\author{Unmesh Kurup}
\affiliation{America Research Center, LG Electronics}
\email{unmesh@ukurup.com}

\author{Mohak Shah}
\affiliation{America Research Center, LG Electronics}
\email{mohak@mohakshah.com}

\begin{document}

\begin{abstract}
    The predominant paradigm for using machine learning models on a device is to train a model in the cloud and perform inference using the trained model on the device. However, with increasing number of smart devices and improved hardware, there is interest in performing model training on the device. Given this surge in interest, a comprehensive survey of the field from a device-agnostic perspective sets the stage for both understanding the state-of-the-art and for identifying open challenges and future avenues of research. However, on-device learning is an expansive field with connections to a large number of related topics in AI and machine learning (including online learning, model adaptation, one/few-shot learning, etc.). Hence, covering such a large number of topics in a single survey is impractical. This survey finds a middle ground by reformulating the problem of on-device learning as resource constrained learning where the resources are compute and memory. This reformulation allows tools, techniques, and algorithms from a wide variety of research areas to be compared equitably. In addition to summarizing the state-of-the-art, the survey also identifies a number of challenges and next steps for both the algorithmic and theoretical aspects of on-device learning.
\end{abstract}

\maketitle

\tableofcontents

\section{Introduction}
\label{sec:intro}






The addition of intelligence to a device carries the promise of a seamless experience that is tailored to each user's specific needs while maintaining the integrity of their personal data. The current approach to making such intelligent devices is based on a cloud paradigm where data is collected at the device level and transferred to the cloud. Once transferred, this data is then aggregated with data collected from other devices, processed, and used to train a machine learning model. When the training is done, the resulting model is pushed from the cloud back to the device where it is used to improve the device's intelligent behavior. In the cloud paradigm, all machine learning that happens on the device is inference, that is, the execution of a model that was trained in the cloud. This separation of roles – data collection and inference on the edge, data processing and model training in the cloud – is natural given that end-user devices have form-factor and cost considerations that impose limits on the amount of computing power and memory they support, as well as the energy that they consume. 

Cloud-based systems have access to nearly limitless resources and are constrained only by cost considerations making them ideal for resource intensive tasks like data storage, data processing, and model building. However, the cloud-based paradigm also has drawbacks that will become more pronounced as AI becomes an ubiquitous aspect of consumer life. The primary considerations are in the privacy and security of user data as this data needs to be transmitted to the cloud and stored there, most often, indefinitely. Transmission of user data is open to interference and capture, and stored data leaves open the possibility of unauthorized access. 

In addition to privacy and security concerns, the expectation for intelligent devices will be that their behavior is tailored specifically to each consumer. However, cloud-trained models are typically less personalized as they are built from data aggregated from many consumers, and each model is built to target broad user segments because building individual models for every consumer and every device is cost prohibitive in most cases. This de-personalization also applies to distributed paradigms like  federated learning that typically tend to improve a global model based on averaging the individual models \cite{McMahanMRA16}. 


Finally, AI-enabled devices will also be expected to learn and respond instantly to new scenarios but cloud-based training is slow because added time is needed to transmit data and models back and forth from the device. Currently, most use cases do not require real-time model updates, and long delays between data collection and model updates is not a serious drawback. But, as intelligent behavior becomes commonplace and expected, there will be a need for real-time updates, like in the case of connected vehicles and autonomous driving. In such situations, long latency becomes untenable and there is a need for solutions where model updates happen locally and not in the cloud. 

As devices become more powerful, it becomes possible to address the drawbacks of the cloud model by moving some or all of the model development onto the device itself. Model training, especially in the age of deep learning, is often the most time-consuming part of the model development process, making it the obvious area of focus to speed up model development on the device. Doing model training on the device is often referred to variously as \textit{Learning on the Edge} and \textit{On-device Learning}. However, we distinguish between these terms, with  \textit{learning on the edge} used as a broad concept to signify the capability of real or quasi-real time learning without uploading data to the cloud while \textit{on-device learning} refers specifically to the concept of doing model training on the resource-constrained device itself. 

\subsection{On-device Learning}
\subsubsection{Definition of an Edge Device}
Before we elaborate on on-device learning, it is helpful to define what we mean by a device, or specifically an edge device, in the context of on-device learning. We define an edge device to be a device whose compute, memory, and energy resources are constrained and cannot be easily increased or decreased. These constraints may be due to form-factor considerations (it is not possible to add more compute or memory or battery without increasing the size of the device) or due to cost considerations (there is enough space to add a GPU to a washing machine but this would increase its cost prohibitively). This definition of an edge device applies to all such consumer and industrial devices where resource constraints place limitations on what is available for building and training AI models. Any cloud solution such as Amazon AWS, Google Cloud Platform, Microsoft Azure, or even on-premise computing clusters do not fit the edge definition because it is easy to provision additional resources as needed. Likewise, a workstation would not be considered an edge device because it is straightforward to replace its CPU, add more memory, and even add an additional GPU card. A standard laptop on the other hand would be considered an edge device as it is not easy to add additional resources as needed, even though their resources generally far exceed what is normally considered as available in a consumer edge device. 

\begin{figure*}[h] 
\centering
\includegraphics[height=5cm]{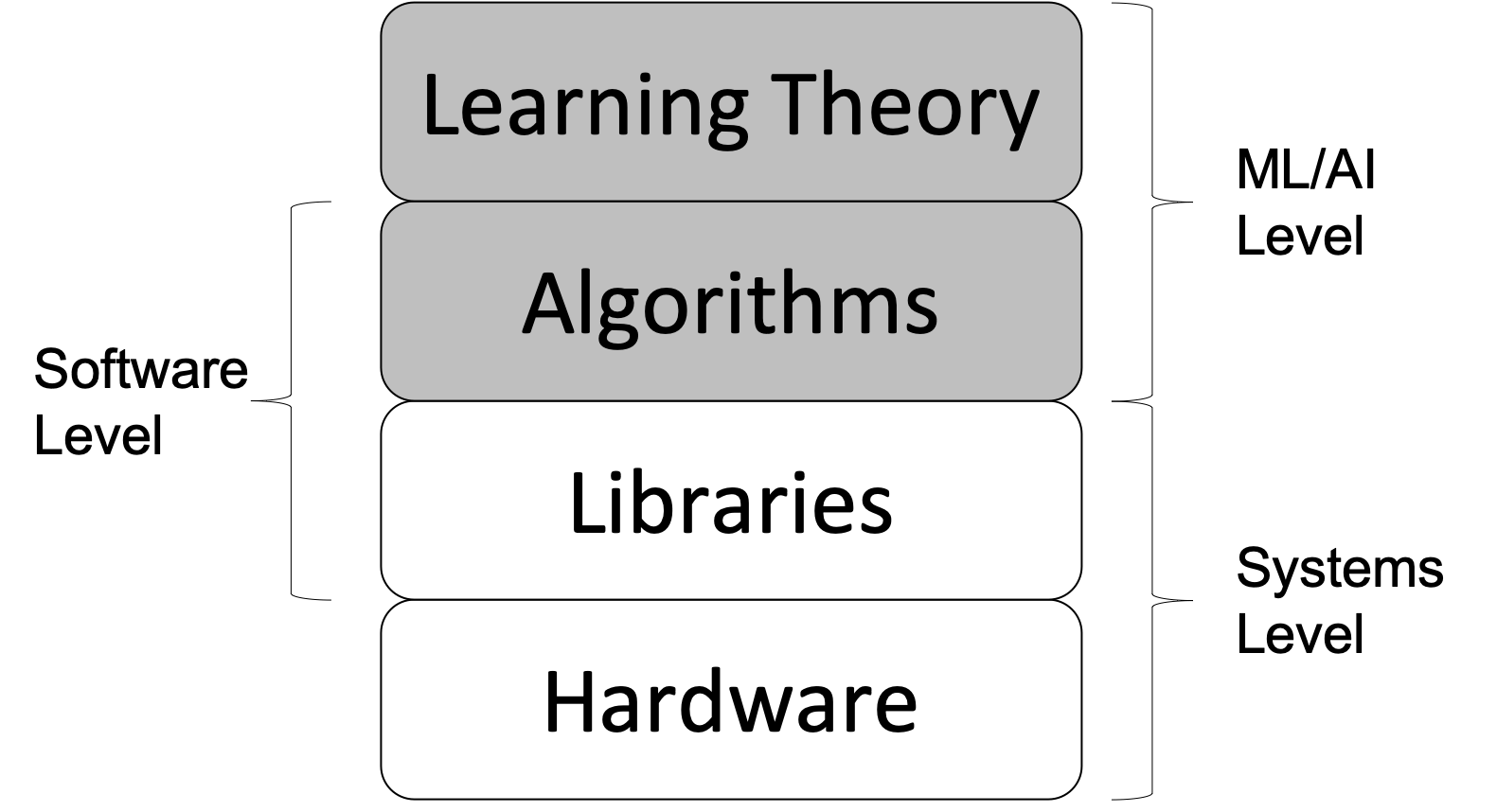}
\caption{Levels of constraints on edge devices. Only the topics in gray (Learning Theory and Algorithms) fall under the scope of this survey.}
\label{fig::research_levels}
\end{figure*}

\subsubsection{Training Models on an Edge Device}
The primary constraints to training models on-device in a reasonable time-frame is the lack of compute and memory on the device. Speeding up training is possible either by adding more resources to the device or using these resources more effectively or some combination of the two. Fig \ref{fig::research_levels} shows a high-level breakdown of the different levels at which these approaches can be applied. Each level in this hierarchy abstracts implementation details of the level below it and presents an independent interface to the level above it. 

\begin{enumerate}
\item \textbf{Hardware:} At the bottom of the hierarchy are the actual chipsets that execute all learning algorithms. Fundamental research in this area aims at improving existing chip design (by developing chips with more compute and memory, and lower power consumption and footprint) or developing new designs with novel architectures that speed up model training.  While hardware research is a fruitful avenue for improving on-device learning, it is an expensive process that requires large capital expenditure to build laboratories and fabrication facilities, and usually involves long timescales for development.

\item \textbf{Libraries:}  Every machine learning algorithm depends on a few key operations (such as Multiply-Add in the case of neural networks). The libraries that support these operations are the interface that separate the hardware from the learning algorithms. This separation allows for algorithm development that is not based on any specific hardware architecture. Improved libraries can support faster execution of algorithms and speed up on-device training. However, these libraries are heavily tuned to the unique aspects of the hardware on which the operations are executed. This dependency limits the amount of improvement that can be gained by new libraries.

\item \textbf{Algorithms:} Since on-device learning techniques are grounded in their algorithmic implementations, research in novel algorithm development is an important part of making model training more efficient. Such algorithm development can take into account resource constraints as part of the model training process. Algorithm development leads to hardware-independent techniques but the actual performance of each algorithm is specific to the exact domain, environment, and hardware, and needs to be verified empirically for each configuration. Depending on the number of choices available in each of these dimensions, the verification space could become very large.

\item \textbf{Theory:} Every learning algorithm is based on an underlying theory that guarantees certain aspects of its performance. Developing novel theories targeted at on-device learning help us understand how algorithms will perform under resource-constrained settings. However, while theoretical research is flexible enough to apply across classes of algorithms and hardware systems, it is limited due to the inherent difficulty of such research and the need to implement a theory in the form of an algorithm before its utility can be realized.
\end{enumerate}

\begin{figure*}[h] 
\centering
\includegraphics[height=11cm]{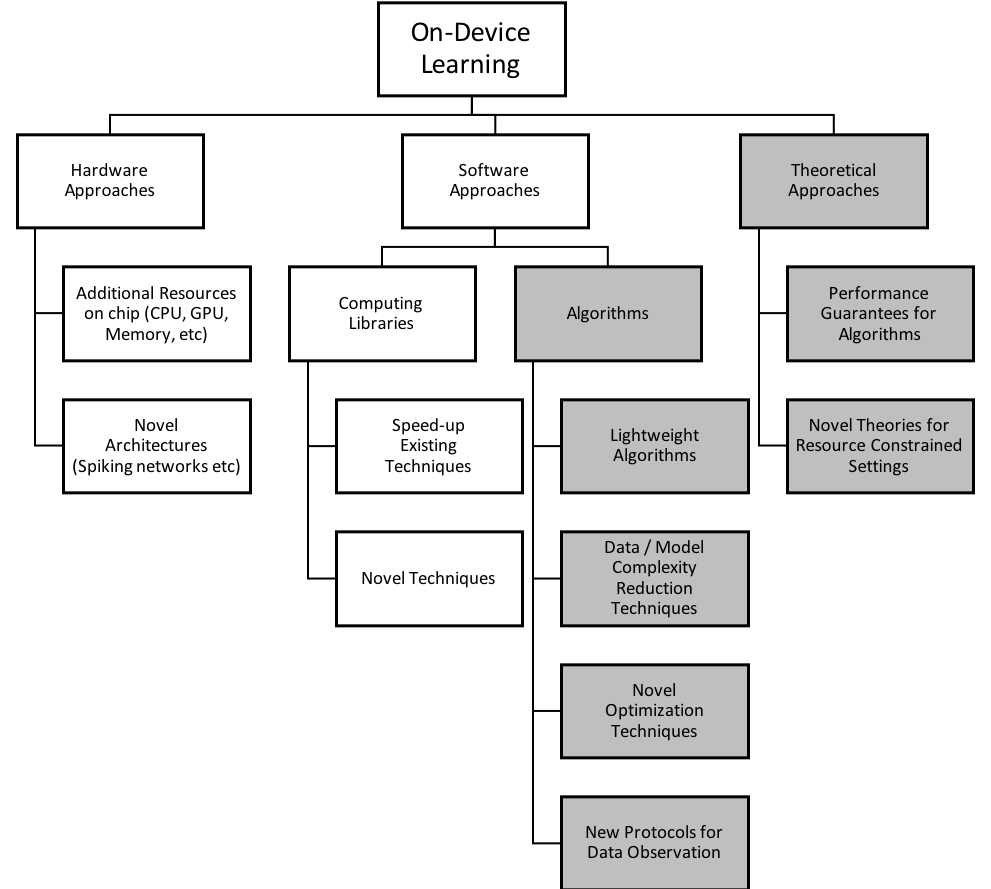}
\caption{Different approaches to improving on-device learning. Topics in gray are covered in this survey.}
\label{fig::ondevice_learning}
\end{figure*}

Fig \ref{fig::ondevice_learning} shows an expanded hierarchical view of the different levels of the edge learning stack and highlight different ways to improve the performance of model training on the device at each level. The hardware approaches involve either adding additional resources to the restricted form-factor of the device or developing novel architectures that are more resource efficient. Software approaches to improve model training involve either improving the performance of computing libraries such as OpenBLAS, Cuda, CuDNN or improving the performance of the machine learning algorithms themselves. Finally, theoretical approaches help direct new research on ML algorithms and improve our understanding of existing techniques and their generalizability to new problems, environments, and hardware.

\subsection{Scope of this survey}
There is a large ongoing research effort, mainly in academia, that looks at on-device learning from multiple points of view including single vs. multiple edge devices, hardware vs. software vs. theory, and domain of application such as healthcare vs. consumer devices vs. autonomous cars. 
Given the significant amount of research in each of these areas it is important to restrict this survey to a manageable subset that targets the most important aspects of on-device learning. 

We first limit this survey to the \textit{Algorithms} and \textit{Learning Theory} levels in Fig \ref{fig::research_levels}. This allows us to focus on the machine learning aspects of on-device learning and develop new techniques that are independent of specific hardware. 
We also limit the scope of this survey to learning on a single device. This restriction makes the scope of the survey manageable while also providing a foundation on which to expanded to distributed settings. In addition, at the right level of abstraction, a distributed edge system can be considered as a single device with an additional resource focused on communication latency. 
This view allows us to extend single-device algorithms and theories to the distributed framework at a later stage. 

The goal of this survey then is to provide a large-scale view of the current state-of-the-art in algorithmic and theoretical advances for on-device learning on single devices. To accomplish this goal, the survey reformulates the problem of on-device learning as one of resource constrained learning. This reformulation describes the efficiency of on-device learning using two resources – compute and memory – and provides a foundation for a fair comparison of different machine learning and AI techniques and their suitability for on-device learning. Finally, this survey identifies challenges in algorithms and theoretical considerations for on-device learning and provides the background needed to develop a road map for future research and development in this field.

\subsection{How to read this survey}
This survey is a comprehensive look at the current state-of-the-art in training models on resource-constrained devices. It is divided into 4 main sections excluding the introduction and the conclusion. Section 2 briefly introduces the resources and their relevance to on-device learning. Sections 3 and 4 respectively focus on the algorithmic and theoretical levels of the edge platform hierarchy. Finally section 5 provides a brief summary and identifies various challenges in making progress towards a robust framework for on-device learning. For those interested in specific aspects of on-device learning, sections 3 and 4 are mostly self-contained and can be read separately.

\textbf{Resource Constraints in On-Device Learning (in Section \ref{sec:resource_constraints}) :} briefly discusses the relevant resources that differentiate on-device learning from a cloud-based system. Most existing research at the various levels in Fig \ref{fig::research_levels} is targeted towards addressing on-device learning when there is limited availability of these resources.

\textbf{Algorithm Research (in Section \ref{sec:algo}):} addresses recent algorithmic developments towards accurately capturing the hardware constraints in a software framework and then surveys the state-of-the-art in machine learning algorithms that take into account resource constraints. This section categorizes the algorithms from a computational perspective (i.e. the underlying computational model used).

\textbf{Theory Research (in Section \ref{sec:theory}) :} addresses on-device learning from a statistical perspective and surveys traditional learning theories forming the basis of most of the algorithm designs addressed in section \ref{sec:algo}. It later addresses the `un-learnability' problem in a resource constrained setting and surveys newer resource constrained learning theories. Such newer theories abstract the resource constraints (i.e. memory, processing speed etc.) as an information bottleneck and provides performance guarantees for learning under these settings.   

Finally, section \ref{sec:challenges} summarises the previous sections and addresses some of the open challenges in on-device learning research.


\section{Resource Constraints in On-Device Learning} 
\label{sec:resource_constraints}



The main difference between traditional machine learning and learning/inference on edge are the additional constraints imposed by device resources. Designing AI capabilities that run on a device necessitates building machine learning algorithms with high model accuracy while concurrently maintaining the resource constraints enforced by the device. This section discusses these critical resource constraints that pose major challenges while designing learning algorithms for edge devices. 

\subsection{Processing Speed}\label{sec:processing_speed}

The response time is often among the  most critical factors for the usability of any on-device application \cite{nielsen1993usability}. The two commonly used measurements are \textit{throughput} and \textit{latency}. Throughput is measured as the rate at which the input data is processed. To maximize throughput it is common to group inputs into batches resulting in higher utilization. But, measuring this incurs additional wait time for aggregating data into batches. Hence, for time-critical use cases, latency is the more frequently used measure. Latency characterizes the time interval between a single input and its response. Although throughput is the inverse of latency (when batch size is fixed to 1), the runtime of an application may vary dramatically depending on whether computations are optimized for throughput vs. latency \cite{venieris2018deploying}. To simplify our discussion, in this survey we use an abstract notion of \textit{runtime} as a proxy for both \textit{throughput} and \textit{latency}.

For the physical system, the processing speed dictates the runtime of an application. This speed is typically measured in clock frequency (i.e. the number of cycles per second) of a processor. Within each cycle, a processor carries out a limited number of operations based on the hardware architecture and the types of the operations. For scientific computations, \acs{FLOPS} is frequently used to measure the number of floating point operations per second. Another frequently used measure specifically for matrix intensive computations such as those in machine learning algorithms is \ac{MAC}. Thus, besides increasing the clock frequency, efficiently combining multiple operations into a single cycle is also an important topic for improving the processing speed. 

On a separate note, the processing speed of a system is also sensitive to the communication latency among the components inside a system. As discussed before, the communication latency aspect is better aligned to the distributed/decentralized computing paradigm for edge learning and has not been addressed in this survey.

\subsection{Memory}\label{sec:memory}

At the heart of any machine learning algorithm is the availability of data for building the model. The second important resource for building AI driven on-device applications is memory. The memory of a computing device provides immediate data access as compared to other storage components. However, this speed of data access comes at higher costs. For reasons of cost, most edge devices are designed with limited memory. As such, the memory footprint of edge applications are typically tailored for a specific target device. 

Advanced machine learning algorithms often take a significant amount of memory during model building through storage of the model parameters and auxiliary variables etc. For instance, even relatively simple image classification models like ResNet-50 can take megabytes of memory space (see \autoref{table:bigCNNcompare}).  Therefore, designing a lightweight model is of a key aspect of accomplishing machine learning on the device. A detailed survey of such techniques is covered in sections \ref{sec:algo} and \ref{sec:theory}.

Besides the model size, querying the model parameters for processing is both time-consuming and energy-intensive \cite{sze2017efficient}.  For example, a single \ac{MAC} operation requires three memory reads and one memory write. In the worst case, these reads and write may be on the off-chip memory rather than the on-chip buffer. This would result in a significant throughput bottleneck and cause orders of magnitude higher energy consumption \cite{chen2017eyeriss}. 


\subsection{Power Consumption and Energy Efficiency}\label{sec:sys_power}

Power consumption is another crucial factor for on-device learning.  An energy efficient solution can prolong the battery lifetime and cut maintenance costs. The system {\it power}, commonly studied in hardware development, is the ratio of energy consumption and time span for a given task. However, it is not a suitable measure for machine learning applications on the edge. First, the power consumption depends on the volume of computation required, e.g. data throughput. Second, the application is often capped at the maximum power of the device when the learning task is intensive. Therefore to better quantify the power consumption, the total energy consumption along with the throughput is recommended for comparing energy efficiency.

Linking energy consumption of a particular AI driven application for a specific device jointly depends on a number of factors like runtime, memory etc. Capturing these dependencies are almost never deterministic. Hence, most existing research estimate the power/energy usage through a surrogate function which typically depends on the memory and runtime of an application. A more detailed survey of such advanced approaches are covered in section \ref{sec:algo}.

\begin{table}[htbp]
\setlength\tabcolsep{2.5pt}
\caption{Comparison of hardware requirements}
    \label{tab:sys}
\begin{threeparttable}
\scriptsize
\begin{tabular}{l|c|c|c|c|c}
\toprule
Use & Device & Hardware Chip & Computing & Memory & Power \\
\noalign{\hrule height 2pt}
Workstation        & NVIDIA DGX-2\tnote{1}            & 16~NVIDIA Tesla V100 \acsp{GPU}         & 2~P\acs{FLOPS}                                      & 512~GB & 10 kW              \\
Mobile Phone       & Pixel 3\tnote{2}                 & Qualcomm Snapdragon™ 845\tnote{3}  & 727~G\acs{FLOPS}  & 4~GB         & 34mW (Snapdragon)  \\
Autonomous Driving & NVDIA DRIVE AGX Xavier\tnote{4}  & NVIDIA Xavier processor            & 30~TOPS                                            & 16~GB        & 30W                \\
Smart home         & Amazon Echo\tnote{5}             & TI DM3725 ARM Cortex-A8            & up to 1~GHz                                        & 256~MB  & 4~W (peak) \\
General IoT        & Qualcomm AI Engine \tnote{6}              & Hexagon 685 and Adreno 615         & 2.1~TOPS                                           & 2-4~GB       & 1W                 \\
General IoT                & Raspberry Pi 3\tnote{7}          & Broadcom ARM Cortex A53            & 1.2~GHz                                            &  1~GB        &  0.58~W \\
General IoT        & Arduino Uno Rev3\tnote{8}        & ATmega328P     &   16~MHz  & 2~KB  & 0.3~W \\
\bottomrule
\end{tabular}
\begin{tablenotes}
 \item[1] \url{http://images.nvidia.com/content/pdf/dgx-2-print-datasheet-738070-nvidia-a4-web.pdf}
 \item[2] \url{https://store.google.com/product/pixel_3_specs}
 \item[3] \url{https://www.qualcomm.com/products/snapdragon-845-mobile-platform}
 \item[4] \url{https://www.nvidia.com/en-us/self-driving-cars/drive-platform/hardware/}
 \item[5] \url{https://www.ifixit.com/Teardown/Amazon+Echo+Teardown/33953}
 \item[6] \url{https://www.qualcomm.com/products/vision-intelligence-400-platform}
 \item[7] \url{https://www.raspberrypi.org/magpi/raspberry-pi-3-specs-benchmarks/}
 \item[8] \url{https://store.arduino.cc/usa/arduino-uno-rev3}
\end{tablenotes}
\normalsize
\end{threeparttable}
\end{table}

\subsection{Typical Use Case and Hardware}\label{sec:hardwarespec}
Finally we conclude this section by providing a brief landscape of the variety of edge devices used in several use-case domains and their resource characterization. As seen in \autoref{tab:sys}, learning on edge spans a wide spectrum of hardware specifications. Hence, designing machine learning models for edge devices requires a very good understanding of the resource constraints, and appropriately incorporating these constraints into the systems, algorithms and theoretical design levels.

\section{Algorithms for On-Device Learning}
\label{sec:algo}
The algorithms approach targets developing resource-efficient techniques that work with existing resource-constrained platforms. This section provides a detailed survey on the various approaches to analyze and estimate a Machine Learning (ML) model's resource footprint, and the state-of-the-art algorithms proposed for on-device learning.

We present these algorithms and model optimization approaches in a task-agnostic manner. This section  discusses general approaches that adapt both traditional ML algorithms and deep learning models to a resource constrained setting. These approaches can be applied to multiple tasks such as classification, detection, regression, image segmentation and super-resolution. 

Many traditional ML algorithms, such as SVM and Random Forest, are already suitable for multiple tasks and do not need special consideration per task. Deep learning approaches, on the other hand, do vary considerably from task to task. However, for the tasks mentioned above, these networks generally deploy a CNN or RNN as the backbone network for feature extraction \cite{jiao2019survey, minaee2020image}. As a consequence the resource footprint of training the backbone networks would directly affect the training performance of the overall model. For example, deep learning based image segmentation methods usually use ResNet as the backbone, while adding additional modules such as a graphical models, modifications to the convolution computation, or combining backbone CNNs with other architectures such as encoder-decoders to achieve their goal.  \cite{minaee2020image}. 

Given these commonalities, we expect the approaches presented in this section to be generalizable to different tasks and refer the readers to surveys \cite{jiao2019survey, minaee2020image, fernandez2019extensive, wang2020deep} for detailed comparison of task-specific models. More importantly, we categorize the directions in which improvements can be made, which could serve as a guideline for proposing novel resource-efficient techniques for new models/tasks. Note that for CNN benchmarking, we use the image classification task as an example as benchmarking datasets are well established for this area. Tasks such as scene segmentation and super-resolution currently lack standard benchmarks making it difficult to equitably compare models.

\subsection{Resource Footprint Characterization}
\label{subsec:resourcefootprint}
Before adapting ML algorithms to the resource-constrained setting, it is important to first understand the resource requirements of common algorithms and identify their resource bottlenecks. The conventional approach adopts an asymptotic analysis (such as the Big-O notation) of the algorithm's implementation. For DNNs, an alternate analysis technique is to use hardware-agnostic metrics such as the number of parameters and number of operations (FLOPs/MACs). However, it has been demonstrated that hardware-agnostic metrics are too crude to predict the real performance of algorithms \cite{yang2017designing, lu2017modeling, marculescu2018hardware} because these metrics depend heavily on the specific platform and framework used. This hardware dependency has lead to a number of efforts aimed at measuring the true resource requirements of many algorithms (mostly DNNs) on specific platforms. A third approach to profiling proposes building regression models that accurately estimate and predict resource consumption of DNNs from their weights and operations. We will provide an overview of all these resource characterization approaches in this subsection as well as new performance metrics for algorithm analysis that incorporates the algorithm's resource footprint.

\subsubsection{Asymptotic Analysis}
\label{subsec:asymptotic}
In this section, we present a comparative overview of the computational and space complexities of both traditional machine learning algorithms and DNNs using asymptotic analysis and hardware-agnostic metrics. 

\textbf{Traditional Machine Learning Algorithms:}
Due to the heterogeneity of model architectures and optimization techniques, there is no unified approach that characterizes the resource utilization performance of traditional machine learning algorithms. The most commonly used method is asymptotic analysis that quantifies computational complexity and space complexity using the Big-O notation. Table \ref{table:MLcompare} summarizes the computational and space complexities of 10 popular machine learning algorithms based on their implementation in Map-Reduce \cite{chu2007map} and Scikit-Learn \cite{treescikit}. Note that algorithm complexity can vary across implementations, but we believe the results demonstrated in Table \ref{table:MLcompare} are representative of the current landscape. For methods that require iterative optimization algorithms or training steps, the training complexity is estimated for one iteration. 

\begin{table}[htbp]
\caption{Comparison of traditional machine learning algorithms. Notation: $m$-number of training samples; $n$-input dimension; $c$-number of classes.}
\centering
\scriptsize
\begin{tabular}{ m{2.1cm}<{\centering}|m{1.7cm}<{\centering}| m{3cm}<{\centering}|m{2.8cm}<{\centering}|m{2.6cm}<{\centering}}
\toprule
Algorithm & Model size & Optimization & Training complexity & Inference complexity\\
\noalign{
\hrule height 2pt
}
Decision tree &$\mathcal{O}(m)$ &- &$\mathcal{O}(mnlog(m))$ & $\mathcal{O}(log(m))$\\
\hline
Random forest & $\mathcal{O}(N_{tree}m)$ &- &$\mathcal{O}(N_{tree}mnlog(m))$ & $\mathcal{O}(N_{tree}log(m))$\\
\hline
SVM & $\mathcal{O}(n)$ & gradient descent &$\mathcal{O}(m^2n)$ &$\mathcal{O}(m_{sv}n)$ \\
\hline
Logistic regression & $\mathcal{O}(n)$ &Newton-Raphson &$\mathcal{O}(mn^2+n^3)$ & $\mathcal{O}(n)$ \\
\hline
kNN &$\mathcal{O}(mn)$&- & - &$\mathcal{O}(mn)$ \\
\hline
Naive Bayes &$\mathcal{O}(nc)$ &-&$\mathcal{O}(mn+nc)$ & $\mathcal{O}(nc)$ \\
\hline
Linear regression &$\mathcal{O}(n)$ &matrix inversion &$\mathcal{O}(mn^2+n^3)$ &$\mathcal{O}(n)$ \\
\noalign{
\hrule height 2pt
}
k-Means &- &- &$\mathcal{O}(mnc)$ &- \\
\hline
EM &-&-&$\mathcal{O}(mn^2+n^3)$ & -\\
\noalign{
\hrule height 2pt
}
PCA &- &eigen-decomposition &$\mathcal{O}(mn^2+n^3)$ & -\\
\bottomrule
\end{tabular}
\normalsize
\label{table:MLcompare}
\end{table}

Some key observations can be made from Table \ref{table:MLcompare}. First, most traditional machine learning algorithms (except tree-based methods and kNN) have a model size that is linear to the input dimension, which do not require much memory (compared to DNNs which will be discussed later). Second, except for kNN which requires distance calculation between the test data and all training samples, the inference step of other algorithms is generally very fast. Third, some methods require complex matrix operations such as matrix inversion and eigen-decomposition with computational complexity around $\mathcal{O}(n^3)$ \cite{chu2007map}. Therefore, one should consider whether these matrix operations can be efficiently supported by the targeted platform when deploying these methods on resource-constrained devices. 

In terms of accuracy of traditional machine learning algorithms, empirical studies have been carried out using multiple datasets \cite{fernandez2014we, alam2016analysis, zhang2017up}. However, for on-device learning, there are few studies that analyze both accuracy and complexity of ML algorithms and the tradeoffs therein. 

\textbf{Deep Neural Networks:}
Deep neural networks have shown superior performance in many computer vision and natural language processing tasks. To harvest their benefits for edge learning, one has to first evaluate whether the models and the number of operations required can fit into the available resources on a given edge device. Accuracy alone cannot justify whether a DNN is suitable for on-device deployment. To this end, recent studies that propose new DNN architectures also consider MACs/FLOPs and number of weights to provide a qualitative estimate of the model size and computational complexity. However, both memory usage and energy consumption also depend on the feature map or activations \cite{chen2017eyeriss,lu2017modeling}. Therefore, in the following, we review popular DNNs in terms of accuracy, weights, activations and MACs. Accuracies of models are excerpted from best accuracies reported on open platforms and the papers that proposed these models. Number of weights, activations and MACs are either excerpted from papers that first proposed the model or calculated using the Netscope tool \cite{netscope}. Note that these papers only count the MACs involved in a forward pass. Since the majority of the computational load in training DNNs happens in the backward pass, the values reported do not give a realistic picture of the computational complexity for model training.

\begin{table}[htbp]
\caption{Comparison of popular CNNs.}
\centering
\scriptsize
\begin{tabular}{ m{2.4cm}<{\centering}|m{1.4cm}<{\centering}| m{1.4cm}<{\centering}| m{1.4cm}<{\centering}|m{1.4cm}<{\centering}|m{1.4cm}<{\centering}|m{1.4cm}<{\centering}}
\toprule
Metric & \makecell{ AlexNet \\ \cite{krizhevsky2012imagenet} } & \makecell{ VGG-16 \\ \cite{simonyan2014very}} & \makecell{ GoogLeNet \\ \cite{szegedy2015going} }& \makecell{ ResNet-18 \\ \cite{he2016deep}} & \makecell{ ResNet-50 \\ \cite{he2016deep}} & \makecell{ Inception\\ v3 \cite{szegedy2016rethinking} }\\
\noalign{
\hrule height 2pt
}
Top-1 acc. &57.2 &71.5 &69.8 &69.6 &76.0 &76.9 \\
\hline
Top-5 acc. &80.2 &91.3 &90.0 &89.2 &93.0 &93.7 \\
\hline
Input size &227$\times$227 &224$\times$224 & 224$\times$224&224$\times$224 &224$\times$224&299$\times$299 \\
\noalign{
\hrule height 2pt
}
$\#$ of stacked CONV layers &5 &13 & 21&17 &49 &16 \\
\hline
Weights &2.3M &14.7M &6.0M & 9.5M&23.6M &22M \\
\hline
Activations &0.94M &15.23M &6.8M &3.2M &11.5M &10.6M \\
\hline
MACs &666M &15.3G &1.43G &1.8G &3.9G & 3.8G\\
\noalign{
\hrule height 2pt
}
$\#$ of FC layers &3 & 3&1 &1 & 1&1 \\
\hline
Weights &58.7M &125M &1M &0.5M &2M & 2M\\
\hline
Activations &9K &9K &2K &1.5K &3K &3K \\
\hline
MACs &58.7M &125M&1M &0.5M &2M &2M \\
\noalign{
\hrule height 2pt
}
Total weights &61M &138M & 7M& 10M&25.6M &24M \\
\hline
Total activations &0.95M &15.24M &6.8M &3.2M &11.5M &10.6M \\
\hline
Total MACs &724M &15.5G &1.43G &1.8G &3.9G &3.8G \\
\bottomrule
\end{tabular}
\normalsize
\label{table:bigCNNcompare}
\end{table}

Compared to other network architectures, most on-device learning research focuses on CNNs, which can be structured as an acyclic graph and analyzed layer-by-layer. Specifically, a CNN consists of two types of layers, namely, the convolutional layer (CONV) and the fully connected layer (FC). It has been shown in \cite{chen2017eyeriss} that the resource requirement of these two layers can be very different due to their different data flow and data reuse patterns. In Table \ref{table:bigCNNcompare}, we summarize the layer-wise statistics of popular high-performance CNN models submitted to the ImageNet challenge. In general, these models are very computation and memory intensive, especially during training. Training requires allocating memory to weights, activations, gradients, data batches, and workspace, which is at least hundreds of MBs if not GBs. These models are hard to deploy on a resource-constrained device, let alone be trained on one. To enable CNN deployment on edge devices, models with smaller sizes and more compact architectures are proposed, which we will review in Section \ref{subsec:reducemodelcomplexity}.


\subsubsection{Resource Profiling} \label{subsec:res_prof}
The most accurate way to quantify the resource requirements of machine learning algorithms is to measure them during deployment. Table \ref{table:DNNbenchmark} summarizes current efforts on DNN benchmarking using various platforms and frameworks. For inference, we only present benchmarks that use at least one edge device such as a mobile phone or an embedded computing system (such as NVidia Jetson TX1). However, as it has not been feasible to train DNNs at large-scale on edge devices yet, we present training benchmarks utilizing single or multiple high-performance CPU or GPUs. Interestingly, apart from measuring model-level performance, three benchmarks \cite{adolf2016fathom, gao2018bigdatabench, deepbench} further decompose the operations involved in running DNNs and profile micro-architectural performance. We believe that these finer-grained measurements can provide more insights into resource requirement estimation for both training and inference, which are composed of these basic operations. 

\begin{sidewaystable}[htbp]
\scriptsize
\caption{DNN profiling benchmarks.}
\label{table:DNNbenchmark}
\centering
\begin{tabularx}{1\linewidth}{X X X X X X}
\toprule
Benchmark & Platform & Framework & Model & Metric & Highlight \\
\noalign{
\hrule height 2pt
}
AI Android\cite{ignatov2018ai}&57 SoC Processors;~200 mobile phones & Tensorflow Lite& CNNs for 9 image processing tests &inference: runtime&the most comprehensive benchmark on CNN deployment on Android system\\
\hline
NAS\cite{cheng2018searching}&Intel i5-7600, NVidia Jetson TX1, Xiaomi Redmi Note 4&N/A&8 light-weight CNNs from neural architecture search methods&inference: time and memory&survey of models generated from NAS methods on resource-limited devices\\
\hline
Fathom\cite{adolf2016fathom}&Skylake i7-6700k CPU with 32GB RAM; NVidia GeForce GtX 960&TensorFlow &3CNNs, 2RNNs, 1DRL, 1Autoencoder, 1Memory Network & training: single-/multi-thread execution time by op type (e.g., MatMul, Conv2D etc., 22 in total) & micro-architectural analysis; exploration of parallelism\\
\hline
BigDataBench\cite{gao2018bigdatabench}&Intel ®Xeon E5-2620 V3 CPU with 64GB RAM & Hadoop, Spark, JStorm, MPI, Impala, Hive, TensorFlow, Caffe (each supporting different models) & Micro benchmarks: 21 ops, e.g., sort, conv, etc.; Comp benchmarks:23, e.g., pagerank, kmeans, etc. & training: runtime, bandwidth, memory (L1-L3 cache and DRAM)&operating system level benchmarking; large variety of models\\
\hline
DAWNBench\cite{coleman2017dawnbench}&Nvidia K80 GPU&TensorFlow, PyTorch&ResNets&training: time-to-accuracy vs. optimization techniques, inference runtime vs. training runtime&defining a new metric by end-to-end training time to a certain validation accuracy\\
\hline
DeepBench\cite{deepbench}&Training:Intel Xeon Phi 7250, 7 types of NVidia GPUs; Inference: 3 NVidia GPUs, iPhone6 \& 7, Raspberry Pi3&N/A & Operations: Matmul, Conv, Recurrent Ops, All-reduce&training and inference runtime, FLOPs&profiling basic operations with respect to input dimensions\\
\hline
TBD\cite{zhu2018tbd}&16 machines each with a Xeon 28-core CPU and 1-4 NVidia Quadro P4000 / NVidia TITAN Xp GPUs&TensorFlow, MXNet, CNTK& 3CNNs, 3RNNs, 1GAN, 1DRL & training: throughput, GPU compute utilization, FP32 utilization, CPU utilization, memory&profiling of same model across frameworks; multi-GPU multi-machine training\\
\hline
SyNERGY\cite{rodriguesfine}& NVidia Jetson TX1 &Caffe&11 CNNs&inference: energy consumption (both per layer and network),SIMD instructions, bus access &proposing a multi-variable regression model to predict energy consumption\\
\hline
DNNAnalysis\cite{canziani2016analysis}& NVidia Jetson TX1&Torch7&14 CNNs&inference: runtime, power, memory, accuracy vs. throughput, parameter utilization& comparison of CNN inference on an embedded computing system\\
\hline
DNNBenchmarks\cite{bianco2018benchmark} & NVidia Jetson TX1; NVidia TITAN Xp GPU &PyTorch&44 CNNs &inference:accuracy, memory, FLOPs, runtime & very comprehensive comparison of CNN inference on both embedded computing system and powerful workstation\\
\bottomrule
\end{tabularx}

\end{sidewaystable}
\normalsize

As opposed to DNNs, there are few profiling results reported for on-device deployment of traditional machine learning algorithms, and usually as a result of comparing these algorithms to newly developed ones. For example, inference time and energy are profiled in \cite{kumar2017resource} for a newly proposed tree-based Bonsai method, local deep kernel learning, single hidden layer neural network, and gradient boosted decision tree on the Arduino Uno micro-controller board. However, memory footprint is not profiled. Some other works empirically analyze the complexity and performance of machine learning algorithms \cite{yao2017complexity, lim2000comparison} where the experiments are conducted on computers but not resource constrained devices. To better understand the resource requirements of traditional machine learning methods, more systematic experiments need to be designed to profile their performance on different platforms and various frameworks. 

\subsubsection{Resource Modeling and Estimation}
\label{subsec:resourcemodeling}
\begin{table}[htbp]
\caption{DNN resource requirements modeling. ASIC: Application-Specific Integrated Circuit. Matmul: matrix multiplication. RMSPE: root mean square percentage error.}
\scriptsize
\begin{tabular}{ m{1.8cm}|m{1.3cm}| m{1.3cm}| m{1.3cm}|m{2.5cm}|m{1.3cm}|m{1.5cm}}
\toprule
Work & Platform & Framework & Metric & \makecell{Measured \\ features} & \makecell{ Regression \\ model} & \makecell{ Relative \\ error} \\
\noalign{
\hrule height 2pt
}
Augur\cite{lu2017modeling} & NVidia TK1, TX1 &Caffe &inference: memory, time &matrix dimensions in matmul, weights, activations&linear & memory: 28\% - 50\%; time: 6\% - 20\%\\
\hline
Paleo\cite{qi2016paleo} & NVidia Titan X GPU cluster & TensorFlow&training \& inference: time&forward \& backward FLOPs, weights, activations, data, platform percent of peak&linear& 4\%-30\% \\
\hline
\makecell{ Gianniti et al. \\ \cite{giannitiperformance}} &NVidia Quadro M6000 GPU & - &training: time & forward \& backward FLOPs of all types of layers & linear & < 23\% \\
\hline
\makecell{ SyNERGY \\ \cite{rodriguesfine}} & Nvidia Jetson TX1 & Caffe & inference: energy & MACs & linear & < 17\% (w/o MobileNet)\\
\hline
\makecell{ NeuralPower \\\cite{cai2017neuralpower} }&Nvidia Titan X \& GTX 1070 &TensorFlow \& Caffe &inference: time, power, energy&layer configuration hyper-parameters, memory access, FLOPs, activations, batch size&polynomial&time: < 24\%; power: < 20\%; energy: < 5\%\\
\hline
\makecell{ HyperPower \\ \cite{stamoulis2018hyperpower}} &Nvidia GTX1070 \& Tegra TX1&Caffe&inference: power, memory&layer configuration hyper-parameters&linear & RMSPE < 7\%\\
\hline
\makecell{ Yang et al. \\ \cite{yang2017designing}} & ASIC Eyeriss\cite{chen2017eyeriss} & - & inference: energy & MACs, memory access&- & -\\
\hline
\makecell{ DeLight \\ \cite{rouhani2016delight}} & Nvidia Tegra TK1& Theano  & training\& inference: energy & layer configuration hyper-parameters & linear & -\\
\bottomrule
\end{tabular}
\normalsize
\label{table:DNNresourcemodeling}
\end{table}

To provide more insights into how efficiently machine learning algorithms can be run on a given edge device, it's helpful to model and predict the resource requirements of these algorithms. Even though performance profiling can provide accurate evaluation of resource requirements, it can be costly as it requires the deployment of all models to be profiled. If such requirements can be modeled before deployment, it will then be more helpful for algorithm design and selection. Recently, there have been an increasing number of studies that attempt to estimate energy, power, memory and runtime of DNNs based on measures such as matrix dimension and MACs. There are many benefits if one can predict resource requirements without deployment and runtime profiling. First, it can provide an estimate of training and deployment cost of a model before actual deployment, which can reduce unnecessary implementation costs. Second, the modeled resource requirements can be integrated into the algorithm design process, which helps to efficiently create models that are tailored to specific user-defined resource constraints. Third, knowing the resource consumption of different operation types can help to decide when offloading and performance optimization are needed to successfully run learning models on edge devices \cite{lu2017modeling}.

In Table \ref{table:DNNresourcemodeling}, we summarize recent studies that propose approaches to model resource requirements of DNNs. Generally, a linear regression model is built upon common features such as FLOPs for certain types of operations, activation size, matrix size, kernel size, layer configuration hyper-parameters, etc., which are not hard to aquire. The Relative Error column shows how these estimation models perform compared to the actual runtime profiling results of a network. We can see that most models demonstrate a relative error between 20\% and 30\%, which indicates that even though hardware dependency and various measures are taken into account, it can still be challenging to make consistently accurate predictions on resource requirements. This large relative error further shows that metrics such as FLOPs or MACs alone are very crude estimates that can provide limited insights. Note that all models are platform and framework dependent, meaning that the coefficients of the regression model will change when the platform or framework changes. Nevertheless, these studies provide a feasible approach to approximate resource requirements of DNNs. It remains to be seen whether similar methodology can be adopted to model the resource requirements of machine learning algorithms other than DNNs.

\subsubsection{New Metrics}
\label{sec:sys_model_performance_metric}
Resource constrained machine learning models cannot be evaluated solely on traditional performance metrics like, accuracy for classification, perplexity in language models, or mean average precision for object detection. In addition to these traditional metrics there is a need for system specific performance metrics. Such metrics can be broadly categorized as below:



\begin{enumerate}

\item Analyze the multi-objective pareto front of the designed model along the dimension of accuracy and resource footprint. Following \cite{cheng2018searching} some commonly used metrics include,
\begin{equation} \label{eq::reward_pareto1}
\text{Reward} =
\begin{cases}
	\text{Accuracy},& \text{if Power}\leq\text{Power budget}\\
	0, & \text{otherwise,}
\end{cases}
\end{equation}
or 
\begin{equation} \label{eq::reward_pareto2}
\text{Reward} =
\begin{cases}
	1-\text{Energy}^*,& \text{if Accuracy}\geq\text{Accuracy threshold}\\
	0, & \text{otherwise,}
\end{cases}
\end{equation}
where $\text{Energy}^*$ is a normalized energy consumption.
Using this measure, a model can be guaranteed to fulfill the resource constraints. However, it is hard to compare several models as there is no single metric. In addition, it is also harder to optimize the model when reward goes to zero.

\item Scalarization of the multi-objective metric into a unified reward \cite{he2018amc}, \ac{FoM} \cite{velasco2018performance}, or NetScore \cite{wong2018netscore} that merges all performance measures into one number, e.g. 
\begin{equation} \label{eq::scalar_pareto1}
\text{Reward} = - \text{Error}\times\log(\text{FLOPs}),
\end{equation}
or
\begin{equation} \label{eq::scalar_pareto2}
\text{FoM}=\frac{\text{Accuracy}}{\text{Runtime}\times\text{Power}},
\end{equation}
or
\begin{equation} \label{eq::scalar_netscore}
\text{NetScore} = 20 \log \bigg(\frac{\text{Accuracy}^\alpha}{\text{Parameters}^\beta \text{MACs}^\gamma}\bigg),
\end{equation}
where $\alpha$, $\beta$ and $\gamma$ are coefficients that control the influence of individual metrics. 
Using a single value, a relative ordering between several models is possible. However, now there are no separate threshold on the resource requirements. Hence, the optimized model may still not fit the hardware limitations. 
\end{enumerate}

Note that, in addition to using the metrics in eq. \eqref{eq::reward_pareto1} to \eqref{eq::scalar_netscore} to evaluate the model performance, the works in 
\cite{cheng2018searching, he2018amc,velasco2018performance} also utilize these metrics to build and optimize their learning algorithms. The main idea is to accurately maintain the system requirements and also improve the model's performance during training \cite{tan2018mnasnet}.  

\subsection{Resource Efficient Training}
\label{sec:efficienttraining}
In this section, we review existing algorithm improvements on resource-constrained training. Note that orthogonal approaches have been made in specialized hardware architecture design and dataflow/memory optimization to enable on-device learning \cite{sze2017efficient, rhu2016vdnn}. While algorithmic and architectural approaches are tightly correlated and in most cases complement each other, we maintain our focus on resource-efficient algorithms that can work across platforms.

In the machine learning task, the available resources are mainly consumed by three sources: 1) the ML model itself; 2) the optimization process that learns the model; and 3) the dataset used for learning. For instance, a complex DNN has a large memory footprint while a kernel based method may require high computational power. The dataset requires storage when it is collected and consumes memory when it is used for training. Correspondingly, there are mainly three approaches that adapt existing ML algorithms to a resource-constrained setting: 1) reducing model complexity by incorporating resource constraints into the hypothesis class; 2) modifying the optimization routine to reduce resource requirements during training; and 3) developing new data representations that utilize less storage. Particularly, for data representation, there is a line of work that proposes new learning protocols for resource-constrained scenarios where full data observability is not possible \cite{ben1998learning, shamir2014fundamental}. In the following, we broadly review resource efficient training methods according to these categories. 

\subsubsection{Lightweight ML Algorithms} \label{sec:algo:low_footprint}
As shown in Table \ref{table:MLcompare} in Section \ref{subsec:asymptotic}, some traditional ML algorithms, such as Naive Bayes, Support Vector Machines (SVMs), Linear Regression, and certain Decision Tree algorithms like C4.5, have relatively low resource footprints \cite{alam2016analysis}. The lightweight nature of these algorithms make them good candidates for on-device learning applications. These algorithms can be readily implemented on a particular embedded sensor system or portable device with few refinements to fit the specific hardware architecture \cite{lin2010tuning, lee2016integrating, haigh2015machine}. 

\subsubsection{Reducing Model Complexity}
\label{subsec:reducemodelcomplexity}
Adding constraints to the hypothesis class or discovering structures in the hypothesis class is a natural approach to selecting or generating simpler hypotheses. This approach is mostly applied to large models, such as decision trees, ensemble-based models, and DNNs. 

Decision trees can have large memory footprint due to the large number of tree nodes. To avoid over-fitting and to reduce this memory footprint, pruning is a typical approach applied for deploying decision trees \cite{kulkarni2012pruning}. Recently, there are also studies that develop shallow and sparse tree learners with powerful nodes that only require a few kilobytes of memory \cite{kumar2017resource}.

For ensemble-based algorithms such as boosting and random forest, a critical question is how to select weak learners to achieve higher accuracy with lower computational cost. To address this issue, \cite{grubb2012speedboost} proposed a greedy approach that selects weak decision-tree-based learners that can generate a prediction at any time, where the accuracy can increase if larger latency is allowed to process weaker learners. 

For DNNs, there have been a plethora of studies that propose more lightweight model architectures for on-device learning. While most of these models are designed for edge deployment, we still include these approaches in efficient training because reductions in weights and connections in these new architectures lead to fewer resource requirements for continued training. In Table \ref{table:smallCNNcompare}, we provide a comparative overview of some representative lightweight CNN architectures. Compared to networks presented in Table \ref{table:bigCNNcompare}, these networks are much smaller in terms of size and computation, while still retaining fairly good accuracy compared to large CNNs. Unfortunately, there is not much work developing lightweight models for DNN architectures other than CNNs.

Among the models presented in Table \ref{table:smallCNNcompare}, MnasNet is a representative model generated via automatic neural architecture search (NAS), whereas all other networks are designed manually. NAS is usually based on reinforcement learning (RL) or evolutionary algorithms where a predefined search space is explored to generate the optimal network architecture that achieves the best tradeoff between accuracy and energy/runtime \cite{wistuba2019survey}. To speed up NAS, the resource modeling techniques mentioned in Section \ref{subsec:resourcemodeling} could be helpful, as they can replace the computationally intensive process of measuring the actual model performance on device. For example, in ChamNet\cite{dai2019chamnet}, predictors for model accuracy, latency, and energy consumption are proposed that can guide the search and reduce time to find desired models from many GPU hours to mere minutes. 

Note that even though for most NAS algorithms, the goal is to optimize for inference performance, the resulting architectures usually contain fewer parameters and require fewer FLOPS, which could lead to a lower resource footprint when trained on-device. In fact, a recent study \cite{radosavovic2020designing} shows that training time can be significantly reduced on the architectures found in their proposed search space. If needed, on-device resource constraints that are specific to model training can be taken into account by NAS. Keeping this in consideration, we include NAS in the resource-efficient training section, as they provide good candidates for on-device model training tasks.

\begin{table}[tbp]
\caption{Comparison of lightweight CNNs.}
\centering
\scriptsize
\begin{tabular}{ m{2.3cm}<{\centering}|m{1.2cm}<{\centering}| m{1.2cm}<{\centering}| m{1.2cm}<{\centering}|m{1.2cm}<{\centering}|m{1.2cm}<{\centering}|m{1.2cm}<{\centering}|m{1.2cm}<{\centering}}
\toprule
Metric & MobileNet V1-1.0\cite{howard2017mobilenets}& MobileNet V2-1.0\cite{sandler2018mobilenetv2} & Squeeze-Net\cite{iandola2016squeezenet} & Squeeze-Next-1.0-23\cite{gholami2018squeezenext} & ShuffleNet $1\times g = 8$\cite{zhang1707shufflenet} & Condense-Net\cite{huang2018condensenet} & \makecell{ MnasNet \\ \cite{tan2018mnasnet} }\\
\noalign{
\hrule height 2pt
}
Top-1 acc. &70.9 &71.8 &57.5 &59.0 &67.6 &71.0 &74.0  \\
\hline
Top-5 acc. &89.9 &91.0 &80.3 & 82.3&- &90.0 &91.8 \\
\hline
Input size &224$\times$224 &224$\times$224 &224$\times$224 &227$\times$227 & 224$\times$224&224$\times$224 &224$\times$224 \\
\noalign{
\hrule height 2pt
}
$\#$ of stacked CONV layers &27 & 20&26 &22 &17 &37 & 18\\
\hline
Weights &3.24M &2.17M &1.25M &0.62M &3.9M &2.8M &3.9M \\
\hline
Activations &5.2M & 1.46M&4.8M &4.7M &3.2M & 1.1M&3.9M \\
\hline
MACs &568M &299M &388M &282M &138M &274M &317M\\
\noalign{
\hrule height 2pt
}
$\#$ of FC layers & 1 &1 &0 &1 &1 &1 &1 \\
\hline
Weights &1M & 1.3M&0 & 0.1M& 1.5M& 0.1M&0.3M \\
\hline
Activations &2K &2.3K &0 &1.1K & 2.5K&1.1K &1.3K \\
\hline
MACs &1M &1.3M & 0&0.1M &1.5M &0.1M & 0.3M\\
\noalign{
\hrule height 2pt
}
Total weights & 4.24M&3.47M &1.25M & 0.72M&5.4M &2.9M &4.2M \\
\hline
Total activations &5.2M &1.46M&4.8M &4.7M &3.2M &1.1M &3.9M \\
\hline
Total MACs & 569M&300M &388M &282M &140M &274M &317M \\
\bottomrule
\end{tabular}
\normalsize
\label{table:smallCNNcompare}
\end{table}

\begin{figure*}[t] 
\centering
\includegraphics[height=8cm]{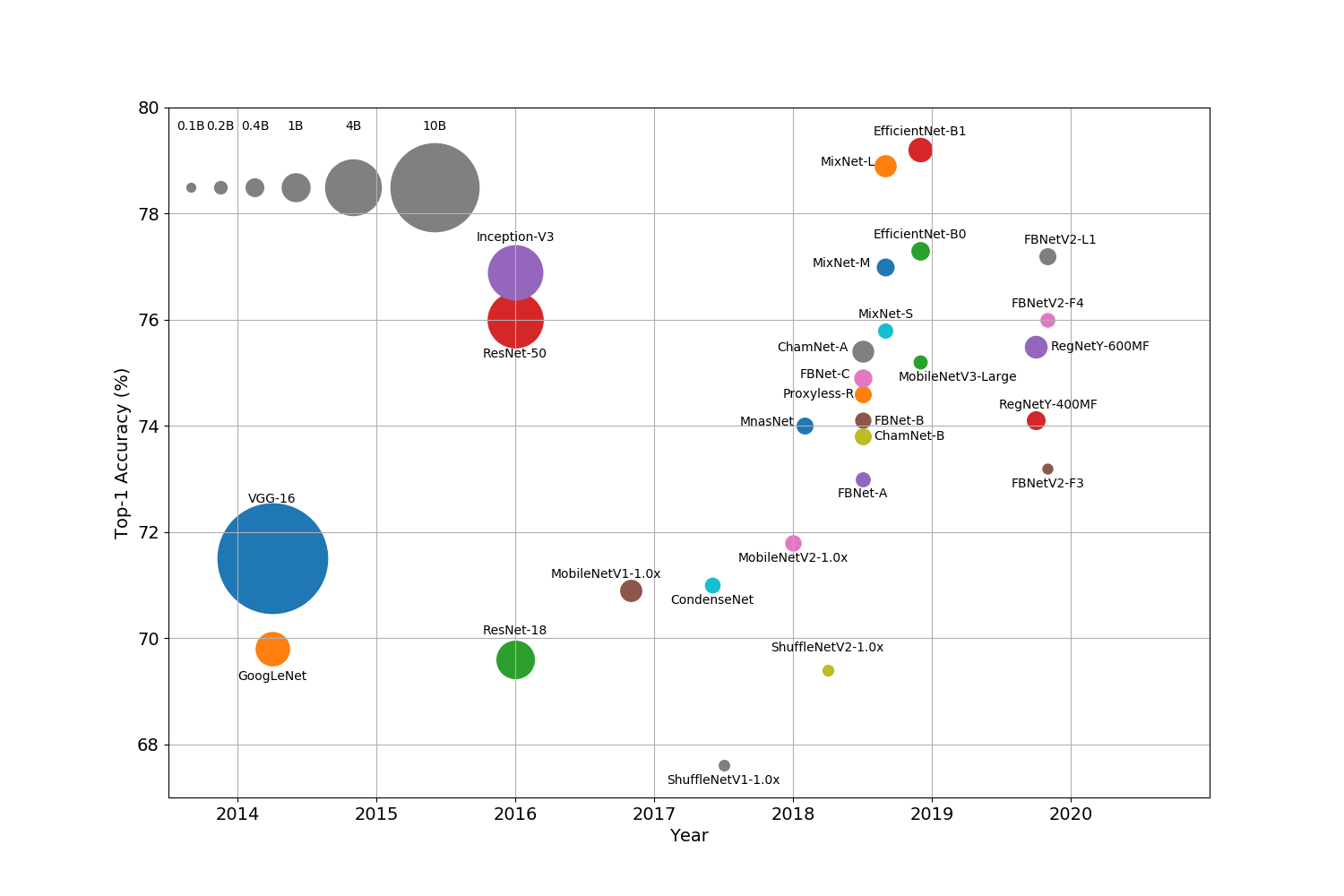}
\caption{Ball chart of the chronological evolution of model complexity. Top-1 accuracy is measured on the ImageNet dataset. The model complexity is represented by FLOPS and reflected by the ball size. The accuracy and FLOPS are taken from original publications of the models. The time of the model is when the associated publication is first made available online. }
\label{fig::models_by_flops}
\end{figure*}

Figure \ref{fig::models_by_flops} shows a chronological plot of the evolution of CNN architectures. We include the models in Table \ref{table:bigCNNcompare}, Table \ref{table:smallCNNcompare}, as well as recent NAS works including FBNet\cite{wu2019fbnet}, FBNetV2\cite{wan2020fbnetv2}, ChamNet\cite{dai2019chamnet}, ProxylessNAS\cite{cai2018proxylessnas}, SinglePath-NAS\cite{stamoulis2019single}, EfficientNet\cite{tan2019efficientnet}, MixNet\cite{tan2019mixconv}, MobileNetV3\cite{howard2019searching}, and RegNet\cite{radosavovic2020designing}. This plot shows the FLOPS each model requires, and only models that achieve more than 60\% top-1 accuracy on the ImageNet dataset are presented. 
Figure \ref{fig::models_by_flops} shows a clear transition from hand-designed CNN architectures to NAS over time, with the best performing models all being generated using NAS techniques. An advantage of NAS over hand-designed architectures is that different accuracy-resource tradeoffs can be made a part of the search space resulting in a group of models that are suitable for platforms with various resource constraints.

Apart from designing new architectures from scratch, another approach to reducing DNN model complexity is to exploit the sparse structure of the model architecture. One efficient way to learn sparse DNN structures is to add a group lasso regularization term in the loss function which encourages sparse structures in various DNN components like filters, channels, filter shapes, layer depth, etc. \cite{feng2015learning, wen2016learning, lebedev2016fast}. Particularly in \cite{wen2016learning}, when designing the regularization term, the authors take into account how computation and memory access involved in matrix multiplication are executed on hardware platforms such that the resulting DNN network can achieve practical computation acceleration.

Another approach adopted for reduced model complexity involves quantization by using low or mixed precision data representation. Quantization was first applied to inference where only the weights and/or activations are quantized post-training. This approach was then extended to training where gradients are also quantized. Conventionally, all numerical values including weights, data, and intermediate results are represented and stored using 32-bit floating point data format. There are a plethora of emerging studies exploring the use of 16-bit or lower precision for storing some or all of these numerical values in model training without much degradation in model accuracy \cite{micikevicius2017mixed, guo2018survey}. Most of these approaches involve modifications in the training routine to account for the quantization error. A more detailed discussion on the training routine modifications of these methods is provided in the next Section.


\subsubsection{Modifying Optimization Routines}  \label{sec:algo:modify_opt}

There are broadly two directions of research aimed at improving the performance of quantized models. 

\noindent \textbf{Resource constrained Model-Centric Optimization Routines}: 
Optimization or other numerical computation involved in training certain traditional ML models can be very resource intensive. For example, for kernel-based algorithms such as SVM, kernel computation usually consumes the most energy and memory. Efficient algorithms have been proposed to reduce the computation complexity and memory footprint of kernel computations, and are beneficial for both training and inference \cite{hsieh2014fast, jose2013local, le2013fastfood}.

The choice of optimization strategy is crucial for DNN training, as a proper optimization technique is critical to finding a good local minimum. While many efficient optimization algorithms such as Adam as well as network initialization techniques are exploited in DNN training, the main objective is to improve convergence and consequently reduce training time. Other resource constraints are not taken into account explicitly. In this part, we focus on efficient training algorithms that consider performance other than runtime and could potentially enable DNN training on devices with limited memory and power.

The most popular approach for reducing memory footprint of DNN training is quantization, as mentioned in section \ref{subsec:reducemodelcomplexity}. However, quantization inevitably introduces error which could make training unstable. To resolve this issue, recent works propose minor modifications to the model training process. A popular approach involves designing a mathematical model to characterize the statistical error introduced when limiting the model's precision through rounding or quantization \cite{xie1992analysis,choi1993fixed,holt1991finite, judd2015reduced}; and proposing improvements \cite{hoehfeld1991learning,kollmann1996chip,magoulas1996new}. In fact, most of the recent research in this line involves modifications in training updates either through stochastic rounding, weight initialization or through introducing the quantization error into gradient updates. The work in \cite{gupta2015deep} uses fixed point representation and stochastic rounding to account for quantization error. \citep{courbariaux2015binaryconnect} further limits the precision to binary representation of weights using stochastic binarization during forward and back propagation. However the full precision is maintained during gradient updates. The authors claim that such unbiased binarization lends to additional regularization of the network. A more detailed theoretical analysis of stochastic rounding and binary connect networks can be found in \cite{li2017training}. 
The work in \cite{lin2015neural} introduced a quantized version of back-propagation by representing the weights as powers of 2. This representation converts the multiplication operations to cheaper bit-shift operations. An alternative noisy back propagation algorithm to account for the error due to binarization of weights is also introduced in \cite{kim2016bitwise}. In a slightly different approach \cite{hubara2017quantized} uses a deterministic quantization but introduces a `straight through estimator' for gradient computation during back propagation. They further introduce a shift-based batch normalization and a new shift-based AdaMax algorithm. \cite{rastegari2016xnor} introduces a new mechanism to compute binarized convolutions and, additionally, introduces a scaling for the binarized gradients. \cite{wiedemann2020dithered} applies a non-subtractive dither quantization function to gradients and shows that this function can induce sparsity and non-zero values with low bitwidth for large enough quantization stepsize. 

A more recent approach in \cite{wang2018training} moves away from fixed-points to a new floating point representation with stochastic rounding and chunk based accumulation during training. Finally, recent work in \cite{jacob2018quantization, fan2020training} simulates the effect of quantization during inference and adds correction to the training updates by introducing quantization noise in the gradient updates. Slightly different from \cite{jacob2018quantization} where all the weights are quantized, \cite{fan2020training} quantizes a random subset of weights and back propagates unbiased gradients for the complimentary subset.  

Apart from the theoretical development of methods that reduce quantization error, some other works focus on making quantization more practical and generalizable. \cite{zhou2016dorefa} generalizes the method of binarized neural networks to allow arbitrary bit-widths for weights, activations, and gradients, and stochastically quantizes the gradients during the backward pass. To generalize the quantization technique to different types of models, \cite{cambier2020shifted} introduces two learnable statistics of DNN tensors, namely, shifted and squeezed factors, which are used to adjust the tensors' range in 8-bits for minimizing quantization loss. They show that their method works out-of-the-box for large-scale DNNs without much tuning. However, in most works, there are still some computation in training that requires full precision representation. To address this issue \cite{yang2020training} proposes a framework that quantizes the complete training path including weights, activations, gradients, error, updates, and batch-norm layers,  and converts them to 8-bit integers. Different quantization functions are used for different compute elements.

In a very different approach,  \cite{soudry2014expectation} propose expectation back propagation (EBP), an algorithm for learning the weights of a binary network using a Variational Bayes technique. The algorithm can be used to train the network such that, each weight can be restricted to be binary or ternary values. This approach is very different from the above gradient-based back-propagation algorithms. However, this approach assumes that the bias is real and is not currently applicable to CNNs. A chronological summary of the approaches modifying the training algorithms to account for quantization error is provided in Table 7. While the above approaches (summarized in Table 7) mainly target modifications to account for the limited bit representations (due to quantization), in the following section, we highlight several other approaches that are not limited  to quantization corrections. 
\begin{table}
\begin{threeparttable}[tb]
\centering
\tabcolsep=0.04cm
\caption{The chronology of the recent approaches which modifies the training algorithm to account for quantization error.} 
\label{quanttable}
\begin{scriptsize}
\begin{tabular}{c|c|c|ccc|cc}  
\noalign{
\hrule height 2pt
}
\multirow{2}{*}{Year} &\multirow{2}{*}{Approach} & \multirow{2}{*}{Keywords} & \multicolumn{3}{c}{Quantization\tnote{1}} &  \multicolumn{2}{c}{Benchmark} 
\\ \cline{4-6} \cline{7-8}
& & & Forward & Backward & \specialcell{Parameter\\ Update} & Data & Model \\ 
\noalign{
\hrule height 2pt
}
2014 &  EBP \cite{soudry2014expectation} & Expectation Back Propagation & 1 bit, FP & - & - & used in \cite{crammer2013adaptive}  & Proprietary MLP \\ 
\noalign{
\hrule height 2pt
}
\multirow{4}{*} {2015} & \multirow{3}{*} {Gupta et. al \cite{gupta2015deep} } & \multirow{2}{*} {Stochastic Rounding} & 16 bits & 16 bits  & 16 bits & MNIST & Proprietary MLP , LeNet-5 \\ \cline{4-8}
& & & 20 bits & 20 bits & 20 bits & CIFAR-10& used in \cite{hinton2012improving}\\  \cline{2-8}
 & Binary Connect \cite{courbariaux2015binaryconnect} & Stochastic Binarization & 1 bit & 1 bit & Float 32 \tnote{2}  & \specialcell{MNIST \\ CIFAR-10\\SVHN} & Proprietary MLP, CNN \\ \noalign{
\hrule height 2pt
}
\multirow{5}{*}{2016} & Lin et. al \cite{lin2015neural} & \specialcell{Stochastic Binarization \\No forward pass multiplication\\
Quantized back propagation}  & 1 bit & 1 bit & Float 32 & \specialcell{MNIST \\ CIFAR-10\\SVHN} & \specialcell{Proprietary \\ MLP, CNN} \\ \cline{2-8}
& Bitwise Net \cite{kim2016bitwise} & \specialcell{Weight Compression\\
Noisy back propagation}  & 1 bit & 1 bit & \specialcell{1 bit \\ Float 32\tnote{3}} & MNIST & Proprietary MLP\\ \cline{2-8}
& XNOR-Net \cite{rastegari2016xnor} & \specialcell{Binary convolution\\Binary dot-product\\
Scaling binary gradient}  & 1 bit & 1 bit & \specialcell{1 bit \\ Float 32\tnote{4}} & ImageNet & \specialcell{AlexNet \\ ResNet-18 \\ GoogLenet} 
\\ \cline{2-8}
& \multirow{2}{*}{DoReFa-Net \cite{zhou2016dorefa}} & \multirow{2}{*}{\specialcell{stochastic gradient quantization \\ arbitrary bit-width}} &\multirow{2}{*}{1-8 bit} & \multirow{2}{*}{1-8 bit} & \multirow{2}{*}{2-32 bit} & SVHN & proprietary CNN \\ \cline{7-8}
&& & & && ImageNet & AlexNet

\\ \noalign{
\hrule height 2pt
}
\multirow{4}{*}{2017} & \multirow{4}{*}{QNN \cite{hubara2017quantized}} & \multirow{4}{*}{\specialcell{Deterministic binarization \\ Straight through estimators \\ to avoid saturation \\ Shift based Batch Normalization \\ Shift based AdaMAX}}  & \multirow{3}{*}{1 bit}& \multirow{3}{*}{1 bit} & \multirow{3}{*}{1 bit \tnote{5}} &  & \\
& & & & & & MNIST & proprietary MLP \\ \cline{7-8}
& & & & & & \specialcell{CiFAR-10\\SVHN} & CNN from \cite{courbariaux2015binaryconnect} \\ \cline{7-8}
& & & & & & \specialcell{ImageNet} & \specialcell{AlexNet\\GoogLenet} \\ \cline{4-8}
 & & & 4 bit & 4 bit & 4 bit \tnote{6} & \specialcell{Penn \\ Treebank} & \specialcell{proprietary RNN\\LSTM}  \\
\noalign{
\hrule height 2pt
}
\multirow{6}{*}{2018} & \multirow{3}{*}{Wang et. al \cite{wang2018training} } & \multirow{3}{*}{\specialcell{novel floating point\\ chunk based accumulation \\ stochastic rounding}}  & \multirow{3}{*}{8 bit}  & \multirow{3}{*}{8 bit} & \multirow{3}{*}{8 bit \tnote{7}} & CIFAR-10 & \specialcell{proprietary CNN\\ResNET}
\\\cline{7-8}
&& & & && BN50 \cite{van2017training}& proprietary MLP 
\\ \cline{7-8}
&&&&&& ImageNet & \specialcell{AlexNet\\ResNET18\\ResNET50}
\\ \cline{2-8}
& \multirow{3}{*}{Jacob et. al \cite{jacob2018quantization}} & \multirow{3}{*}{\specialcell{training with simulated\\quantization}}& \multirow{3}{*}{8 bit} & \multirow{3}{*}{8 bit} & \multirow{3}{*}{8 bit \tnote{8}} & Imagenet &  \specialcell{Resnet\\Inception v3\\MobileNet}  \\ \cline{7-8}
&& & & && COCO & MobileNet SSD  
\\ \cline{7-8}
&& & & && Flickr \cite{howard2017mobilenets} & MobileNet SSD 

\\ 
\noalign{
\hrule height 2pt
}
2019
& WAGEUBN \cite{yang2020training} & \specialcell{batch-norm layer quantization \\8-bit integer representation \\combination of direct, constant \\and shift quantization} &8 bit & 8 bit & 8 bit & ImageNet & ResNet18/34/50 \\
\noalign{
\hrule height 2pt
}
\multirow{12}{*}{2020}
&\multirow{4}{*}{S2FP8 \cite{cambier2020shifted}} & \multirow{4}{*}{\specialcell{shifted and squeezed FP8 \\ representation of tensors \\ tensor distribution learning }} &\multirow{4}{*}{8 bit} & \multirow{4}{*}{8 bit} & \multirow{4}{*}{32 bit} & CIFAR-10 & ResNet20/34/50 
\\ \cline{7-8}
&& & & && ImageNet & ResNet18/50
\\ \cline{7-8}
&& & & && English-Vietnamese & Transformer-Tiny
\\ \cline{7-8}
&& & & && MovieLens & \specialcell{Neural Collaborative\\ Filtering (NCF)}
\\ \cline{2-8}
& \multirow{3}{*}{Wiedemann et. al \cite{wiedemann2020dithered}} &
\multirow{3}{*}{\specialcell{stochastic gradient quantization \\ induce sparsity \\ non-subtractive dither}} &\multirow{3}{*}{8 bit} & \multirow{3}{*}{8 bit} & \multirow{3}{*}{32 bit} & MNIST & LeNet
\\ \cline{7-8}
&& & & && CIFAR-10/100 & \specialcell{AlexNet\\ResNet18\\VGG11}
\\ \cline{7-8}
&& & & && ImageNet &ResNet18
\\ \cline{2-8}
& \multirow{3}{*}{Quant-Noise \cite{fan2020training} }& \multirow{3}{*}{\specialcell{training using\\quantization noise}} & \multirow{3}{*}{8 bit} & \multirow{3}{*}{8 bit} & \multirow{3}{*}{8 bit} & Wikitext-103 & RoBERT \\ 
&& & & && MNLI & RoBERT\\
&& & & && ImageNet & EfficientNet-B3\\
\\ \noalign{
\hrule height 2pt
}
\end{tabular}

\begin{tablenotes}
 \item[1] minimum quantization for best performing model reported.
 \item[2] all real valued vectors are reported as Float 32 by default.
 \item[3] involves tuning a separate set of parameters with floating point precision.
 \item[4] becomes Float 32 if gradient scaling is used.
 \item[5] except the first layer input of 8 bits.
 \item[6] contains results with 2 bit, 3 bit and floating point precision.
 \item[7] additional 16 bit for accumulation.
 \item[8] uses 7 bit precision for some Inception v3 experiments.
\end{tablenotes}
\end{scriptsize}
\end{threeparttable}
\end{table}

Apart from quantization, there are mainly two other approaches targeted at reducing memory footprint of DNN training, namely, layer-wise training \cite{greff2016highway, chen2016training}, and trading computation for memory \cite{gruslys2016memory}. One major cause for the heavy memory footprint of DNN training is end-to-end back-propagation which requires storing all intermediate feature maps for gradient calculation. Sequential layer-by-layer training has been proposed as an alternative \cite{chen2016training}. While this method was originally proposed for better DNN interpretation, it requires less memory usage while retaining the generalization ability of the network. Trading computation for memory reduces memory footprint by releasing some of the memory allocated to intermediate results, and recomputing these results as needed. This approach is shown to tightly fit within almost any user-set memory budget while minimizing the computational cost \cite{gruslys2016memory}. Apart from these two approaches, some other studies focus on exploiting the specific DNN architectures and targeted datasets, and propose methods using novel loss functions \cite{zen2016fast, chen2016efficient}, and training pipelines \cite{chen2016efficient} to improve the memory and computation efficiency of DNNs. 

While the aforementioned studies demonstrate promising experimental results, they barely provide any analysis on the generalization error nor the sample complexity under a limited memory budget. A few studies target at this issue by developing memory-bounded learning techniques with some performance guarantees. For example, \cite{langford2009sparse, golovin2013large, steinhardt2015minimax, tai2018sketching} proposed memory-bounded optimization routines for sparse linear regression and provided upper bounds on regret in an online learning setting.  For PCA, \cite{mitliagkas2013memory,allen2017first,zhang2017randomization,li2016rivalry,yang2015streaming} proposed memory or computation-efficient optimization algorithms while guaranteeing optimal sample complexity.

Note that in an environment with a cluster of CPUs or GPUs, the most popular technique for speeding up training is parallelization, where multiple computational entities share the computation and storage for both data and model \cite{dean2012large}. However, in IoT applications with embedded systems, it is not common to have similar settings as a computer cluster. Usually the devices are scattered and operated in a distributed fashion. To enable cooperation among a group of devices, distributed learning techniques such as consensus, federated learning and many others have been proposed \cite{tsianos2012consensus, mcmahan2017communication}. However, the primary concern of these techniques are data privacy and the lack of central control entity in the application field, which is out of the scope of this paper.

\noindent \textbf{Resource constrained Generic Optimization Routines}: While the methods in the previous subsection are designed for specific algorithms, the approaches in this section target generic optimization routines. One of the first works  in this line of research is BuckWild! \cite{de2015taming} which introduces low precision SGD and provides theoretical convergence proofs of the algorithm. An implementation of the Buckwild! algorithm through a computation model called "Dataset Model Gradients Communicate" (DMGC) is provided in \cite{de2017understanding}. As an improvement to the Buckwild! approach, the work in \cite{de2018high} proposes novel `bit-centering' quantization for low precision SGD and stochastic variance reduction gradient (SVRG). An alternative approach to improve upon low-precision SGD in \cite{de2015taming} is also introduced in \cite{yang2019swalp}. This approach introduces the SWALP algorithm, an extension of the stochastic weight averaging (SWA) scheme for SGDs for low precision arithmetic. While most of the above approaches mainly analyze the effect of quantization to SGD and its variants; \cite{stich2018sparsified} adopts an approach of improving the SGD algorithm under sparse representation. Finally, in a more recent work, \cite{li2019dimension} provides the first dimension-free bound for the convergence of low precision SGD algorithms. There has been significant research on analyzing traditional first order algorithms like SGD, SVRG etc., in parallel and distributed settings with low precision / quantized bit representation \cite{lin2017deep,wu2018error,strom2015scalable,alistarh2017qsgd,wangni2018gradient,tang2018communication,gandikota2019vqsgd,wen2017terngrad,mayekar2019ratq}. Even though these approaches generalize to single device learning, the main focus of these research is in reducing the communication bottleneck for parallel and distributed settings. This area is not within the scope of the current survey. A more detailed survey on communication-efficient distributed optimization is available in \cite{lim2020federated,gu2019distributed,li2020federated,tang2020communicationefficient}.  

An alternative line of work involves implementing fixed point Quadratic Programs (QP) for solving linear Model Predictive Control (MPC). Most of these algorithms involve modifying the fast gradient methods developed in \cite{nesterov2013introductory} to obtain a suboptimal solution in a finite number of iterations under resource constrained settings. In fact these algorithms extend the generic Interior point methods, Active-set methods, Fast Gradient Methods, Alternating Direction method of multipliers and Alternating minimization algorithm, to handle quantization error introduced due to fixed point implementations. However, all these approaches are targeted towards linear MPC problems, and is not the main focus of this survey. A comprehensive survey on these fixed point QPs for linear MPCs is available in \cite{mcinerney2018survey}.

\subsubsection{Data Compression} \label{sec:algo:dat_compress}

Besides complexity in models or optimization routines, the dimensionality and volume of training data significantly dictates the algorithm design for on-device learning. One critical aspect is the \textit{sample complexity} \footnote{Formal introduction provided in Section \ref{sec:theory:trad_learn_theory}} i.e. amount of training data needed by the algorithm to achieve a desired statistical accuracy. This aspect has been widely studied by the machine learning community and has led to newer learning settings like Semi-Supervised learning \cite{chapelle2009semi}, Transductive learning \cite{chapelle2009semi,vapnik2006estimation}, Universum learning \cite{vapnik2006estimation,dhar2019multiclass}, Adversarial learning \cite{goodfellow2014explaining}, Learning under Privileged Information (LUPI) \cite{vapnik2006estimation}, Learning Using Statistical Invariants (LUSI) \cite{vapnik2019rethinking} and many more. These settings allow for the design of advanced algorithms that can achieve high test accuracies even with limited training data. A detailed coverage of these approaches is outside the scope of this survey but readers are directed to \cite{chapelle2009semi,vapnik2006estimation,dhar2019multiclass,goodfellow2014explaining,vapnik2019rethinking,van2020survey,geng2020recent,zhu2005semi,cherkassky2007learning} for a more exhaustive reading.     

In this paper we focus on those approaches that reduce the resource footprint imposed by the dimensionality and the volume of training data. 

 For example, the storage and memory requirement of k-Nearest Neighbor (kNN) approaches are large due the fact that all training samples need to be stored for inference. Therefore, techniques that compress training data are usually used to reduce the memory footprint of the algorithm \cite{kusner2014stochastic, wang2016deep,zhang2016zipml}. Data compression or sparsification is also used to improve DNN training efficiency. Some approaches exploit matrix sparsity to reduce the memory footprint to store training data. For example, \cite{rouhani2017tinydl} transforms the input data to a lower-dimensional embedding which can be further factorized into the product of a dictionary matrix and a block-sparse coefficient matrix. Training using the transformed data is shown to be more efficient in memory, runtime and energy consumption. In \cite{li2016lightrnn}, a novel data embedding technique is used in RNN training, which can represent the vocabulary for language modeling in a more compressed manner. 

\subsubsection{New Protocols for Data Observation} \label{sec:algo:new_protocols}

For most learning algorithms, a major assumption is that full i.i.d. data is accessible in batch or streaming fashion. However, under certain constraints, only partial features of part or the entire dataset can be observed. For example, feature extraction can be costly due to either expensive feature computation or labor-intensive feature acquisition. Another instance is that memory is not sufficient to store a small batch of high-dimensional data samples. This problem leads to studies on limited attribute observation \cite{ben1998learning, cesa2011efficient}, where both memory and computation are limited by restricted observation of sample attributes. Under this setting, a few studies propose new learning algorithms and analyze sample complexity for applications such as sparse PCA\cite{shamir2014fundamental}, sparse linear regression \cite{cesa2011efficient, hazan2012linear, murata2018sample, ito2018online}, parity learning\cite{steinhardt2016memory}, etc. While this new learning protocol provides valuable insights into the trade-offs of sample complexity, data observability, and memory footprint, it remains to be explored how they can be used to design resource-efficient algorithms for a wider range of ML models. 



\subsection{Resource Efficient Inference} \label{sec:algo:resource_inference}
While inference is not the main focus of this paper, for the sake of completeness, we provide a brief overview of resource efficient approaches that enable on-device inference. Due to the fact that most traditional ML algorithms are not resource intensive during inference, we will only focus on methods proposed for DNNs.

Apart from designing lightweight DNN architectures as discussed in Section \ref{subsec:reducemodelcomplexity}, another popular approach is static model compression, where, during or after training, the model is compressed in size via techniques such as network pruning \cite{han2015deep, liu2020pruning}, vector quantization \cite{gong2014compressing, han2015deep, wang2018haq}, distillation \cite{44873}, hashing \cite{chen2015compressing}, network projection \cite{ravi2017projectionnet}, binarization \cite{courbariaux2016binarized}, etc. These studies demonstrate significant reduction in model size while retaining most of the network's predictive capacity. 

However, since these models will not change after compression, their complexities cannot further adapt to dynamic on-device resources or inputs. To address this issue, some novel adaptive techniques for faster inference on embedded devices are explored in recent works. For example, \cite{dey2019embedded} uses dynamic layer-wise partitioning and partial execution of CNN based model inference, allowing for more robust support of dynamic sensing rates and large data volume. Another approach proposed in \cite{marco2020optimizing} involves developing a low cost predictive model which dynamically selects models from a set of pre-trained DNNs by weighing desired accuracy and inference time as metrics for embedded devices. \cite{lee2019neuro} introduced Neuro.ZERO to provide energy and intermittence aware DNN inference and training along with adaptive high-precision fixed-point arithmetic to allow for accelerated run-time embedded hardware performance. 

There have been observations that not all inputs require the same amount of computational power to be processed. A simple model may be sufficient to classify samples that are easy to distinguish, while a more complex model is only needed to process difficult samples. Based on this reasoning, \cite{bolukbasi2017adaptive, wang2017skipnet, huang2017multi} focus on building and running adaptive models that dynamically scale computation according to inputs. The specific methods proposed in these works include early termination, exploration of cascaded models, and selectively using/skipping parts of the network. These adaptive approaches are complementary to the compression approach, and further resource efficiency can be expected if these two approaches are applied altogether. 

Apart from the `model-centric` approaches as discussed in Section \ref{sec:algo:modify_opt}, hardware and system optimizations have also been exploited for efficient model deployment. To explore how the choice of computing devices can impact performance of these models on the edge we direct the reader to \cite{murshed2019machine}. The survey discusses how various low power hardware such as ASICs, FPGAs, RISC-V, and embedded devices are used for efficient inferencing of Deep Learning models. Another interesting survey \cite{wang2020convergence} expands on the various communication and computation modes for deep learning models where edge devices and the cloud server work in tandem. Their work explores concepts such as integral offloading, partial offloading, vertical collaboration, and horizontal collaboration to allow efficient edge-based inference in collaboration with the cloud. \cite{zhou2019edge} further advances these ideas from the point of view of the network latency between the cloud server and edge device, exploring techniques such as model partitioning, edge caching, input filtering, early exit strategies etc., to provide efficient on device inference. Lately, the embedded ML research community has also focused on interconnected and smart home devices for processing data privately and locally with stronger privacy restrictions and lower latencies. \cite{hu2019deephome} aggregates processing capability of potential embedded devices at home with comparisons between processing on mobile phones and specific hardware for efficient DL processing such as Coral TPU and the NVIDIA Jetson Nano. 

\subsection{Challenges in Resource-Efficient Algorithm Development}
\label{subsec:algchallenges}
From the discussions in this section, we observe the following challenges for resource characterization of existing ML algorithms and development of new resource-efficient algorithms:
\begin{enumerate}
\item \textbf{Hardware dependency}: As observed from existing hardware platform benchmarks and resource requirement modeling approaches, resource characterization greatly depends on the platform, framework, and the computing library used. For example, on the Nvidia Jetson TX1 platform, the memory footprint of a CNN model can be very different based on the frameworks (Torch vs. Caffe) used \cite{canziani2016analysis, rodriguesfine}. Among all resource constraints, memory footprint is particularly hard to quantify as many frameworks and libraries deploy unique memory allocation and optimization techniques. For example, MXNet provides multiple memory optimization mechanisms and the memory footprint is very different under each mechanism \cite{MXNetmemory}. Consequently, given specific on-device resource budgets and performance requirements, it is challenging to choose the optimal algorithm that fits in the resource constraints.

With all the variability and heterogeneity in implementation, it is thereby important to draw insights and discover trends on what is invariant across platforms. As discussed in \cite{marculescu2018hardware}, cross-platform models which account for chip variability are needed to transfer knowledge from one type of hardware to another. Furthermore, we postulate that cross-framework or cross-library models are also important to enable resource prediction for all types of implementations.

\item \textbf{Metric design}:While a few novel metrics are proposed for on-device ML algorithm analysis, they either cannot be used directly for algorithm comparison, or cannot provide guarantees that the training process will fit the resource budget. Particularly, for metrics devised based on hardware agnostic measures such as number of parameters or FLOPs, they rarely accurately reflect  realistic algorithm performance on a specific platform. Therefore, more practical, commensurable and interpretable metrics need to be designed to determine which algorithm can provide the best trade-off between accuracy and multiple resource constraints.

\item \textbf{Algorithm focus}: Most studies on on-device learning focus on DNNs, especially CNNs, because their layer-wise architectures carry a declarative specification of computational requirements. In contrast, there has been very little focus on RNNs and traditional machine learning methods. This is mainly due to their complexity or heterogeneity in structure or optimization method, and lack of a benchmarking dataset. However, as observed in Table \ref{table:DNNbenchmark}, the edge platforms used for DNN profiling are generally mobile phones or embedded computing cards which still have gigabytes of RAM and large computational power. For IoT or embedded devices with megabytes or even kilobytes of RAM and low computational power, DNNs can barely fit. Therefore, traditional machine learning methods are equally important for edge learning, and their resource requirements and performance on edge devices need to be better profiled, estimated and understood. In addition, designing advanced learning algorithms that require only small amounts of training data to achieve high test accuracies is a huge challenge. Addressing this area can significantly improve the resource footprint (i.e. low memory footprint to store the training data), while maintaining highly accurate models.    

\item \textbf{Dynamic resource budget}: Most studies assume that the resource available for a given algorithm or application is static. However, resource budget for a specific application can be dynamic on platforms such as mobile phone due to events such as starting or closing applications and application priority changes \cite{fang2018nestdnn}. Contention can occur when resource is smaller than the model requirements and resource can be wasted when it is abundant. To solve this issue, \cite{fang2018nestdnn} proposed an approach for multi-DNN applications. When trained offline, for each application, five models with diverse accuracy-latency tradeoffs are generated and their resource requirements are profiled. The models are nested with shared weights to reduce memory footprint. During deployment, a greedy heuristic approach is used to choose the best models for multiple applications at runtime that achieve user-defined accuracy and latency requirements while not exceeding the total memory of the device.



\end{enumerate}

\section{Theoretical Considerations for On-Device Learning}
\label{sec:theory}

\begin{table}[h]
\caption{Broad Categorization of resource constrained algorithms in \autoref{sec:algo} with respect to their underlying learning theories.}  
\label{tab::TheoryvsAlgos}
\begin{footnotesize}
\begin{tabular}{|c|l|}  
\hline
Theory & Algorithms \\
\noalign{
\hrule height 2pt
}
\multirow{4}{*}{Traditional Machine Learning Theory} & Low-Footprint ML algorithms (section \ref{sec:algo:low_footprint})\\
& Reducing model-complexity (section \ref{subsec:reducemodelcomplexity}) \\
&  Modifying Optimization routines (section \ref{sec:algo:modify_opt})\\
& Data Compression (section \ref{sec:algo:dat_compress})\\ \hline
Novel resource constrained machine learning theory & New protocols for data observation (section \ref{sec:algo:new_protocols})\\
\hline
\end{tabular}
\end{footnotesize}
\end{table}

While Section \ref{sec:algo} categorizes the approaches used for building resource-constrained machine learning models, this section focuses on the computational learning theory that is used to design and analyze these algorithms. Nearly every approach in the previous section is based on one of the following underlying theories,
\begin{itemize}
\item \textbf{Traditional machine learning theory}: Most of the existing approaches used to build the resource efficient machine learning models for inference (in section \ref{sec:algo:resource_inference}) follow what is canonically referred to as traditional machine learning theory. In addition,  approaches like, \textit{low-footprint machine learning} (section \ref{sec:algo:low_footprint}), \textit{reducing model complexity} (section \ref{subsec:reducemodelcomplexity}), \textit{data compression} (section \ref{sec:algo:dat_compress}), \textit{modifying optimization routines} (section \ref{sec:algo:modify_opt}) used for resource efficient training, also follow this traditional learning theory. For example, low-footprint machine learning involves using algorithms with inherently low resource footprints designed under traditional learning theory. Approaches like \textit{reducing model complexity} and \textit{data compression}, incorporates additional resource constraints for algorithms designed under traditional learning theory. Finally, the approach of \textit{modifying optimization routines} simply modifies the optimization algorithms used to solve the machine learning problem designed under such theories.     

\item \textbf{Novel resource constrained machine learning theory}: These approaches highlight the gaps in traditional learning theory and propose newer notions of learnability with resource constraint settings. Most of the algorithms in \textit{new protocols for data observation} (section \ref{sec:algo:new_protocols}) fall in this category. Typically, such approaches modify the traditional assumption of i.i.d data being presented in a batch or streaming fashion and introduces a specific protocol of data observability. Additional algorithmic details are available in \ref{sec:algo:new_protocols}.  
\end{itemize}

In this section we provide a detailed survey of the advancements in the above categories. Note that there is an abundance of literature addressing traditional learning theories \cite{vapnik2006estimation, shalev2014understanding, mohri2012foundations}. For completeness however we still include a very brief summary of such traditional theories in section \ref{sec:theory:trad_learn_theory}. This summary also acclimatizes the readers with the notations used throughout this section. 

Section \ref{sec:theory:learn_prob_formalize} formalizes the learning problem for both traditional as well as resource constrained settings. Next, section \ref{sec:theory:trad_learn_theory} provides a brief survey of some of the popular traditional learning theories. The advanced learning theories developed for resource-constraint settings are provided in section \ref{sec:theory:res_const_learn_theory}. Finally, we conclude by discussing the existing gaps and challenges in section \ref{sec:theory:challenges}.


\subsection{Formalization of the Learning problem} \label{sec:theory:learn_prob_formalize}
Before delving into the details of the learning theories, we first formalize the learning problem for traditional machine learning algorithms. We then extend this formalization to novel machine learning algorithms under resource constraint settings in section \ref{sec:theory:res_const_learn_prob}. For simplicity we focus on supervised problems under inductive settings \cite{vapnik2006estimation,cherkassky2007learning}. Extensions to other advanced learning settings like transductive, semi-supervised, universum etc., can be found in \cite{vapnik2006estimation, shalev2014understanding, mohri2012foundations}.       


\subsubsection{Traditional Machine Learning Problem} \label{sec:theory:trad_learn_prob}

The supervised learning problem assumes an underlying label generating process $y = g(\mathbf{x})$, where $y \in \mathcal{Y}$ and  $\mathbf{x} \in \mathcal{X} \text{(data domain)} \subseteq \mathbb{R}^d$ ; which is characterized by a data generating distribution $\mathcal{D}$. Under the inductive learning setting, the machine learning problem can be formalized as,

\begin{definition} \textbf{Inductive Learning} 
\begin{enumerate}
\item[\textbf{Given.}] independent and identically distributed (i.i.d) training samples $S = (\mathbf{x}_i,y_i)_{i=1}^n$ from the underlying data generating distribution $S \sim \mathcal{D}^n$, and a predefined loss function $l:\mathcal{X}\times \mathcal{Y} \rightarrow \mathcal{O}$ (output domain). 
\item[\textbf{Task.}] estimate a function/hypothesis $\hat{h}:\mathcal{X} \rightarrow \mathcal{Y}$ from a set of hypotheses $\mathcal{H}$ (a.k.a Hypothesis class), that best approximates the underlying data generating process $y = g(\mathbf{x})$ i.e. which minimizes 
\begin{flalign} \label{eqloss}
\mathbb{E}_{\mathcal{D}}(l(y , \hat{h}(\mathbf{x})) = R(\hat{h})
\end{flalign}
\end{enumerate}
Here,
\begin{flalign}
\mathbb{E}_{\mathcal{D}}(\cdot) &= \text{Expectation operator under distribution } \mathcal{D} ; \quad \mathbbm{1}_{(\cdot)} = Indicator function && \nonumber \\ 
\text{and } R(\cdot) &= \text{True risk of a hypothesis} && \nonumber 
\end{flalign}
\end{definition}  

Some popular examples following this problem setting are,
\begin{itemize}
\item Binary Classification problems using 0/1 loss where $\mathcal{Y} = \{-1,+1\}$, $\mathcal{O} = \{0,1\}$ and $l(y , \hat{h}(\mathbf{x})) = \mathbbm{1}_{y \neq \hat{h}(\mathbf{x})}$.
\item Regression problems with additive gaussian noise (i.e. least-square loss) where $\mathcal{Y} = \mathbb{R}$, $\mathcal{O} = \mathbb{R}$ and $l(y , \hat{h}(\mathbf{x})) = (y - \hat{h}(\mathbf{x}))^2$.
\end{itemize}

 
Typically users incorporate apriori information about the domain while constructing the hypothesis class. In the simplest sense, the hypothesis class gets defined through the methodologies used to solve the above problems. For example, solving the regression problem using,
\begin{itemize}
\item \textit{least-squares} linear regression: the hypothesis class includes all \textit{linear} models i.e. $\mathcal{H} = \{h : h(\mathbf{x}) = \mathbf{w}^T \mathbf{x} + b ;  \mathbf{w} \in \mathbb{R}^d; b \in \mathbb{R}\}$.
\item \textit{lasso} $L1-$ linear regression: the hypothesis class includes all \textit{linear} models of the form:   $\mathcal{H} = \{h : h(\mathbf{x}) = \mathbf{w}^T \mathbf{x} + b ;  \mathbf{w} \in \mathbb{R}^d; b \in \mathbb{R} ; ||\mathbf{w}||_1 \leq B \}$.
\end{itemize}

\subsubsection{Resource Constrained Machine Learning Problem} \label{sec:theory:res_const_learn_prob}
In addition to the goals highlighted in the above section, building resource constrained machine learning algorithms introduces additional constraints to the learning problem. The goal now is not just to minimize the true risk, but also ensure that the resource constraints discussed in section \ref{sec:resource_constraints} are met. Typically, these resource constraints are imposed during \textit{inference} or \textit{training} phase of a machine learning pipeline. 
A mathematical formalization of this problem can therefore be given as,
\begin{definition} \textbf{Resource Constrained Machine Learning} 
\begin{enumerate}
\item[\textbf{Given.}] i.i.d training samples $S = (\mathbf{x}_i,y_i)_{i=1}^n$ from the underlying data generating distribution $S \sim \mathcal{D}^n$, a predefined loss function $l:\mathcal{X}\times \mathcal{Y} \rightarrow \mathcal{O}$, and a predefined resource constraint $\mathcal{C}(\cdot)$. 
\item[\textbf{Inference.}] estimate a function $\hat{h}:\mathcal{X} \rightarrow \mathcal{Y}$ that best approximates the underlying data generating process i.e. eq. \eqref{eqloss} and simultaneously satisfies  $\mathcal{C}(\hat{h})$. 
\item[\textbf{Training.}] estimate a function $\hat{h}:\mathcal{X} \rightarrow \mathcal{Y}$ that best approximates the underlying data generating process i.e. eq. \eqref{eqloss} and simultaneously satisfies  $\mathcal{C}(A(S))$. Here, $A:S \rightarrow \mathcal{H}$ is an algorithm (model building process) that inputs the training data and outputs a hypothesis $\hat{h} \in \mathcal{H}$.
\end{enumerate}
\end{definition}  

As seen above the main difference of this setting with respect to the traditional inductive learning setting is the additional constraint $\mathcal{C}(\cdot)$ on the final model (during inference) or the Algorithm (for training). As an example, consider the least squares linear regression example discussed above. Typical resource constraints for such a problem may look like,
\begin{itemize}
\item Memory constraint during inference enforced by requiring the model parameters $(\mathbf{w},b)$ of the estimated model $y = \mathbf{w}^T \mathbf{x} + b$ be represented using float32 precision ($\sim 2^{32(d+1)}$ bit memory footprint) or int16 precision ($\sim 2^{16(d+1)}$ bit memory footprint); with $\mathbf{x} \in \mathbb{R}^d$. This is equivalent to a constraint on the final model $\mathcal{C}_{\text{float32}}(\hat{w},\hat{b}) = \{ \hat{w} \in \{-3.4E+38,\ldots,3.4E+38\}^d ;\hat{b} = \{-3.4E+38,\ldots,3.4E+38\} \}$ or $\mathcal{C}_{\text{int16}}(\hat{w},\hat{b}) = \{\hat{w} = \{-32768,\ldots , 32767\}^d ;\hat{b} =  \{-32768,\ldots , 32767\} \}$. 
\item Computation constraint during inference can be enforced by requiring $\hat{w}$ to be $k-$ sparse with $k<d$. This, would ensure $k+1$ FLOPS per sample and equivalently constraints final model as $\mathcal{C}(\hat{w},\hat{b}) = \{ \hat{w} \in \mathbb{R}^k \}$.  
\end{itemize}
Similar, memory/computation constraint can be imposed onto the learning algorithm $A(\mathcal{S})$ during training.

\subsection{Learning theories} \label{sec:learn_theory}
In this section we discuss the learning theories developed for the problems discussed above.

\subsubsection{Traditional Learning Theories} \label{sec:theory:trad_learn_theory}

These theories target the traditional learning problem discussed in \ref{sec:theory:learn_prob_formalize}. Most traditional learning theories provide probabilistic guarantees of the goodness of a model (actually guarantees for all models in the Hypothesis class) with respect to the metric in eq.\eqref{eqloss}. Such theories typically decompose eq. \eqref{eqloss} into two main components,  
\begin{flalign} \label{errordecomp}
R(h) - R^* = \underbrace{(\underset{h \in \mathcal{H}}{min}\; R(h)-R^*)}_{\text{approximation error}} + \underbrace{(R(h)-\underset{h \in \mathcal{H}}{min}\; R(h))}_{\text{estimation error}}
\end{flalign}
Here,
\begin{flalign}
\text{Bayes Error, } R^* &= \underset{\text{all measurable } h}{min}\; R(h) && \nonumber \\
\text{and} \; \underset{h \in \mathcal{H}}{min}\; R(h)) & = \text{smallest in-class error} && \nonumber
\end{flalign}

The approximation error in eq. \eqref{errordecomp}, typically depends on the choice of the hypothesis class and the problem domain. The value of this error is problem dependent. Traditional learning theories then characterize the estimation error. One of the fundamental theoretical frameworks used for such analyses is the PAC-learning framework.

\begin{definition} \label{th::PAC}
\textbf{(Efficient) Agnostic PAC-learning} \cite{mohri2012foundations}\\
A hypothesis class $\mathcal{H}$ is agnostic PAC-learnable using an algorithm $A(S)$, if for a given $\epsilon,\delta \in (0,1)$ and for any distribution $\mathcal{D}$ over $\mathcal{X} \times \mathcal{Y}$ (with $\mathcal{X} \subseteq R^d$) there exists a polynomial function $p_1(\frac{1}{\epsilon},\frac{1}{\delta},d,|\mathcal{H}|)$ such that for any sample size $n \geq p_1(\frac{1}{\epsilon},\frac{1}{\delta},d,|\mathcal{H}|) $ the following holds,
\begin{flalign} \label{agnosticPAC}
\mathbb{P}_{\mathcal{D}}(R(h_S)-\underset{h \in \mathcal{H}}{min}\; R(h) \leq \epsilon) \geq 1-\delta; \quad where \; h_S = A(S) \in \mathcal{H}
\end{flalign} 
Further it is efficiently PAC learnable if the computation complexity is $O(p_2(\frac{1}{\epsilon},\frac{1}{\delta},d,|\mathcal{H}|))$ for some polynomial function $p_2$.   
\end{definition}

As seen from Definition \ref{th::PAC} there are two aspects of PAC-learning.

\noindent \textbf{Sample Complexity characterizing Generalization Bounds:} Although Definition \ref{th::PAC} provides the learnability framework for a wide range of machine learning problems like regression, multi-class classification, recommendation etc. For simplicity, in this section we mainly focus on the \textit{sample-complexity} (and equivalently the generalization bounds) for binary classification problems. Extensions of these theories for other learning problems can be found in \cite{shalev2014understanding,mohri2012foundations}.

Definition \ref{th::PAC} guarantees that an estimated model using the algorithm $A(S)$ and sample size $n \geq p_1(\frac{1}{\epsilon},\frac{1}{\delta},d,|\mathcal{H}|)$, will guarantee predictions with error tolerance of $\epsilon$ and a probability $1-\delta$. One popular class of algorithm choice is the Empirical Risk Minimization (ERM) based algorithms. Here, the algorithms return the hypothesis that minimizes an empirical estimate of the risk function in \eqref{eqloss} given by, 
\begin{flalign} \label{eq::ERM}
\text{ERM estimate } & h_S \;= \; ERM(S) = \underset{h \in \mathcal{H}}{argmax} \; R_S(h) && \\
\text{where, Empirical Risk } & R_S(h) \; = \; \frac{1}{n} \sum\limits_{i=1}^n \mathbbm{1}_{y_i \neq h(\mathbf{x}_i)} && \nonumber
\end{flalign}


A very interesting property of the ERM based algorithms is the \textit{uniform convergence} property (see \cite{shalev2014understanding}). This property dictates that for the ERM algorithm to return a \textit{good} model from within $\mathcal{H}$, we need to bound the term  $|R_S(h) - R(h)|; \forall h \in \mathcal{H}$. In fact, most popular learning theories characterize the number of samples (a.k.a sample complexity) needed for bounding this term $|R_S(h) - R(h)|; \forall h \in \mathcal{H}$ for any given $\epsilon,\delta \in (0,1)$. The canonical form adopted in most such learning theories is provided next,



\begin{definition} \textbf{Canonical Forms} \\ \\
\underline{Generalization Bound}: For a given hypothesis class $\mathcal{H}$, training set $S \sim \mathcal{D}^n$, and $\epsilon, \delta \in (0,1)$, a typical generalization bound adopts the following form,
\begin{flalign} \label{eq::conc_canonical}
\text{with probability at least } 1-\delta \text{ we have, } \forall h \in \mathcal{H} \quad R(h) \leq R_S(h)+ \underbrace{ O(p_1(\frac{1}{\delta},n,|\mathcal{H}|))}_{\text{confidence term}}
\end{flalign}
A direct implication of \eqref{eq::conc_canonical} and uniform convergence property is the sample complexity provided next, \\ \\
\underline{Sample Complexity} A hypothesis class $\mathcal{H}$ is PAC learnable using ERM based algorithms with a training set $S$ of sample complexity $n \geq p_2(\frac{1}{\epsilon},\frac{1}{\delta},|\mathcal{H}|)$
\end{definition}

The exact forms of the confidence terms and the sample complexity term depends on the way the complexity of the hypothesis class is captured. A brief survey of some the popular theories used to capture the complexity of the hypothesis class is provided in Table \ref{tab::SLT}. 




\begin{sidewaystable}[htbp]
\caption{Popular Traditional Learning Theories} 
\label{tab::SLT}
\scriptsize
\begin{tabular}{|c|p{0.1\textwidth}|p{0.15\textwidth}|c|p{0.37\textwidth}|}  
\toprule
\multicolumn{2}{|c|}{\textbf{Theory}} & \specialcell{\textbf{Confidence term}\\$p_1(\frac{1}{\delta},n,|\mathcal{H}|)$} & \specialcell{\textbf{Sample Complexity} \\ $p_2(\frac{1}{\epsilon},\frac{1}{\delta},|\mathcal{H}|)$} & \quad \quad \textbf{Additional  Remarks}\\
\noalign{
\hrule height 2pt
}
\textbf{Finite Hypothesis} ($|\mathcal{H}|<\infty$) &&&&\\
Realizable ($\underset{h \in \mathcal{H}}{min}\; R(h) = 0$) & PAC Bound \cite{valiant1984theory}& $\frac{log|\mathcal{H}| + log\frac{1}{\delta}}{n}$&$\frac{1}{\epsilon}(log|\mathcal{H}| + log\frac{1}{\delta})$&\\ 
\cline{2-5}
Non-Realizable ($\underset{h \in \mathcal{H}}{min}\; R(h) \neq 0$)& Agnostic PAC \cite{valiant1984theory} & $\sqrt{\frac{log|\mathcal{H}| + log\frac{2}{\delta}}{2n}}$& $\frac{log|\mathcal{H}| + log\frac{2}{\delta}}{\epsilon^2}$ &\\ 
\noalign{
\hrule height 2pt
}
\textbf{Countably Infinite} ($\mathcal{H} = \cup_{n \in N} h_n$) & MDL \cite{shalev2014understanding}& $\sqrt{\frac{|d(h)| + log\frac{2}{\delta}}{2n}}$& $\frac{log|d(h)| + log\frac{2}{\delta}}{\epsilon^2}$ & $d(h) =$ minimum description length of hypothesis. \\
\noalign{
\hrule height 2pt
}
& VC (dimension) Theory \cite{vapnik71uniform} & 
\begin{itemize}[leftmargin=*]
\item[--] $\sqrt{2\frac{log \mathcal{G}_{\mathcal{H}}(n) + log\frac{2}{\delta}}{n}}$
\item[--] $\sqrt{2\frac{dlog\frac{2en}{d} + log\frac{2}{\delta}}{n}}$
\end{itemize} (Sauer's Lema) & $\frac{log \mathcal{G}_{\mathcal{H}}(n) + log\frac{2}{\delta}}{\epsilon^2}$  & 
\begin{itemize}[leftmargin=*]
\item[--] Growth Function $\mathcal{G}_{\mathcal{H}}(n) = \underset{z_1\ldots z_n\sim \mathcal{D}^n}{sup} |\mathcal{H}_{z_1\ldots z_n}|$.
\item[--] VC dim, $d = \underset{n \in N:\mathcal{G}_{\mathcal{H}}(n) = 2^n}{max} n, \quad z_i=(\mathbf{x}_i,y_i)$.
\end{itemize}
\\
\cline{2-5}
\multirow{7}{*}{\textbf{Infinite Hypothesis}}& VC (Entropy) Theory \cite{vapnik71uniform}& $\sqrt{2\frac{log E_D [N(\mathcal{H}, 2n)] + log\frac{2}{\delta}}{n}}$ & $\frac{log E_D [N(\mathcal{H}, 2n)] + log\frac{2}{\delta}}{\epsilon^2}$ & 
Size of function class $ N(\mathcal{H},n) = |\mathcal{H}_{z_1\ldots z_n}|$.
\\ \cline{2-5}
& Covering Number \cite{wolf2018mathematical} & $\sqrt{\frac{log(4\Gamma_{1}(2n,\varepsilon /8,\mathcal{F})+log(\frac{1}{\delta}))}{2n}}$ & $ \frac{log(4\Gamma_{1}(2n,\varepsilon /8,\mathcal{F})+log(\frac{1}{\delta}))}{\epsilon^2}$ & \begin{itemize}[leftmargin=*]
\item[--] Covering number $\Gamma_p(n,\varepsilon,\mathcal{H})= \text{max} \{N_{in}(\varepsilon,\mathcal{H},||\cdot||_{p,x})|\mathbf{x} \in \mathcal{X} \}$
\item[--] $N_{in}(\varepsilon,\mathcal{H},||\cdot||_{p,x})$ smallest cardinality of internal $\varepsilon$ cover of $\mathcal{H}$ with $||g||_{p,z} = (\frac{1}{n}\sum\limits_{i=1}^n |g(z_i)|^p)^\frac{1}{p}$.
\end{itemize} \\ \cline{2-5}
& Radamacher Complexity  \cite{bartlett2002rademacher} & 
\begin{itemize}[leftmargin=*]
\item[--] $2\mathcal{R}_n(\mathcal{H}) + \sqrt{\frac{log\frac{1}{\delta}}{2n}}$ 
\item[--] or, $2\mathcal{R}_S(\mathcal{H}) + 3\sqrt{\frac{log\frac{2}{\delta}}{2n}}$ 
\end{itemize} &  & 
\begin{itemize}[leftmargin=*]
\item[--] Empirical Radamacher Complexity, $\mathcal{R}_S (\mathcal{H}) = \underset{\sigma}{[ \underset{h \in \mathcal{H}}{sup \frac{\sum\limits_{i=1}^n \sigma_i h(\mathbf{x}_i)}{n}}]}$ 
\item[--] Radamacher Complexity, $\mathcal{R}_n (\mathcal{H}) = \underset{S \in \mathcal{D}^n}{R_S(\mathcal{H})}$. For equivalence with, Gaussian complexity, Maximum Discrepancy etc. see \cite{bartlett2002rademacher}.
\end{itemize}
\\ 
\cline{2-5}
& Algorithm Stability \cite{mohri2012foundations} & $\beta + (2m\beta+M)\sqrt{\frac{log(\frac{1}{\delta})}{2m}}$ &  & \begin{itemize}[leftmargin=*]
\item[--] Applies to $\beta$ stable algorithms $h_S = A(S)$.
\item[--] $A$ is uniformly $\beta$ stable if $\forall (\mathbf{x},y) \in \mathcal{X} \times \{-1,+1\} \;  |l(y,h_S(\mathbf{x}))-l(y,h_S^\prime(\mathbf{x}))| \leq \beta ; S,S^\prime \sim \mathcal{D}^n$.
\end{itemize}  \\ 
\cline{2-5}
& Compression Bounds \cite{shalev2014understanding} & $\sqrt{R_{S^\prime}(A(S))\frac{4klog(m/\delta)}{m}} + \frac{8klog(m/\delta)}{m}$ & & \begin{itemize}[leftmargin=*]
\item[--] $S^\prime = S_I$ where $\exists$ an index set $I$ and a compression scheme $B: S_I \rightarrow \mathcal{H} \; s.t. \; A(S)=B(S_I)$ with $|S|>2|S_I|$.
\item[--] The bound is on the Algorithm $A$'s output in terms of error on the hold-out set $S^\prime$. 
\end{itemize} \\ 
\cline{2-5}
\hline
\end{tabular}
\normalsize
\end{sidewaystable}

Note that, Table \ref{tab::SLT} provides a very brief highlight of some of the popular traditional learning theories. For a more detailed coverage of advanced learning theories please see \cite{shalev2014understanding,mohri2012foundations}. In addition, there are a few alternative theories which uses different learning mechanisms like., mistake bounds \cite{mohri2012foundations}, learning by distance \cite{ben1990learning}, statistical query learning \cite{kearns1990computational}. Most such settings are adapted for very specific learning tasks and have some connections to the PAC learning theory. Interested readers are directed to the above references for further details.

\noindent \textbf{Computation complexity:}
Another aspect of the PAC learning theory is the computational complexity of the algorithm. Note that, there can be several algorithms to obtain the same solution. For example, to sort an array of numbers both merge sort and binary sort will provide the sorted output. Although the outcome (solution) of the algorithm may be the same; the worst-case computation complexity of each algorithm is different i.e. binary sort $O(n^2)$ and merge sort $O(nlogn)$. PAC theory captures the efficiency of an algorithm in terms of its computation complexity. However, the computation times of any algorithm is machine dependent. To decouple such dependencies most computation complexity analyses assumes an underlying abstract machine, like a Turing Machine over reals etc., and provides a comparative machine independent computational analysis in big-O notation. For example, an $O(nlogn)$ implementation of the merge sort algorithm would mean the actual runtime in seconds on any machine would follow (see \cite{shalev2014understanding}): \newline 
\textit{"there exist constants $c$ and $n_0$, which can depend on the actual machine, such that, for any value of $n > n_0$, the runtime in seconds of sorting any $n$ items
will be at most $c n log(n)$"}. \newline 
A brief coverage of the computation times for several popular algorithms in big-O notation is provided in Table \ref{table:MLcompare}. For a more in-depth discussion on several computational models readers are directed to \cite{kearns1990computational}.

As seen in this section, most popular traditional learning theories mainly target the sample complexity and in turn the generalization capability of an algorithm to learn a hypothesis class followed by the computation complexity of the algorithm to learn such a hypothesis class. From a resource constrained perspective although the computational (processing power) aspect of an algorithm is (asymptotically) handled in such theories, the interplay between computation vs. sample complexity is disjunctive. Even so, most of the existing methodologies adopted for resource constrained machine learning (discussed in sections sections \ref{sec:algo:resource_inference}, \ref{sec:algo:low_footprint}, \ref{subsec:reducemodelcomplexity}, \ref{sec:algo:dat_compress}, \ref{sec:algo:modify_opt}) follow this learning paradigm.

\subsubsection{Resource-Constrained Learning Theories} \label{sec:theory:res_const_learn_theory}

Although the PAC learning theory is the dominant theory behind majority of the ML algorithms, the notion of sample complexity and resource complexity (computation, memory, communication bandwidth, power etc.) is disjunctive in such theories. In fact, the works in \cite{decatur2000computational,ryabko2007sample} raise some critical questions about the adequacy of the PAC learning framework for designing ML algorithms with limited computation complexity. These questions have led to a substantial amount of work modifying the PAC learning paradigm to provide an explicit tradeoff between sample and computation complexity. Good surveys of sample-computation complexity theory can be found in \cite{decatur2000computational,shalev2012using,servedio2000computational,daniely2013more,daniely2013average,feldman2007efficiency,
srebro2011theoretical,lucic2017computational} and is not the main focus of this paper. Rather, in this category we target the more exclusive class of literature which additionally captures the effect of space complexity on the sample complexity for learnability. There is limited research that provides an explicit characterization of such an interplay for any generic algorithm. Most such works modify the traditional assumption of i.i.d data being presented in a batch or streaming fashion and introduces a specific protocol of data observability. Such theories limit the memory/space footprint through this restricted data observability. These advanced theories provide a platform to utilize existing computationally efficient algorithms under memory constrained settings to build machine learning models with strong error guarantees.  We discuss a few examples of such work next. 

A seminal work in this line can be found in \cite{ben1998learning}, where the authors introduce a new protocol for data observation called \textit{Restricted focus of attention} (RFA). This modified protocol is formalized through a projection (or focusing) function and limits the algorithm's memory (space) and computation footprint through selective observation of the available samples' attributes/features. The authors introduce a new notion of k-RFA or (k-weak-RFA) learnability, where k is the number of observed bits (or features) per sample; and provides a framework to analyze the sample complexity of any learning algorithm under such memory constrained observation. The interplay between space and sample complexity of any algorithm designed in this RFA framework is provided in Table \ref{tab::RLT}. Several follow-up works target a class of problems like $l_1/l_2-$linear regression, support vector regression etc., in a similar RFA framework and modified the projection function providing lower sample complexity \cite{hazan2012linear,bullins2016limits,ito2017efficient,ito2018online}. They also provide computationally efficient algorithms for these specific methods. 

Another approach to characterize the sample vs. space complexity is provided in \cite{steinhardt2016memory}. Here the authors introduce a memory efficient streaming data observation protocol and utilizes the statistical query (SQ) based learning paradigm (originally introduced in \cite{kearns1990computational}) to characterize the sample vs. space complexity for learnability of finite hypothesis classes. Allowing the SQ algorithm for improper learning, the authors justify applicability to infinite function classes through an $\epsilon$ - cover under some metric. However, such extensions have not been provided in the paper. One major advantage of this approach is that it opens up the gamut of machine learning algorithms developed in the SQ paradigm for resource constrained settings \cite{feldman2013statistical,feldman2017statistical}. As an example, the authors illustrate the applicability of their theory for designing resource efficient k-sparse linear regression algorithm following \cite{feldman2013statistical,feldman2017statistical}. Additional details on the space/sample complexity under this framework is provided in Table \ref{tab::RLT}.

More recently \cite{moshkovitz2017mixing,moshkovitz2018entropy} introduced a graph based approach to model the version space of a hypothesis class in the form of a \textit{hypothesis graph}. The authors introduced a notion of \textit{d-mixing} of the hypothesis graph as a measure of (un)-learnability in bounded memory settings. This notion of mixing was  formalized as a complexity measure in \cite{moshkovitz2017mixing} and further utilized to show the un-learnability property of most generic neural network architectures. In another line of work a similar (yet complimentary) notion of \textit{separability} was introduced in \cite{moshkovitz2017general}. Rather than showing the negative (unlearnability) results under bounded memory settings; this framework was used to characterize the lower bound sample vs. space complexity for learnability of a hypothesis class using an $(n, b, \delta, \epsilon)$ - bounded memory algorithm. Here, the data generating distribution is assumed to be uniform in the domain, $n$ = number of training samples, $b$ = bits of memory, $\delta, \epsilon$ = confidence, tolerance values for PAC learnability. However, the framework supports a very limited set of machine learning algorithms and its applicability for modifying popular machine learning algorithms have not yet been shown. 

Finally, \cite{raz2017time} proposes a new protocol for data access called \textit{branching program} and translates the learning algorithm under resource (memory) constraints in the form of a matrix (as opposed to a graph in \cite{moshkovitz2017mixing}). The authors build a connection between the stability of the matrix norm (in the form of an upper bound on its maximum singular value) and the learnability of the hypothesis class with limited memory. This work provides the interplay between the memory-size and minimum number of samples required for exact learning of the concept class. \cite{beame2017time} extended this work for the class of problems where the sample space of tests is smaller than the sample space of inputs. \cite{garg2018extractor} further improves upon the bounds proposed in \cite{raz2017time, beame2017time}. However, most existing work in this framework focus on analyzing the learnability of a problem, rather than providing an algorithm guaranteeing learnability for resource constrained settings.

In addition to the above theories there are a few research works that target a niche yet interesting class of machine learning problems like hypothesis testing \cite{lakshmanan1979compound,hellman1971memory,hellman1972effects,leighton1986estimating,drakopoulos2013learning} or function estimation \cite{alon1999space,morris1978counting,henzinger1998computing}. However, their extension to any generic loss function has not been provided and is non-trivial. Hence, we do not delve into the details of such settings. However, one specific research on estimating biased coordinates \cite{shamir2014fundamental} needs special consideration, mainly because it covers a set of very interesting problems like principal component analysis, singular value decomposition, correlation analysis etc. This work adopts an information theory centric general framework to handle most memory constrained estimation problems. The authors introduce a generic $(b,n_t,T)$ protocol of data observability for most iterative mini-batch based algorithms. Here, $b$ = space complexity of intermediate results, $n_t$ = size of a mini-batch (of i.i.d samples with dimension $d$) at $t \in T$ iteration, where $T$ = number of epochs for the iterative algorithm. In a loose sense the overall space complexity including the data and intermediate results become $O(T(b+n_td))$. Under such a $(b,n_t,T)$ protocol the authors casts most estimation problems like sparse PCA, Covariance estimation, correlation analysis etc., into a `\textit{hide-and-seek}' problem. Using this framework the authors explore the limitations in memory constrained settings and provide the sample complexity for good estimation with high probability (see Table \ref{tab::RLT}). However, although this provides a framework to analyze specific estimation problems, there are no guidelines on how to cast/modify existing algorithms to optimize the space vs. sample complexities within this framework.



\begin{sidewaystable}[htbp]
\caption{Resource Constrained Learning Theories} 
\label{tab::RLT}
\scriptsize
\begin{tabular}{|c|c|p{0.2\textwidth}|p{0.35\textwidth}|}  
\toprule
\textbf{Theory} & \textbf{Sample Complexity} & \textbf{Space Complexity} & \quad \quad \textbf{Additional  Remarks} \\
\noalign{
\hrule height 2pt
}
\specialcell{$k-$RFA  \\ ( \cite{ben1998learning} \sc{Corollary 4.3})} & $max \{ \frac{4r^2}{\epsilon} log \frac{2r}{\delta}, \frac{8r^2 VCdim(\mathcal{H})}{\epsilon} log \frac{13r}{\epsilon} \}$ & $O(k)$ per sample & 
\begin{itemize}[leftmargin=*]
\item[--] $\mathcal{H} = \Psi(\mathcal{F}_i)_{i=1}^r$, where Composition function $\Psi(\mathcal{F}_i)_{i=1}^r = \{ \psi(f_1,\ldots,f_r)| f_i \in \mathcal{F} \}$ defines an ensemble of functions.
\item[--] $\mathcal{F}$ is PAC learnable over domain $X_k$. $k<d$ is a pre-selected subset of features.   
\item[--] $r$ number of functions used in the ensemble function $\Psi$. This is user-defined.
\item[--] Captures the per-sample space complexity and not the overall algorithm space complexity.
\end{itemize}
\\
\hline
\specialcell{Statistical Query \\(\cite{steinhardt2016memory} \sc{Theorem} 7)} & $O(\frac{\lceil m_0/k log |\mathcal{H}|}{\tau^2}(log log |\mathcal{H}| + log m_0 + log(1/\delta)))$ & $O(log |\mathcal{H}|(log m_0 + log log (1/\tau)) + k log(1/\tau))$ per state variable& 
\begin{itemize}[leftmargin=*]
\item[--] Assumption : $\mathcal{H}$ is $(\epsilon, 0)-$ learnable with $m_0$ statistical queries of tolerance $\tau$.
\item[--] $k$ = user defined trade-off with inverse-dependence on sample complexity and direct dependence on space complexity. 
\item[--] $|\mathcal{H}|$ = size of the finite hypothesis class of probability distributions. 
\item[--] Following this framework provides a k-sparse linear regression implementation with sample complexity $\tilde{O}(\frac{nk^8 log(1/\delta)}{\epsilon^4})$ and space complexity $O(k log^2(\frac{d}{\epsilon}))$ bits. $\tilde{O}(\cdot)$ is big - O which additionally hides the log terms. 
\end{itemize}
\\
\hline
\specialcell{$\mathcal{H}-$ graph separability \\(\cite{moshkovitz2017general} \sc{Theorem} 7)} & $\frac{k}{\alpha} log \frac{|\mathcal{H}|}{\alpha^2}$ & $log \frac{|\mathcal{H}|}{\alpha^2} + log \frac{k}{\alpha}$ bits &  
\begin{itemize}[leftmargin=*]
\item[--] Assumption : $\mathcal{H}-$ graph is $(\alpha, \epsilon)-$ separable (see \cite{steinhardt2016memory} for definition).
\item[--] $\mathcal{H}$ is PAC learnable for given ($\epsilon , \delta = e^{-k\alpha^2/8} + e^{-2k \epsilon^2}$).
\item[--] Limited to uniform distributions and targets sample complexity lower bounds.
\end{itemize}
\\
\hline
\specialcell{Hide-n-Seek \\ (\cite{shamir2014fundamental} \sc{Theorem} 3)} & $\Omega(\text{max}\{(d/\rho b), 1/\rho^2\})$ & $O(T(b+n_td))$ & 
\begin{itemize}[leftmargin=*]
\item[--] Applicable only to estimation problems like PCA, SVD, CCA etc., falling in the hide-n-seek framework.
\item[--] $b$ = space complexity of intermediate results. 
\item[--] $n_t$ = size of a mini-batch (of i.i.d samples with dimension $d$) at $t \in T$ iteration. 
\item[--] $T$ = number of epochs for the iterative algorithm.
\end{itemize}
\\
\hline
\end{tabular}
\normalsize
\end{sidewaystable}

\subsection{Challenges in Resource-Efficient Theoretical Research} \label{sec:theory:challenges}
As presented in the above section there are some major challenges underlying the resource efficient theories discussed in section \ref{sec:theory:res_const_learn_theory}.
\begin{enumerate}
\item Most of the new theories are mainly developed to analyze the un-learnability of a hypothesis class $\mathcal{H}$ under resource constraints. Adapting such frameworks to develop resource efficient algorithms guaranteeing the learnability of $\mathcal{H}$ is needed. Although there are a few theoretical frameworks like \cite{ben1998learning,steinhardt2016memory} which provide guidelines towards developing resource efficient algorithms. However, the underlying assumptions rule out a wide range of hypothesis classes. Showing the practicality of such assumptions towards developing a wide range of machine learning algorithms is a huge challenge in such existing frameworks.  
\item In addition, as shown in Table \ref{tab::RLT} most of the existing theories deal with a specific aspect of the space-complexity. For example, \cite{ben1998learning} mainly considers the per-sample space complexity while \cite{steinhardt2016memory} considers the space complexity of the intermediate state representation. A more comprehensive analysis of the overall space/computation complexity of the algorithm $A(S)$ needs to be developed.
\item A general limitation of most of the theoretical frameworks is the limited empirical analysis of the algorithms designed using these frameworks. For example, most of the analyses are asymptotic in nature. How such mathematical expressions translate in practicality is still an open problem.
\item Most of the existing theories introduces error guarantees in terms of the hypothesis class. Selecting the optimal model through hyperparameter optimization, model selection routines in a resource constrained setting and guaranteeing the correctness of the selected model is missing.  
\item Finally, a comparison between the frameworks introduced in Table \ref{tab::RLT} is missing. 
\end{enumerate}

\section{Discussion}
\label{sec:challenges}
The previous sections provide a comprehensive look at the current state of on-device learning from an algorithm and theory perspective. In this section we provide a brief summary of these findings and elaborate on the research and development challenges facing the adoption of an edge learning paradigm. We also highlight the effort needed for on-device learning using a few typical edge-learning use cases and explain how research in the different areas (algorithms and theory) has certain advantages and disadvantages when it comes to their usability. 

\subsection{Summary of the Current State-of-the-art in On-device Learning}
We begin by briefly summarizing our findings from Sections 3 and 4. If you have read those sections, you can skip this summary and move directly to \autoref{subsec:R&D Challenges}.

\noindent \textbf{Algorithm Research (details in \autoref{sec:algo}):} Algorithm research mainly targets the computational aspects of model building under limited resource settings. The main goal is to design optimized machine learning algorithms which best satisfies a surrogate software-centric resource constraint. Such surrogate measures are designed to approximate the hardware constraints through asymptotic analysis (\autoref{subsec:asymptotic}), resource profiling (\autoref{subsec:res_prof}) or resource modeling (\autoref{subsec:resourcemodeling}). For a given software-centric resource constraint, the state-of-art algorithm designs adopt one of the following approaches:
\begin{enumerate}
    \item Lightweight ML Algorithms (see \autoref{sec:algo:low_footprint}): This approach utilizes already available algorithms with low resource footprints. There are no additional modifications for resource constrained model building. As such, for cases where the available device's resources are smaller than the resource footprint of the selected lightweight algorithm, this approach will fail. Additionally, in most cases the lightweight ML algorithms result in models with low complexity that may fail to fully capture the underlying process. 

    \item Reducing Model complexity (see \autoref{subsec:reducemodelcomplexity}): This approach controls the size (memory footprint) and computation complexity of the machine learning algorithm by adding additional constraints on to the model architecture (e.g. by selecting a smaller hypothesis class). For a pre-specified constrained model architecture motivated by the available resource constraints, this approach adopts traditional optimization routines. Apart from model building, this is one of the dominant approaches for deploying resource efficient models for model inference. Compared to the lightweight ML algorithms approach, model complexity reduction techniques can accommodate a broader class of ML algorithms and can more effectively capture the underlying process.  
    
    \item Modifying optimization routines (see \autoref{sec:algo:modify_opt}): This approach designs specific optimization routines for resource efficient model building. Here the resource constraints are incorporated during the model building (training) phase. Note that as opposed to the previous technique of limiting the model architectures beforehand, this approach can adapt the optimization routines to fit the resource constraints for any given model architecture (hypothesis class). In certain cases, this approach can also dynamically modify the architecture to fit the resource constraints. Although this approach provides a wider choice of the class of models, the design process is still tied to a specific problem type (classification, regression, etc.) and adopted method/loss function (linear regression, ridge regression for regression problems).
    
    \item Data Compression (see \autoref{sec:algo:dat_compress}):  Rather than constraining the model size/complexity, this approach targets building models on compressed data. The goal is to limit the memory usage via reduced data storage and computation through fixed per-sample computation cost. In addition a more generic approach includes adopting advanced learning settings that accommodates algorithms with smaller sample complexity. However, this is a broader research topic and is not just limited to on-device learning. A detailed analysis of these approaches have been delegated to other existing surveys.
    
    \color{black}
    
    \item New protocols for data observation (see \autoref{sec:algo:new_protocols}): This approach completely changes the traditional data observation protocol (like availability of i.i.d data in batch or online settings), and builds resource efficient models under limited data observation. These approaches are guided by an underlying resource constrained learning theory (discussed in \autoref{sec:theory:res_const_learn_theory}) which captures the interplay between resource constraints and the goodness of the model in terms of the generalization capacity. Additionally, compared to the above approaches, this framework provides a generic mechanism to design resource constrained algorithms for a wider range of learning problems applicable to any method/loss function targeting that problem type.   

\end{enumerate}
Obviously one of the major challenges in this research is proper software-centric characterization of the hardware constraints and appropriately using this characterization for better metric designs. Some other important challenges include applicability to a wider range of algorithms and dynamic resource budgeting. A more detailed discussion is available in \autoref{subsec:algchallenges}.

\noindent \textbf{Theory Research (details in \autoref{sec:theory}):} Research into the theory underlying on-device learning focuses mainly on developing frameworks to analyze the statistical aspects (i.e. error guarantees) of a designed algorithm with or without associated resource constraints. There are two broad categories into which most of the existing resource constrained algorithms can be categorized,
\begin{enumerate}
    \item Traditional Learning Theories (see \autoref{sec:theory:trad_learn_theory}): Most of the existing resource constrained algorithm design (summarized in $1 - 4$ above) follow this traditional machine learning theory. A limitation of this approach is that such theories are built mainly for analyzing the error guarantees of the algorithm used for model estimation. The effect of resource constraints on the generalization capability of the algorithm is not directly addressed through such theories. For example, algorithms developed using the approach of \textit{reducing the model complexity} typically adopts a two step approach. First the size of the hypothesis class is constrained \textit{a priori} to incorporate the resource constraints. Next, an algorithm is designed guaranteeing the best-in-class model within that hypothesis class. The direct interplay between the error guarantees and resource constraints is missing in such theoretical frameworks.  
    
    \item Resource constrained learning theories (see \autoref{sec:theory:res_const_learn_theory}): Modern research has shown that it may be impossible to learn a hypothesis class under resource constrained settings. To circumvent such inconsistencies in traditional learning theories, newer resource constraint learning theories have been developed. Such theories provide learning guarantees in light of the resource constraints imposed by the device. The algorithms designed using the approach summarized in point 5 above follow these newer learning theories. Although such theory motivated design provides a generic framework through which algorithms can be designed for a wide range of learning problems for any loss function addressing the problem type, till date, very few algorithms based on these theories have been developed.
\end{enumerate}
Overall, the newer resource constrained theory research provides a generic framework for designing algorithms with error guarantees under resource constrained settings which apply to a broader range of problems. However, currently, very few algorithms have been developed which utilize these frameworks. Developing additional algorithms in such advanced frameworks need significant effort. Also, the application of such theories to a complete ML pipeline including hyperparameter optimization, data preprocessing etc., has not yet been addressed.

\subsection{Research \& Development Challenges}
\label{subsec:R&D Challenges}
A study of the current state-of-the-art in on-device learning also provides an understanding of the existing challenges in the field that prevent its adoption as a real alternative to cloud-based solutions. In this section we identify a few directions for research that will allow us to build learning algorithms that run on the edge. As before, we limit our discussion to the algorithm and theory levels. 



\noindent \textbf{Algorithm:} The challenges facing the research \& development of resource constrained optimized algorithms falls into three areas:
\begin{enumerate}
    \item Software-centric Resource Constraints: A correct characterization of the software-centric resource constraints that best approximates the hardware resources is absolutely critical to developing resource constrained algorithms because the optimal algorithm design is very specific to a particular device (i.e. computational model) and its available resources. 
    
    
    \item Understanding how traditional ML algorithms can contribute to edge learning: Majority of current on-device algorithm research focuses on adapting modern deep learning approaches like CNN, DNN to run on the edge. Very little focus has been given to traditional machine learning methods. Traditional machine learning methods are of huge interest for building edge learning capability especially when memory is limited (for devices with megabytes or even kilobytes of RAM) and computational power is low. These are precisely the areas where modern deep architectures are wholly unsuited due to their out-sized memory and compute requirements. Hence, there is a need to explore the traditional machine learning approaches for learning on-device. In addition, designing advanced learning algorithms that requires small training datasets to achieve high test accuracies is a huge challenge. Addressing this capability can significantly improve the resource footprint (i.e low memory footprint to store the training data).
    
    \item Dynamic resource budget: Most existing research assume the availability of dedicated resources while training their edge-learning algorithms. However, for many applications such an assumption is invalid. For example, in mobile phones an application's priority may change \cite{fang2018nestdnn} based on user's actions. Hence, there is a need for algorithm research incorporating the accuracy-latency trade-off under dynamic resource constraints. 
\end{enumerate}

For building edge-learning capability for use-cases like object detection, user identification etc. specific to products like., mobile phones, refrigerator etc. ; significant groundwork needs to be laid in terms of a complete analysis of existing approaches that cater to the specific requirements of the use-cases and the products. This analysis also forms the basis for extending existing approaches and developing newer algorithms in those cases where current methods fail to perform satisfactorily. 

\noindent \textbf{Theory:} The research on algorithms developed using (extensions of) advanced theoretical frameworks provides a mechanism to characterize the performance (error) guarantees of a model in resource constrained settings. The effort on such research needs careful consideration of the following points, 
\begin{enumerate}
    \item Analysis of Resource Constraints: Although advanced theories abstract resource constraints as a form of information bottleneck, a good understanding of how such abstractions practically impact a resource (say memory vs. computation) needs to be properly analyzed.
    \item Applicability: With the appropriate abstraction in place, applicability (of the underlying assumptions) of the theory for practical use-cases needs to be analyzed.
    \item Algorithm development: Once newer learning theories supporting resource constraints have been developed, there is need to develop new algorithms based on these new theoretical frameworks.
    \item Extending existing theories: The existing theories majorly cater to a very specific aspect of the ML pipeline (i.e. training an algorithm for a pre-specified hypothesis class). Extending this approach to the entire model building pipeline (for e.g., Hyper-parameter optimization, data pre-processing, etc.) remains to be developed. Also, a majority of the underlying theories target the un-learnability of a problem. Extending such concepts to newer theories to develop practical algorithms needs to be developed.
\end{enumerate}
Overall, the theory research provides a mechanism to characterize a wide range of edge-specific ML problems. However, extending research at this level needs significant effort as compared to the previous algorithmic approaches.

\section{Conclusion} 
On-device learning has so far remained in the purview of academic research, but with the increasing number of smart devices and improved hardware, there is interest in performing learning on the device rather than in the cloud. In the industry, this interest is fueled mainly by hardware manufacturers promoting AI-specific chipsets that are optimized for certain mathematical operations, and startups providing ad hoc solutions to certain niche domains mostly in computer vision and IoT. Given this surge in interest and corresponding availability of edge hardware suitable for on-device learning, a comprehensive survey of the field from an algorithms and learning theory perspective sets the stage for both understanding the state-of-the-art and for identifying open challenges and future avenues of research. However, on-device learning is an expansive field with connections to a large number of related topics in AI and machine learning including online learning, model adaptation, one/few-shot learning, resource-constrained learning etc. to name just a few. Covering such a large number of research topics in a single survey is impractical but, at the same time, ignoring the work that has been done in these areas leaves significant gaps in any comparison of approaches. This survey finds a middle ground by reformulating the problem of on-device learning as resource-constrained learning where the resources are compute and memory. This reformulation allows tools, techniques, and algorithms from a wide variety of research areas to be compared equitably.  

We limited the survey to learning on single devices with the understanding that the ideas discussed can be extended in a normal fashion to the distributed setting via an additional constraint based on communication latency. We also focused the survey on the algorithmic and theoretical aspects of on-device learning leaving out the effects of the systems level (hardware and libraries). This choice was deliberate and allowed us to separate out the algorithmic aspects of on-device learning from implementation and hardware choices. This distinction also allows us to identify challenges and future research that can be applied to a variety of systems. 

Based on the reformulation of on-device learning as resource constrained learning, the survey found that there are a number of areas where more research and development is needed.  

At the algorithmic level, it is clear that current efforts are mainly targeted at either utilizing already lightweight machine learning algorithms or modifying existing algorithms in ways that reduce resource utilization. There are a number of challenges we identified in the algorithm space including the need for decoupling algorithms from hardware constraints, designing effective loss functions and metrics that capture resource constraints, an expanded focus on traditional as well as advanced ML algorithms with low sample complexity in addition to the current work on DNNs, and dealing with situations where the resource budget is dynamic rather than static. In addition, improved methods for model profiling are needed to more accurately calculate an algorithm's resource consumption. Current approaches to such measurements are abstract and focus on applying software engineering principles such as asymptotic analysis or low-level measures like FLOPS or MACs (Multiply-Add Computations). None of these approaches give a holistic idea of resource requirements and in many cases represent an insignificant portion of the total resources required by the system during learning.

Finally, current research in the field of learning theory for resource constrained algorithms is focused on the un-learnability of an algorithm under resource constraints. The natural step forward is to identify techniques that can instead provide guarantees on the learnability of an algorithm and the associated estimation error. Existing theoretical techniques also mainly focus on the space(memory) complexity of these algorithms and not their compute requirements. Even in cases where an ideal hypothesis class can be identified that satisfies resource constraints, further work is needed to select the optimal model from within that class.

\small
\bibliographystyle{ACM-Reference-Format}  
\bibliography{OnDevice}


\begin{thebibliography}{214}


\ifx \showCODEN    \undefined \def \showCODEN     #1{\unskip}     \fi
\ifx \showDOI      \undefined \def \showDOI       #1{#1}\fi
\ifx \showISBNx    \undefined \def \showISBNx     #1{\unskip}     \fi
\ifx \showISBNxiii \undefined \def \showISBNxiii  #1{\unskip}     \fi
\ifx \showISSN     \undefined \def \showISSN      #1{\unskip}     \fi
\ifx \showLCCN     \undefined \def \showLCCN      #1{\unskip}     \fi
\ifx \shownote     \undefined \def \shownote      #1{#1}          \fi
\ifx \showarticletitle \undefined \def \showarticletitle #1{#1}   \fi
\ifx \showURL      \undefined \def \showURL       {\relax}        \fi
\providecommand\bibfield[2]{#2}
\providecommand\bibinfo[2]{#2}
\providecommand\natexlab[1]{#1}
\providecommand\showeprint[2][]{arXiv:#2}

\bibitem[\protect\citeauthoryear{Adolf, Rama, Reagen, Wei, and Brooks}{Adolf
  et~al\mbox{.}}{2016}]%
        {adolf2016fathom}
\bibfield{author}{\bibinfo{person}{Robert Adolf}, \bibinfo{person}{Saketh
  Rama}, \bibinfo{person}{Brandon Reagen}, \bibinfo{person}{Gu-Yeon Wei}, {and}
  \bibinfo{person}{David Brooks}.} \bibinfo{year}{2016}\natexlab{}.
\newblock \showarticletitle{Fathom: Reference workloads for modern deep
  learning methods}. In \bibinfo{booktitle}{\emph{Workload Characterization
  (IISWC), 2016 IEEE International Symposium on}}. IEEE,
  \bibinfo{pages}{1--10}.
\newblock


\bibitem[\protect\citeauthoryear{Alam, Mehmood, Katib, and Albeshri}{Alam
  et~al\mbox{.}}{2016}]%
        {alam2016analysis}
\bibfield{author}{\bibinfo{person}{Furqan Alam}, \bibinfo{person}{Rashid
  Mehmood}, \bibinfo{person}{Iyad Katib}, {and} \bibinfo{person}{Aiiad
  Albeshri}.} \bibinfo{year}{2016}\natexlab{}.
\newblock \showarticletitle{Analysis of eight data mining algorithms for
  smarter Internet of Things (IoT)}.
\newblock \bibinfo{journal}{\emph{Procedia Computer Science}}
  \bibinfo{volume}{98} (\bibinfo{year}{2016}), \bibinfo{pages}{437--442}.
\newblock


\bibitem[\protect\citeauthoryear{Alistarh, Grubic, Li, Tomioka, and
  Vojnovic}{Alistarh et~al\mbox{.}}{2017}]%
        {alistarh2017qsgd}
\bibfield{author}{\bibinfo{person}{Dan Alistarh}, \bibinfo{person}{Demjan
  Grubic}, \bibinfo{person}{Jerry Li}, \bibinfo{person}{Ryota Tomioka}, {and}
  \bibinfo{person}{Milan Vojnovic}.} \bibinfo{year}{2017}\natexlab{}.
\newblock \showarticletitle{QSGD: Communication-efficient SGD via gradient
  quantization and encoding}. In \bibinfo{booktitle}{\emph{Advances in Neural
  Information Processing Systems}}. \bibinfo{pages}{1709--1720}.
\newblock


\bibitem[\protect\citeauthoryear{Allen-Zhu and Li}{Allen-Zhu and Li}{2017}]%
        {allen2017first}
\bibfield{author}{\bibinfo{person}{Zeyuan Allen-Zhu} {and}
  \bibinfo{person}{Yuanzhi Li}.} \bibinfo{year}{2017}\natexlab{}.
\newblock \showarticletitle{First efficient convergence for streaming k-pca: a
  global, gap-free, and near-optimal rate}. In
  \bibinfo{booktitle}{\emph{Foundations of Computer Science (FOCS), 2017 IEEE
  58th Annual Symposium on}}. IEEE, \bibinfo{pages}{487--492}.
\newblock


\bibitem[\protect\citeauthoryear{Alon, Matias, and Szegedy}{Alon
  et~al\mbox{.}}{1999}]%
        {alon1999space}
\bibfield{author}{\bibinfo{person}{Noga Alon}, \bibinfo{person}{Yossi Matias},
  {and} \bibinfo{person}{Mario Szegedy}.} \bibinfo{year}{1999}\natexlab{}.
\newblock \showarticletitle{The space complexity of approximating the frequency
  moments}.
\newblock \bibinfo{journal}{\emph{Journal of Computer and system sciences}}
  \bibinfo{volume}{58}, \bibinfo{number}{1} (\bibinfo{year}{1999}),
  \bibinfo{pages}{137--147}.
\newblock


\bibitem[\protect\citeauthoryear{Bartlett and Mendelson}{Bartlett and
  Mendelson}{2002}]%
        {bartlett2002rademacher}
\bibfield{author}{\bibinfo{person}{Peter~L Bartlett} {and}
  \bibinfo{person}{Shahar Mendelson}.} \bibinfo{year}{2002}\natexlab{}.
\newblock \showarticletitle{Rademacher and Gaussian complexities: Risk bounds
  and structural results}.
\newblock \bibinfo{journal}{\emph{Journal of Machine Learning Research}}
  \bibinfo{volume}{3}, \bibinfo{number}{Nov} (\bibinfo{year}{2002}),
  \bibinfo{pages}{463--482}.
\newblock


\bibitem[\protect\citeauthoryear{Beame, Gharan, and Yang}{Beame
  et~al\mbox{.}}{2017}]%
        {beame2017time}
\bibfield{author}{\bibinfo{person}{Paul Beame}, \bibinfo{person}{Shayan~Oveis
  Gharan}, {and} \bibinfo{person}{Xin Yang}.} \bibinfo{year}{2017}\natexlab{}.
\newblock \showarticletitle{Time-Space Tradeoffs for Learning from Small Test
  Spaces: Learning Low Degree Polynomial Functions}.
\newblock \bibinfo{journal}{\emph{arXiv preprint arXiv:1708.02640}}
  (\bibinfo{year}{2017}).
\newblock


\bibitem[\protect\citeauthoryear{Ben-David and Dichterman}{Ben-David and
  Dichterman}{1998}]%
        {ben1998learning}
\bibfield{author}{\bibinfo{person}{Shai Ben-David} {and} \bibinfo{person}{Eli
  Dichterman}.} \bibinfo{year}{1998}\natexlab{}.
\newblock \showarticletitle{Learning with restricted focus of attention}.
\newblock \bibinfo{journal}{\emph{J. Comput. System Sci.}}
  \bibinfo{volume}{56}, \bibinfo{number}{3} (\bibinfo{year}{1998}),
  \bibinfo{pages}{277--298}.
\newblock


\bibitem[\protect\citeauthoryear{Ben-David, Itai, and Kushilevitz}{Ben-David
  et~al\mbox{.}}{1990}]%
        {ben1990learning}
\bibfield{author}{\bibinfo{person}{Shai Ben-David}, \bibinfo{person}{Alon
  Itai}, {and} \bibinfo{person}{Eyal Kushilevitz}.}
  \bibinfo{year}{1990}\natexlab{}.
\newblock \showarticletitle{Learning by Distances.}. In
  \bibinfo{booktitle}{\emph{COLT}}. \bibinfo{pages}{232--245}.
\newblock


\bibitem[\protect\citeauthoryear{Bianco, Cadene, Celona, and Napoletano}{Bianco
  et~al\mbox{.}}{2018}]%
        {bianco2018benchmark}
\bibfield{author}{\bibinfo{person}{Simone Bianco}, \bibinfo{person}{Remi
  Cadene}, \bibinfo{person}{Luigi Celona}, {and} \bibinfo{person}{Paolo
  Napoletano}.} \bibinfo{year}{2018}\natexlab{}.
\newblock \showarticletitle{Benchmark Analysis of Representative Deep Neural
  Network Architectures}.
\newblock \bibinfo{journal}{\emph{IEEE Access}} (\bibinfo{year}{2018}).
\newblock


\bibitem[\protect\citeauthoryear{Bolukbasi, Wang, Dekel, and
  Saligrama}{Bolukbasi et~al\mbox{.}}{2017}]%
        {bolukbasi2017adaptive}
\bibfield{author}{\bibinfo{person}{Tolga Bolukbasi}, \bibinfo{person}{Joseph
  Wang}, \bibinfo{person}{Ofer Dekel}, {and} \bibinfo{person}{Venkatesh
  Saligrama}.} \bibinfo{year}{2017}\natexlab{}.
\newblock \showarticletitle{Adaptive neural networks for efficient inference}.
\newblock \bibinfo{journal}{\emph{arXiv preprint arXiv:1702.07811}}
  (\bibinfo{year}{2017}).
\newblock


\bibitem[\protect\citeauthoryear{Bullins, Hazan, and Koren}{Bullins
  et~al\mbox{.}}{2016}]%
        {bullins2016limits}
\bibfield{author}{\bibinfo{person}{Brian Bullins}, \bibinfo{person}{Elad
  Hazan}, {and} \bibinfo{person}{Tomer Koren}.}
  \bibinfo{year}{2016}\natexlab{}.
\newblock \showarticletitle{The limits of learning with missing data}. In
  \bibinfo{booktitle}{\emph{Advances in Neural Information Processing
  Systems}}. \bibinfo{pages}{3495--3503}.
\newblock


\bibitem[\protect\citeauthoryear{Cai, Juan, Stamoulis, and Marculescu}{Cai
  et~al\mbox{.}}{2017}]%
        {cai2017neuralpower}
\bibfield{author}{\bibinfo{person}{Ermao Cai}, \bibinfo{person}{Da-Cheng Juan},
  \bibinfo{person}{Dimitrios Stamoulis}, {and} \bibinfo{person}{Diana
  Marculescu}.} \bibinfo{year}{2017}\natexlab{}.
\newblock \showarticletitle{Neuralpower: Predict and deploy energy-efficient
  convolutional neural networks}.
\newblock \bibinfo{journal}{\emph{arXiv preprint arXiv:1710.05420}}
  (\bibinfo{year}{2017}).
\newblock


\bibitem[\protect\citeauthoryear{Cai, Zhu, and Han}{Cai et~al\mbox{.}}{2018}]%
        {cai2018proxylessnas}
\bibfield{author}{\bibinfo{person}{Han Cai}, \bibinfo{person}{Ligeng Zhu},
  {and} \bibinfo{person}{Song Han}.} \bibinfo{year}{2018}\natexlab{}.
\newblock \showarticletitle{Proxylessnas: Direct neural architecture search on
  target task and hardware}.
\newblock \bibinfo{journal}{\emph{arXiv preprint arXiv:1812.00332}}
  (\bibinfo{year}{2018}).
\newblock


\bibitem[\protect\citeauthoryear{Cambier, Bhiwandiwalla, Gong, Nekuii, Elibol,
  and Tang}{Cambier et~al\mbox{.}}{2020}]%
        {cambier2020shifted}
\bibfield{author}{\bibinfo{person}{L{\'e}opold Cambier},
  \bibinfo{person}{Anahita Bhiwandiwalla}, \bibinfo{person}{Ting Gong},
  \bibinfo{person}{Mehran Nekuii}, \bibinfo{person}{Oguz~H Elibol}, {and}
  \bibinfo{person}{Hanlin Tang}.} \bibinfo{year}{2020}\natexlab{}.
\newblock \showarticletitle{Shifted and Squeezed 8-bit Floating Point format
  for Low-Precision Training of Deep Neural Networks}.
\newblock \bibinfo{journal}{\emph{arXiv preprint arXiv:2001.05674}}
  (\bibinfo{year}{2020}).
\newblock


\bibitem[\protect\citeauthoryear{Canziani, Paszke, and Culurciello}{Canziani
  et~al\mbox{.}}{2016}]%
        {canziani2016analysis}
\bibfield{author}{\bibinfo{person}{Alfredo Canziani}, \bibinfo{person}{Adam
  Paszke}, {and} \bibinfo{person}{Eugenio Culurciello}.}
  \bibinfo{year}{2016}\natexlab{}.
\newblock \showarticletitle{An analysis of deep neural network models for
  practical applications}.
\newblock \bibinfo{journal}{\emph{arXiv preprint arXiv:1605.07678}}
  (\bibinfo{year}{2016}).
\newblock


\bibitem[\protect\citeauthoryear{Cesa-Bianchi, Shalev-Shwartz, and
  Shamir}{Cesa-Bianchi et~al\mbox{.}}{2011}]%
        {cesa2011efficient}
\bibfield{author}{\bibinfo{person}{Nicolo Cesa-Bianchi}, \bibinfo{person}{Shai
  Shalev-Shwartz}, {and} \bibinfo{person}{Ohad Shamir}.}
  \bibinfo{year}{2011}\natexlab{}.
\newblock \showarticletitle{Efficient learning with partially observed
  attributes}.
\newblock \bibinfo{journal}{\emph{Journal of Machine Learning Research}}
  \bibinfo{volume}{12}, \bibinfo{number}{Oct} (\bibinfo{year}{2011}),
  \bibinfo{pages}{2857--2878}.
\newblock


\bibitem[\protect\citeauthoryear{Chapelle, Scholkopf, and Zien}{Chapelle
  et~al\mbox{.}}{2009}]%
        {chapelle2009semi}
\bibfield{author}{\bibinfo{person}{Olivier Chapelle}, \bibinfo{person}{Bernhard
  Scholkopf}, {and} \bibinfo{person}{Alexander Zien}.}
  \bibinfo{year}{2009}\natexlab{}.
\newblock \showarticletitle{Semi-supervised learning (chapelle, o. et al.,
  eds.; 2006)[book reviews]}.
\newblock \bibinfo{journal}{\emph{IEEE Transactions on Neural Networks}}
  \bibinfo{volume}{20}, \bibinfo{number}{3} (\bibinfo{year}{2009}),
  \bibinfo{pages}{542--542}.
\newblock


\bibitem[\protect\citeauthoryear{Chen, Xu, Zhang, and Guestrin}{Chen
  et~al\mbox{.}}{2016b}]%
        {chen2016training}
\bibfield{author}{\bibinfo{person}{Tianqi Chen}, \bibinfo{person}{Bing Xu},
  \bibinfo{person}{Chiyuan Zhang}, {and} \bibinfo{person}{Carlos Guestrin}.}
  \bibinfo{year}{2016}\natexlab{b}.
\newblock \showarticletitle{Training deep nets with sublinear memory cost}.
\newblock \bibinfo{journal}{\emph{arXiv preprint arXiv:1604.06174}}
  (\bibinfo{year}{2016}).
\newblock


\bibitem[\protect\citeauthoryear{Chen, Wilson, Tyree, Weinberger, and
  Chen}{Chen et~al\mbox{.}}{2015}]%
        {chen2015compressing}
\bibfield{author}{\bibinfo{person}{Wenlin Chen}, \bibinfo{person}{James
  Wilson}, \bibinfo{person}{Stephen Tyree}, \bibinfo{person}{Kilian
  Weinberger}, {and} \bibinfo{person}{Yixin Chen}.}
  \bibinfo{year}{2015}\natexlab{}.
\newblock \showarticletitle{Compressing neural networks with the hashing
  trick}. In \bibinfo{booktitle}{\emph{International Conference on Machine
  Learning}}. \bibinfo{pages}{2285--2294}.
\newblock


\bibitem[\protect\citeauthoryear{Chen, Liu, Wang, Gales, and Woodland}{Chen
  et~al\mbox{.}}{2016a}]%
        {chen2016efficient}
\bibfield{author}{\bibinfo{person}{Xie Chen}, \bibinfo{person}{Xunying Liu},
  \bibinfo{person}{Yongqiang Wang}, \bibinfo{person}{Mark~JF Gales}, {and}
  \bibinfo{person}{Philip~C Woodland}.} \bibinfo{year}{2016}\natexlab{a}.
\newblock \showarticletitle{Efficient training and evaluation of recurrent
  neural network language models for automatic speech recognition}.
\newblock \bibinfo{journal}{\emph{IEEE/ACM Transactions on Audio, Speech, and
  Language Processing}} \bibinfo{volume}{24}, \bibinfo{number}{11}
  (\bibinfo{year}{2016}), \bibinfo{pages}{2146--2157}.
\newblock


\bibitem[\protect\citeauthoryear{Chen, Krishna, Emer, and Sze}{Chen
  et~al\mbox{.}}{2017}]%
        {chen2017eyeriss}
\bibfield{author}{\bibinfo{person}{Yu-Hsin Chen}, \bibinfo{person}{Tushar
  Krishna}, \bibinfo{person}{Joel~S Emer}, {and} \bibinfo{person}{Vivienne
  Sze}.} \bibinfo{year}{2017}\natexlab{}.
\newblock \showarticletitle{Eyeriss: An energy-efficient reconfigurable
  accelerator for deep convolutional neural networks}.
\newblock \bibinfo{journal}{\emph{IEEE Journal of Solid-State Circuits}}
  \bibinfo{volume}{52}, \bibinfo{number}{1} (\bibinfo{year}{2017}),
  \bibinfo{pages}{127--138}.
\newblock


\bibitem[\protect\citeauthoryear{Cheng, Dong, Hsu, Chang, Sun, Chang, Pan,
  Chen, Wei, and Juan}{Cheng et~al\mbox{.}}{2018}]%
        {cheng2018searching}
\bibfield{author}{\bibinfo{person}{An-Chieh Cheng}, \bibinfo{person}{Jin-Dong
  Dong}, \bibinfo{person}{Chi-Hung Hsu}, \bibinfo{person}{Shu-Huan Chang},
  \bibinfo{person}{Min Sun}, \bibinfo{person}{Shih-Chieh Chang},
  \bibinfo{person}{Jia-Yu Pan}, \bibinfo{person}{Yu-Ting Chen},
  \bibinfo{person}{Wei Wei}, {and} \bibinfo{person}{Da-Cheng Juan}.}
  \bibinfo{year}{2018}\natexlab{}.
\newblock \showarticletitle{Searching Toward Pareto-Optimal Device-Aware Neural
  Architectures}.
\newblock \bibinfo{journal}{\emph{arXiv preprint arXiv:1808.09830}}
  (\bibinfo{year}{2018}).
\newblock


\bibitem[\protect\citeauthoryear{Cherkassky and Mulier}{Cherkassky and
  Mulier}{2007}]%
        {cherkassky2007learning}
\bibfield{author}{\bibinfo{person}{Vladimir Cherkassky} {and}
  \bibinfo{person}{Filip~M Mulier}.} \bibinfo{year}{2007}\natexlab{}.
\newblock \bibinfo{booktitle}{\emph{Learning from data: concepts, theory, and
  methods}}.
\newblock \bibinfo{publisher}{John Wiley \& Sons}.
\newblock


\bibitem[\protect\citeauthoryear{Choi, Burleson, and Phatak}{Choi
  et~al\mbox{.}}{1993}]%
        {choi1993fixed}
\bibfield{author}{\bibinfo{person}{H Choi}, \bibinfo{person}{WP Burleson},
  {and} \bibinfo{person}{DS Phatak}.} \bibinfo{year}{1993}\natexlab{}.
\newblock \showarticletitle{Fixed-point roundoff error analysis of large
  feedforward neural networks}. In \bibinfo{booktitle}{\emph{Proceedings of
  1993 International Conference on Neural Networks (IJCNN-93-Nagoya, Japan)}},
  Vol.~\bibinfo{volume}{2}. IEEE, \bibinfo{pages}{1947--1950}.
\newblock


\bibitem[\protect\citeauthoryear{Chu, Kim, Lin, Yu, Bradski, Olukotun, and
  Ng}{Chu et~al\mbox{.}}{2007}]%
        {chu2007map}
\bibfield{author}{\bibinfo{person}{Cheng-Tao Chu}, \bibinfo{person}{Sang~K
  Kim}, \bibinfo{person}{Yi-An Lin}, \bibinfo{person}{YuanYuan Yu},
  \bibinfo{person}{Gary Bradski}, \bibinfo{person}{Kunle Olukotun}, {and}
  \bibinfo{person}{Andrew~Y Ng}.} \bibinfo{year}{2007}\natexlab{}.
\newblock \showarticletitle{Map-reduce for machine learning on multicore}. In
  \bibinfo{booktitle}{\emph{Advances in neural information processing
  systems}}. \bibinfo{pages}{281--288}.
\newblock


\bibitem[\protect\citeauthoryear{Coleman, Narayanan, Kang, Zhao, Zhang, Nardi,
  Bailis, Olukotun, R{\'e}, and Zaharia}{Coleman et~al\mbox{.}}{2017}]%
        {coleman2017dawnbench}
\bibfield{author}{\bibinfo{person}{Cody Coleman}, \bibinfo{person}{Deepak
  Narayanan}, \bibinfo{person}{Daniel Kang}, \bibinfo{person}{Tian Zhao},
  \bibinfo{person}{Jian Zhang}, \bibinfo{person}{Luigi Nardi},
  \bibinfo{person}{Peter Bailis}, \bibinfo{person}{Kunle Olukotun},
  \bibinfo{person}{Chris R{\'e}}, {and} \bibinfo{person}{Matei Zaharia}.}
  \bibinfo{year}{2017}\natexlab{}.
\newblock \showarticletitle{DAWNBench: An End-to-End Deep Learning Benchmark
  and Competition}.
\newblock \bibinfo{journal}{\emph{Training}} \bibinfo{volume}{100},
  \bibinfo{number}{101} (\bibinfo{year}{2017}), \bibinfo{pages}{102}.
\newblock


\bibitem[\protect\citeauthoryear{Courbariaux, Bengio, and David}{Courbariaux
  et~al\mbox{.}}{2015}]%
        {courbariaux2015binaryconnect}
\bibfield{author}{\bibinfo{person}{Matthieu Courbariaux},
  \bibinfo{person}{Yoshua Bengio}, {and} \bibinfo{person}{Jean-Pierre David}.}
  \bibinfo{year}{2015}\natexlab{}.
\newblock \showarticletitle{Binaryconnect: Training deep neural networks with
  binary weights during propagations}. In \bibinfo{booktitle}{\emph{Advances in
  neural information processing systems}}. \bibinfo{pages}{3123--3131}.
\newblock


\bibitem[\protect\citeauthoryear{Courbariaux, Hubara, Soudry, El-Yaniv, and
  Bengio}{Courbariaux et~al\mbox{.}}{2016}]%
        {courbariaux2016binarized}
\bibfield{author}{\bibinfo{person}{Matthieu Courbariaux}, \bibinfo{person}{Itay
  Hubara}, \bibinfo{person}{Daniel Soudry}, \bibinfo{person}{Ran El-Yaniv},
  {and} \bibinfo{person}{Yoshua Bengio}.} \bibinfo{year}{2016}\natexlab{}.
\newblock \showarticletitle{Binarized neural networks: Training deep neural
  networks with weights and activations constrained to+ 1 or-1}.
\newblock \bibinfo{journal}{\emph{arXiv preprint arXiv:1602.02830}}
  (\bibinfo{year}{2016}).
\newblock


\bibitem[\protect\citeauthoryear{Crammer, Kulesza, and Dredze}{Crammer
  et~al\mbox{.}}{2013}]%
        {crammer2013adaptive}
\bibfield{author}{\bibinfo{person}{Koby Crammer}, \bibinfo{person}{Alex
  Kulesza}, {and} \bibinfo{person}{Mark Dredze}.}
  \bibinfo{year}{2013}\natexlab{}.
\newblock \showarticletitle{Adaptive regularization of weight vectors}.
\newblock \bibinfo{journal}{\emph{Machine learning}} \bibinfo{volume}{91},
  \bibinfo{number}{2} (\bibinfo{year}{2013}), \bibinfo{pages}{155--187}.
\newblock


\bibitem[\protect\citeauthoryear{Dai, Zhang, Wu, Yin, Sun, Wang, Dukhan, Hu,
  Wu, Jia, et~al\mbox{.}}{Dai et~al\mbox{.}}{2019}]%
        {dai2019chamnet}
\bibfield{author}{\bibinfo{person}{Xiaoliang Dai}, \bibinfo{person}{Peizhao
  Zhang}, \bibinfo{person}{Bichen Wu}, \bibinfo{person}{Hongxu Yin},
  \bibinfo{person}{Fei Sun}, \bibinfo{person}{Yanghan Wang},
  \bibinfo{person}{Marat Dukhan}, \bibinfo{person}{Yunqing Hu},
  \bibinfo{person}{Yiming Wu}, \bibinfo{person}{Yangqing Jia}, {et~al\mbox{.}}}
  \bibinfo{year}{2019}\natexlab{}.
\newblock \showarticletitle{Chamnet: Towards efficient network design through
  platform-aware model adaptation}. In \bibinfo{booktitle}{\emph{Proceedings of
  the IEEE Conference on Computer Vision and Pattern Recognition}}.
  \bibinfo{pages}{11398--11407}.
\newblock


\bibitem[\protect\citeauthoryear{Daniely, Linial, and Shalev-Shwartz}{Daniely
  et~al\mbox{.}}{2013a}]%
        {daniely2013average}
\bibfield{author}{\bibinfo{person}{Amit Daniely}, \bibinfo{person}{Nati
  Linial}, {and} \bibinfo{person}{Shai Shalev-Shwartz}.}
  \bibinfo{year}{2013}\natexlab{a}.
\newblock \showarticletitle{From average case complexity to improper learning
  complexity}.
\newblock \bibinfo{journal}{\emph{arXiv preprint arXiv:1311.2272}}
  (\bibinfo{year}{2013}).
\newblock


\bibitem[\protect\citeauthoryear{Daniely, Linial, and Shalev-Shwartz}{Daniely
  et~al\mbox{.}}{2013b}]%
        {daniely2013more}
\bibfield{author}{\bibinfo{person}{Amit Daniely}, \bibinfo{person}{Nati
  Linial}, {and} \bibinfo{person}{Shai Shalev-Shwartz}.}
  \bibinfo{year}{2013}\natexlab{b}.
\newblock \showarticletitle{More data speeds up training time in learning
  halfspaces over sparse vectors}. In \bibinfo{booktitle}{\emph{Advances in
  Neural Information Processing Systems}}. \bibinfo{pages}{145--153}.
\newblock


\bibitem[\protect\citeauthoryear{Dasgupta and Gschwend}{Dasgupta and
  Gschwend}{[n.d.]}]%
        {netscope}
\bibfield{author}{\bibinfo{person}{Saumitro Dasgupta} {and}
  \bibinfo{person}{David Gschwend}.} \bibinfo{year}{[n.d.]}\natexlab{}.
\newblock \bibinfo{booktitle}{\emph{Netscope CNN Analyzer}}.
\newblock
\urldef\tempurl%
\url{https://dgschwend.github.io/netscope/quickstart.html}
\showURL{%
\tempurl}


\bibitem[\protect\citeauthoryear{De~Sa, Feldman, R{\'e}, and Olukotun}{De~Sa
  et~al\mbox{.}}{2017}]%
        {de2017understanding}
\bibfield{author}{\bibinfo{person}{Christopher De~Sa}, \bibinfo{person}{Matthew
  Feldman}, \bibinfo{person}{Christopher R{\'e}}, {and} \bibinfo{person}{Kunle
  Olukotun}.} \bibinfo{year}{2017}\natexlab{}.
\newblock \showarticletitle{Understanding and optimizing asynchronous
  low-precision stochastic gradient descent}. In
  \bibinfo{booktitle}{\emph{Proceedings of the 44th Annual International
  Symposium on Computer Architecture}}. \bibinfo{pages}{561--574}.
\newblock


\bibitem[\protect\citeauthoryear{De~Sa, Leszczynski, Zhang, Marzoev, Aberger,
  Olukotun, and R{\'e}}{De~Sa et~al\mbox{.}}{2018}]%
        {de2018high}
\bibfield{author}{\bibinfo{person}{Christopher De~Sa}, \bibinfo{person}{Megan
  Leszczynski}, \bibinfo{person}{Jian Zhang}, \bibinfo{person}{Alana Marzoev},
  \bibinfo{person}{Christopher~R Aberger}, \bibinfo{person}{Kunle Olukotun},
  {and} \bibinfo{person}{Christopher R{\'e}}.} \bibinfo{year}{2018}\natexlab{}.
\newblock \showarticletitle{High-accuracy low-precision training}.
\newblock \bibinfo{journal}{\emph{arXiv preprint arXiv:1803.03383}}
  (\bibinfo{year}{2018}).
\newblock


\bibitem[\protect\citeauthoryear{De~Sa, Zhang, Olukotun, and R{\'e}}{De~Sa
  et~al\mbox{.}}{2015}]%
        {de2015taming}
\bibfield{author}{\bibinfo{person}{Christopher~M De~Sa}, \bibinfo{person}{Ce
  Zhang}, \bibinfo{person}{Kunle Olukotun}, {and} \bibinfo{person}{Christopher
  R{\'e}}.} \bibinfo{year}{2015}\natexlab{}.
\newblock \showarticletitle{Taming the wild: A unified analysis of
  hogwild-style algorithms}. In \bibinfo{booktitle}{\emph{Advances in neural
  information processing systems}}. \bibinfo{pages}{2674--2682}.
\newblock


\bibitem[\protect\citeauthoryear{Dean, Corrado, Monga, Chen, Devin, Mao,
  Senior, Tucker, Yang, Le, et~al\mbox{.}}{Dean et~al\mbox{.}}{2012}]%
        {dean2012large}
\bibfield{author}{\bibinfo{person}{Jeffrey Dean}, \bibinfo{person}{Greg
  Corrado}, \bibinfo{person}{Rajat Monga}, \bibinfo{person}{Kai Chen},
  \bibinfo{person}{Matthieu Devin}, \bibinfo{person}{Mark Mao},
  \bibinfo{person}{Andrew Senior}, \bibinfo{person}{Paul Tucker},
  \bibinfo{person}{Ke Yang}, \bibinfo{person}{Quoc~V Le}, {et~al\mbox{.}}}
  \bibinfo{year}{2012}\natexlab{}.
\newblock \showarticletitle{Large scale distributed deep networks}. In
  \bibinfo{booktitle}{\emph{Advances in neural information processing
  systems}}. \bibinfo{pages}{1223--1231}.
\newblock


\bibitem[\protect\citeauthoryear{Decatur, Goldreich, and Ron}{Decatur
  et~al\mbox{.}}{2000}]%
        {decatur2000computational}
\bibfield{author}{\bibinfo{person}{Scott~E Decatur}, \bibinfo{person}{Oded
  Goldreich}, {and} \bibinfo{person}{Dana Ron}.}
  \bibinfo{year}{2000}\natexlab{}.
\newblock \showarticletitle{Computational sample complexity}.
\newblock \bibinfo{journal}{\emph{SIAM J. Comput.}} \bibinfo{volume}{29},
  \bibinfo{number}{3} (\bibinfo{year}{2000}), \bibinfo{pages}{854--879}.
\newblock


\bibitem[\protect\citeauthoryear{Dey, Mukherjee, and Pal}{Dey
  et~al\mbox{.}}{2019}]%
        {dey2019embedded}
\bibfield{author}{\bibinfo{person}{Swarnava Dey}, \bibinfo{person}{Arijit
  Mukherjee}, {and} \bibinfo{person}{Arpan Pal}.}
  \bibinfo{year}{2019}\natexlab{}.
\newblock \showarticletitle{Embedded Deep Inference in Practice: Case for Model
  Partitioning}. In \bibinfo{booktitle}{\emph{Proceedings of the 1st Workshop
  on Machine Learning on Edge in Sensor Systems}}. \bibinfo{pages}{25--30}.
\newblock


\bibitem[\protect\citeauthoryear{Dhar, Cherkassky, and Shah}{Dhar
  et~al\mbox{.}}{2019}]%
        {dhar2019multiclass}
\bibfield{author}{\bibinfo{person}{Sauptik Dhar}, \bibinfo{person}{Vladimir
  Cherkassky}, {and} \bibinfo{person}{Mohak Shah}.}
  \bibinfo{year}{2019}\natexlab{}.
\newblock \showarticletitle{Multiclass Learning from Contradictions}. In
  \bibinfo{booktitle}{\emph{Advances in Neural Information Processing
  Systems}}. \bibinfo{pages}{8400--8410}.
\newblock


\bibitem[\protect\citeauthoryear{Drakopoulos, Ozdaglar, and
  Tsitsiklis}{Drakopoulos et~al\mbox{.}}{2013}]%
        {drakopoulos2013learning}
\bibfield{author}{\bibinfo{person}{Kimon Drakopoulos}, \bibinfo{person}{Asuman
  Ozdaglar}, {and} \bibinfo{person}{John~N Tsitsiklis}.}
  \bibinfo{year}{2013}\natexlab{}.
\newblock \showarticletitle{On learning with finite memory}.
\newblock \bibinfo{journal}{\emph{IEEE Transactions on Information Theory}}
  \bibinfo{volume}{59}, \bibinfo{number}{10} (\bibinfo{year}{2013}),
  \bibinfo{pages}{6859--6872}.
\newblock


\bibitem[\protect\citeauthoryear{Fan, Stock, Graham, Grave, Gribonval, Jegou,
  and Joulin}{Fan et~al\mbox{.}}{2020}]%
        {fan2020training}
\bibfield{author}{\bibinfo{person}{Angela Fan}, \bibinfo{person}{Pierre Stock},
  \bibinfo{person}{Benjamin Graham}, \bibinfo{person}{Edouard Grave},
  \bibinfo{person}{Remi Gribonval}, \bibinfo{person}{Herve Jegou}, {and}
  \bibinfo{person}{Armand Joulin}.} \bibinfo{year}{2020}\natexlab{}.
\newblock \bibinfo{title}{Training with Quantization Noise for Extreme Model
  Compression}.
\newblock
\newblock
\showeprint[arxiv]{cs.LG/2004.07320}


\bibitem[\protect\citeauthoryear{Fang, Zeng, and Zhang}{Fang
  et~al\mbox{.}}{2018}]%
        {fang2018nestdnn}
\bibfield{author}{\bibinfo{person}{Biyi Fang}, \bibinfo{person}{Xiao Zeng},
  {and} \bibinfo{person}{Mi Zhang}.} \bibinfo{year}{2018}\natexlab{}.
\newblock \showarticletitle{{NestDNN}: Resource-Aware Multi-Tenant On-Device
  Deep Learning for Continuous Mobile Vision}. In
  \bibinfo{booktitle}{\emph{Proceedings of the 24th Annual International
  Conference on Mobile Computing and Networking}}. ACM,
  \bibinfo{pages}{115--127}.
\newblock


\bibitem[\protect\citeauthoryear{Feldman et~al\mbox{.}}{Feldman
  et~al\mbox{.}}{2007}]%
        {feldman2007efficiency}
\bibfield{author}{\bibinfo{person}{Vitaly Feldman} {et~al\mbox{.}}}
  \bibinfo{year}{2007}\natexlab{}.
\newblock \bibinfo{booktitle}{\emph{Efficiency and computational limitations of
  learning algorithms}}. Vol.~\bibinfo{volume}{68}.
\newblock


\bibitem[\protect\citeauthoryear{Feldman, Grigorescu, Reyzin, Vempala, and
  Xiao}{Feldman et~al\mbox{.}}{2013}]%
        {feldman2013statistical}
\bibfield{author}{\bibinfo{person}{Vitaly Feldman}, \bibinfo{person}{Elena
  Grigorescu}, \bibinfo{person}{Lev Reyzin}, \bibinfo{person}{Santosh Vempala},
  {and} \bibinfo{person}{Ying Xiao}.} \bibinfo{year}{2013}\natexlab{}.
\newblock \showarticletitle{Statistical algorithms and a lower bound for
  detecting planted cliques}. In \bibinfo{booktitle}{\emph{Proceedings of the
  forty-fifth annual ACM symposium on Theory of computing}}. ACM,
  \bibinfo{pages}{655--664}.
\newblock


\bibitem[\protect\citeauthoryear{Feldman, Guzm{\'a}n, and Vempala}{Feldman
  et~al\mbox{.}}{2017}]%
        {feldman2017statistical}
\bibfield{author}{\bibinfo{person}{Vitaly Feldman},
  \bibinfo{person}{Crist{\'o}bal Guzm{\'a}n}, {and} \bibinfo{person}{Santosh
  Vempala}.} \bibinfo{year}{2017}\natexlab{}.
\newblock \showarticletitle{Statistical query algorithms for mean vector
  estimation and stochastic convex optimization}. In
  \bibinfo{booktitle}{\emph{Proceedings of the Twenty-Eighth Annual ACM-SIAM
  Symposium on Discrete Algorithms}}. SIAM, \bibinfo{pages}{1265--1277}.
\newblock


\bibitem[\protect\citeauthoryear{Feng and Darrell}{Feng and Darrell}{2015}]%
        {feng2015learning}
\bibfield{author}{\bibinfo{person}{Jiashi Feng} {and} \bibinfo{person}{Trevor
  Darrell}.} \bibinfo{year}{2015}\natexlab{}.
\newblock \showarticletitle{Learning the structure of deep convolutional
  networks}. In \bibinfo{booktitle}{\emph{Proceedings of the IEEE international
  conference on computer vision}}. \bibinfo{pages}{2749--2757}.
\newblock


\bibitem[\protect\citeauthoryear{Fern{\'a}ndez-Delgado, Cernadas, Barro, and
  Amorim}{Fern{\'a}ndez-Delgado et~al\mbox{.}}{2014}]%
        {fernandez2014we}
\bibfield{author}{\bibinfo{person}{Manuel Fern{\'a}ndez-Delgado},
  \bibinfo{person}{Eva Cernadas}, \bibinfo{person}{Sen{\'e}n Barro}, {and}
  \bibinfo{person}{Dinani Amorim}.} \bibinfo{year}{2014}\natexlab{}.
\newblock \showarticletitle{Do we need hundreds of classifiers to solve real
  world classification problems?}
\newblock \bibinfo{journal}{\emph{The Journal of Machine Learning Research}}
  \bibinfo{volume}{15}, \bibinfo{number}{1} (\bibinfo{year}{2014}),
  \bibinfo{pages}{3133--3181}.
\newblock


\bibitem[\protect\citeauthoryear{Fern{\'a}ndez-Delgado, Sirsat, Cernadas,
  Alawadi, Barro, and Febrero-Bande}{Fern{\'a}ndez-Delgado
  et~al\mbox{.}}{2019}]%
        {fernandez2019extensive}
\bibfield{author}{\bibinfo{person}{M Fern{\'a}ndez-Delgado},
  \bibinfo{person}{MS Sirsat}, \bibinfo{person}{Eva Cernadas},
  \bibinfo{person}{Sadi Alawadi}, \bibinfo{person}{Sen{\'e}n Barro}, {and}
  \bibinfo{person}{Manuel Febrero-Bande}.} \bibinfo{year}{2019}\natexlab{}.
\newblock \showarticletitle{An extensive experimental survey of regression
  methods}.
\newblock \bibinfo{journal}{\emph{Neural Networks}}  \bibinfo{volume}{111}
  (\bibinfo{year}{2019}), \bibinfo{pages}{11--34}.
\newblock


\bibitem[\protect\citeauthoryear{Gandikota, Maity, and Mazumdar}{Gandikota
  et~al\mbox{.}}{2019}]%
        {gandikota2019vqsgd}
\bibfield{author}{\bibinfo{person}{Venkata Gandikota},
  \bibinfo{person}{Raj~Kumar Maity}, {and} \bibinfo{person}{Arya Mazumdar}.}
  \bibinfo{year}{2019}\natexlab{}.
\newblock \showarticletitle{vqSGD: Vector quantized stochastic gradient
  descent}.
\newblock \bibinfo{journal}{\emph{arXiv preprint arXiv:1911.07971}}
  (\bibinfo{year}{2019}).
\newblock


\bibitem[\protect\citeauthoryear{Gao, Zhan, Wang, Luo, Zheng, Ren, Zheng, Lu,
  Li, Cao, et~al\mbox{.}}{Gao et~al\mbox{.}}{2018}]%
        {gao2018bigdatabench}
\bibfield{author}{\bibinfo{person}{Wanling Gao}, \bibinfo{person}{Jianfeng
  Zhan}, \bibinfo{person}{Lei Wang}, \bibinfo{person}{Chunjie Luo},
  \bibinfo{person}{Daoyi Zheng}, \bibinfo{person}{Rui Ren},
  \bibinfo{person}{Chen Zheng}, \bibinfo{person}{Gang Lu},
  \bibinfo{person}{Jingwei Li}, \bibinfo{person}{Zheng Cao}, {et~al\mbox{.}}}
  \bibinfo{year}{2018}\natexlab{}.
\newblock \showarticletitle{BigDataBench: A Dwarf-based Big Data and AI
  Benchmark Suite}.
\newblock \bibinfo{journal}{\emph{arXiv preprint arXiv:1802.08254}}
  (\bibinfo{year}{2018}).
\newblock


\bibitem[\protect\citeauthoryear{Garg, Raz, and Tal}{Garg
  et~al\mbox{.}}{2018}]%
        {garg2018extractor}
\bibfield{author}{\bibinfo{person}{Sumegha Garg}, \bibinfo{person}{Ran Raz},
  {and} \bibinfo{person}{Avishay Tal}.} \bibinfo{year}{2018}\natexlab{}.
\newblock \showarticletitle{Extractor-based time-space lower bounds for
  learning}. In \bibinfo{booktitle}{\emph{Proceedings of the 50th Annual ACM
  SIGACT Symposium on Theory of Computing}}. ACM, \bibinfo{pages}{990--1002}.
\newblock


\bibitem[\protect\citeauthoryear{Geng, Huang, and Chen}{Geng
  et~al\mbox{.}}{2020}]%
        {geng2020recent}
\bibfield{author}{\bibinfo{person}{Chuanxing Geng}, \bibinfo{person}{Sheng-jun
  Huang}, {and} \bibinfo{person}{Songcan Chen}.}
  \bibinfo{year}{2020}\natexlab{}.
\newblock \showarticletitle{Recent advances in open set recognition: A survey}.
\newblock \bibinfo{journal}{\emph{IEEE Transactions on Pattern Analysis and
  Machine Intelligence}} (\bibinfo{year}{2020}).
\newblock


\bibitem[\protect\citeauthoryear{Gholami, Kwon, Wu, Tai, Yue, Jin, Zhao, and
  Keutzer}{Gholami et~al\mbox{.}}{2018}]%
        {gholami2018squeezenext}
\bibfield{author}{\bibinfo{person}{Amir Gholami}, \bibinfo{person}{Kiseok
  Kwon}, \bibinfo{person}{Bichen Wu}, \bibinfo{person}{Zizheng Tai},
  \bibinfo{person}{Xiangyu Yue}, \bibinfo{person}{Peter Jin},
  \bibinfo{person}{Sicheng Zhao}, {and} \bibinfo{person}{Kurt Keutzer}.}
  \bibinfo{year}{2018}\natexlab{}.
\newblock \showarticletitle{SqueezeNext: Hardware-Aware Neural Network Design}.
\newblock \bibinfo{journal}{\emph{arXiv preprint arXiv:1803.10615}}
  (\bibinfo{year}{2018}).
\newblock


\bibitem[\protect\citeauthoryear{Gianniti, Zhang, and Ardagna}{Gianniti
  et~al\mbox{.}}{[n.d.]}]%
        {giannitiperformance}
\bibfield{author}{\bibinfo{person}{Eugenio Gianniti}, \bibinfo{person}{Li
  Zhang}, {and} \bibinfo{person}{Danilo Ardagna}.}
  \bibinfo{year}{[n.d.]}\natexlab{}.
\newblock \showarticletitle{Performance Prediction of GPU-based Deep Learning
  Applications}.
\newblock  (\bibinfo{year}{[n.\,d.]}).
\newblock


\bibitem[\protect\citeauthoryear{Golovin, Sculley, McMahan, and Young}{Golovin
  et~al\mbox{.}}{2013}]%
        {golovin2013large}
\bibfield{author}{\bibinfo{person}{Daniel Golovin}, \bibinfo{person}{D
  Sculley}, \bibinfo{person}{Brendan McMahan}, {and} \bibinfo{person}{Michael
  Young}.} \bibinfo{year}{2013}\natexlab{}.
\newblock \showarticletitle{Large-scale learning with less ram via
  randomization}. In \bibinfo{booktitle}{\emph{International Conference on
  Machine Learning}}. \bibinfo{pages}{325--333}.
\newblock


\bibitem[\protect\citeauthoryear{Gong, Liu, Yang, and Bourdev}{Gong
  et~al\mbox{.}}{2014}]%
        {gong2014compressing}
\bibfield{author}{\bibinfo{person}{Yunchao Gong}, \bibinfo{person}{Liu Liu},
  \bibinfo{person}{Ming Yang}, {and} \bibinfo{person}{Lubomir Bourdev}.}
  \bibinfo{year}{2014}\natexlab{}.
\newblock \showarticletitle{Compressing deep convolutional networks using
  vector quantization}.
\newblock \bibinfo{journal}{\emph{arXiv preprint arXiv:1412.6115}}
  (\bibinfo{year}{2014}).
\newblock


\bibitem[\protect\citeauthoryear{Goodfellow, Shlens, and Szegedy}{Goodfellow
  et~al\mbox{.}}{2014}]%
        {goodfellow2014explaining}
\bibfield{author}{\bibinfo{person}{Ian~J Goodfellow}, \bibinfo{person}{Jonathon
  Shlens}, {and} \bibinfo{person}{Christian Szegedy}.}
  \bibinfo{year}{2014}\natexlab{}.
\newblock \showarticletitle{Explaining and harnessing adversarial examples}.
\newblock \bibinfo{journal}{\emph{arXiv preprint arXiv:1412.6572}}
  (\bibinfo{year}{2014}).
\newblock


\bibitem[\protect\citeauthoryear{Greff, Srivastava, and Schmidhuber}{Greff
  et~al\mbox{.}}{2017}]%
        {greff2016highway}
\bibfield{author}{\bibinfo{person}{Klaus Greff}, \bibinfo{person}{Rupesh~K
  Srivastava}, {and} \bibinfo{person}{J{\"u}rgen Schmidhuber}.}
  \bibinfo{year}{2017}\natexlab{}.
\newblock \showarticletitle{Highway and residual networks learn unrolled
  iterative estimation}. In \bibinfo{booktitle}{\emph{ICLR}}.
\newblock


\bibitem[\protect\citeauthoryear{Grubb and Bagnell}{Grubb and Bagnell}{2012}]%
        {grubb2012speedboost}
\bibfield{author}{\bibinfo{person}{Alex Grubb} {and} \bibinfo{person}{Drew
  Bagnell}.} \bibinfo{year}{2012}\natexlab{}.
\newblock \showarticletitle{Speedboost: Anytime prediction with uniform
  near-optimality}. In \bibinfo{booktitle}{\emph{Artificial Intelligence and
  Statistics}}. \bibinfo{pages}{458--466}.
\newblock


\bibitem[\protect\citeauthoryear{Gruslys, Munos, Danihelka, Lanctot, and
  Graves}{Gruslys et~al\mbox{.}}{2016}]%
        {gruslys2016memory}
\bibfield{author}{\bibinfo{person}{Audrunas Gruslys}, \bibinfo{person}{R{\'e}mi
  Munos}, \bibinfo{person}{Ivo Danihelka}, \bibinfo{person}{Marc Lanctot},
  {and} \bibinfo{person}{Alex Graves}.} \bibinfo{year}{2016}\natexlab{}.
\newblock \showarticletitle{Memory-efficient backpropagation through time}. In
  \bibinfo{booktitle}{\emph{Advances in Neural Information Processing
  Systems}}. \bibinfo{pages}{4125--4133}.
\newblock


\bibitem[\protect\citeauthoryear{Gu, Yang, and Wu}{Gu et~al\mbox{.}}{2019}]%
        {gu2019distributed}
\bibfield{author}{\bibinfo{person}{Renjie Gu}, \bibinfo{person}{Shuo Yang},
  {and} \bibinfo{person}{Fan Wu}.} \bibinfo{year}{2019}\natexlab{}.
\newblock \showarticletitle{Distributed machine learning on mobile devices: A
  survey}.
\newblock \bibinfo{journal}{\emph{arXiv preprint arXiv:1909.08329}}
  (\bibinfo{year}{2019}).
\newblock


\bibitem[\protect\citeauthoryear{Guo}{Guo}{2018}]%
        {guo2018survey}
\bibfield{author}{\bibinfo{person}{Yunhui Guo}.}
  \bibinfo{year}{2018}\natexlab{}.
\newblock \showarticletitle{A survey on methods and theories of quantized
  neural networks}.
\newblock \bibinfo{journal}{\emph{arXiv preprint arXiv:1808.04752}}
  (\bibinfo{year}{2018}).
\newblock


\bibitem[\protect\citeauthoryear{Gupta, Agrawal, Gopalakrishnan, and
  Narayanan}{Gupta et~al\mbox{.}}{2015}]%
        {gupta2015deep}
\bibfield{author}{\bibinfo{person}{Suyog Gupta}, \bibinfo{person}{Ankur
  Agrawal}, \bibinfo{person}{Kailash Gopalakrishnan}, {and}
  \bibinfo{person}{Pritish Narayanan}.} \bibinfo{year}{2015}\natexlab{}.
\newblock \showarticletitle{Deep learning with limited numerical precision}. In
  \bibinfo{booktitle}{\emph{International Conference on Machine Learning}}.
  \bibinfo{pages}{1737--1746}.
\newblock


\bibitem[\protect\citeauthoryear{Haigh, Mackay, Cook, and Lin}{Haigh
  et~al\mbox{.}}{2015}]%
        {haigh2015machine}
\bibfield{author}{\bibinfo{person}{Karen~Zita Haigh}, \bibinfo{person}{Allan~M
  Mackay}, \bibinfo{person}{Michael~R Cook}, {and} \bibinfo{person}{Li~G Lin}.}
  \bibinfo{year}{2015}\natexlab{}.
\newblock \showarticletitle{Machine learning for embedded systems: A case
  study}.
\newblock \bibinfo{journal}{\emph{Technical Report}} (\bibinfo{year}{2015}).
\newblock


\bibitem[\protect\citeauthoryear{Han, Mao, and Dally}{Han
  et~al\mbox{.}}{2015}]%
        {han2015deep}
\bibfield{author}{\bibinfo{person}{Song Han}, \bibinfo{person}{Huizi Mao},
  {and} \bibinfo{person}{William~J Dally}.} \bibinfo{year}{2015}\natexlab{}.
\newblock \showarticletitle{Deep compression: Compressing deep neural networks
  with pruning, trained quantization and huffman coding}.
\newblock \bibinfo{journal}{\emph{arXiv preprint arXiv:1510.00149}}
  (\bibinfo{year}{2015}).
\newblock


\bibitem[\protect\citeauthoryear{Hazan and Koren}{Hazan and Koren}{2012}]%
        {hazan2012linear}
\bibfield{author}{\bibinfo{person}{Elad Hazan} {and} \bibinfo{person}{Tomer
  Koren}.} \bibinfo{year}{2012}\natexlab{}.
\newblock \showarticletitle{Linear regression with limited observation}.
\newblock \bibinfo{journal}{\emph{arXiv preprint arXiv:1206.4678}}
  (\bibinfo{year}{2012}).
\newblock


\bibitem[\protect\citeauthoryear{He, Zhang, Ren, and Sun}{He
  et~al\mbox{.}}{2016}]%
        {he2016deep}
\bibfield{author}{\bibinfo{person}{Kaiming He}, \bibinfo{person}{Xiangyu
  Zhang}, \bibinfo{person}{Shaoqing Ren}, {and} \bibinfo{person}{Jian Sun}.}
  \bibinfo{year}{2016}\natexlab{}.
\newblock \showarticletitle{Deep residual learning for image recognition}. In
  \bibinfo{booktitle}{\emph{Proceedings of the IEEE conference on computer
  vision and pattern recognition}}. \bibinfo{pages}{770--778}.
\newblock


\bibitem[\protect\citeauthoryear{He, Lin, Liu, Wang, Li, and Han}{He
  et~al\mbox{.}}{2018}]%
        {he2018amc}
\bibfield{author}{\bibinfo{person}{Yihui He}, \bibinfo{person}{Ji Lin},
  \bibinfo{person}{Zhijian Liu}, \bibinfo{person}{Hanrui Wang},
  \bibinfo{person}{Li-Jia Li}, {and} \bibinfo{person}{Song Han}.}
  \bibinfo{year}{2018}\natexlab{}.
\newblock \showarticletitle{{AMC: AutoML for Model Compression and Acceleration
  on Mobile Devices}}. In \bibinfo{booktitle}{\emph{The European Conference on
  Computer Vision (ECCV)}}. \bibinfo{pages}{784--800}.
\newblock


\bibitem[\protect\citeauthoryear{Hellman}{Hellman}{1972}]%
        {hellman1972effects}
\bibfield{author}{\bibinfo{person}{Martin Hellman}.}
  \bibinfo{year}{1972}\natexlab{}.
\newblock \showarticletitle{The effects of randomization on finite-memory
  decision schemes}.
\newblock \bibinfo{journal}{\emph{IEEE Transactions on Information Theory}}
  \bibinfo{volume}{18}, \bibinfo{number}{4} (\bibinfo{year}{1972}),
  \bibinfo{pages}{499--502}.
\newblock


\bibitem[\protect\citeauthoryear{Hellman, Cover, et~al\mbox{.}}{Hellman
  et~al\mbox{.}}{1971}]%
        {hellman1971memory}
\bibfield{author}{\bibinfo{person}{Martin~E Hellman}, \bibinfo{person}{Thomas~M
  Cover}, {et~al\mbox{.}}} \bibinfo{year}{1971}\natexlab{}.
\newblock \showarticletitle{On memory saved by randomization}.
\newblock \bibinfo{journal}{\emph{The Annals of Mathematical Statistics}}
  \bibinfo{volume}{42}, \bibinfo{number}{3} (\bibinfo{year}{1971}),
  \bibinfo{pages}{1075--1078}.
\newblock


\bibitem[\protect\citeauthoryear{Henzinger, Raghavan, and
  Rajagopalan}{Henzinger et~al\mbox{.}}{1998}]%
        {henzinger1998computing}
\bibfield{author}{\bibinfo{person}{Monika~Rauch Henzinger},
  \bibinfo{person}{Prabhakar Raghavan}, {and} \bibinfo{person}{Sridhar
  Rajagopalan}.} \bibinfo{year}{1998}\natexlab{}.
\newblock \showarticletitle{Computing on data streams.}
\newblock \bibinfo{journal}{\emph{External memory algorithms}}
  \bibinfo{volume}{50} (\bibinfo{year}{1998}), \bibinfo{pages}{107--118}.
\newblock


\bibitem[\protect\citeauthoryear{Hinton, Vinyals, and Dean}{Hinton
  et~al\mbox{.}}{2015}]%
        {44873}
\bibfield{author}{\bibinfo{person}{Geoffrey Hinton}, \bibinfo{person}{Oriol
  Vinyals}, {and} \bibinfo{person}{Jeffrey Dean}.}
  \bibinfo{year}{2015}\natexlab{}.
\newblock \showarticletitle{Distilling the Knowledge in a Neural Network}. In
  \bibinfo{booktitle}{\emph{NIPS Deep Learning and Representation Learning
  Workshop}}.
\newblock
\urldef\tempurl%
\url{http://arxiv.org/abs/1503.02531}
\showURL{%
\tempurl}


\bibitem[\protect\citeauthoryear{Hinton, Srivastava, Krizhevsky, Sutskever, and
  Salakhutdinov}{Hinton et~al\mbox{.}}{2012}]%
        {hinton2012improving}
\bibfield{author}{\bibinfo{person}{Geoffrey~E Hinton}, \bibinfo{person}{Nitish
  Srivastava}, \bibinfo{person}{Alex Krizhevsky}, \bibinfo{person}{Ilya
  Sutskever}, {and} \bibinfo{person}{Ruslan~R Salakhutdinov}.}
  \bibinfo{year}{2012}\natexlab{}.
\newblock \showarticletitle{Improving neural networks by preventing
  co-adaptation of feature detectors}.
\newblock \bibinfo{journal}{\emph{arXiv preprint arXiv:1207.0580}}
  (\bibinfo{year}{2012}).
\newblock


\bibitem[\protect\citeauthoryear{Hoehfeld and Fahlman}{Hoehfeld and
  Fahlman}{1991}]%
        {hoehfeld1991learning}
\bibfield{author}{\bibinfo{person}{Markus Hoehfeld} {and}
  \bibinfo{person}{Scott~E Fahlman}.} \bibinfo{year}{1991}\natexlab{}.
\newblock \bibinfo{booktitle}{\emph{Learning with limited numerical precision
  using the cascade-correlation algorithm}}.
\newblock \bibinfo{publisher}{Citeseer}.
\newblock


\bibitem[\protect\citeauthoryear{Holt and Hwang}{Holt and Hwang}{1991}]%
        {holt1991finite}
\bibfield{author}{\bibinfo{person}{JL Holt} {and} \bibinfo{person}{Jenq-Neng
  Hwang}.} \bibinfo{year}{1991}\natexlab{}.
\newblock \showarticletitle{Finite precision error analysis for neural network
  learning}. In \bibinfo{booktitle}{\emph{Proceedings of the First
  International Forum on Applications of Neural Networks to Power Systems}}.
  IEEE, \bibinfo{pages}{237--241}.
\newblock


\bibitem[\protect\citeauthoryear{Howard, Sandler, Chu, Chen, Chen, Tan, Wang,
  Zhu, Pang, Vasudevan, et~al\mbox{.}}{Howard et~al\mbox{.}}{2019}]%
        {howard2019searching}
\bibfield{author}{\bibinfo{person}{Andrew Howard}, \bibinfo{person}{Mark
  Sandler}, \bibinfo{person}{Grace Chu}, \bibinfo{person}{Liang-Chieh Chen},
  \bibinfo{person}{Bo Chen}, \bibinfo{person}{Mingxing Tan},
  \bibinfo{person}{Weijun Wang}, \bibinfo{person}{Yukun Zhu},
  \bibinfo{person}{Ruoming Pang}, \bibinfo{person}{Vijay Vasudevan},
  {et~al\mbox{.}}} \bibinfo{year}{2019}\natexlab{}.
\newblock \showarticletitle{Searching for mobilenetv3}. In
  \bibinfo{booktitle}{\emph{Proceedings of the IEEE International Conference on
  Computer Vision}}. \bibinfo{pages}{1314--1324}.
\newblock


\bibitem[\protect\citeauthoryear{Howard, Zhu, Chen, Kalenichenko, Wang, Weyand,
  Andreetto, and Adam}{Howard et~al\mbox{.}}{2017}]%
        {howard2017mobilenets}
\bibfield{author}{\bibinfo{person}{Andrew~G Howard}, \bibinfo{person}{Menglong
  Zhu}, \bibinfo{person}{Bo Chen}, \bibinfo{person}{Dmitry Kalenichenko},
  \bibinfo{person}{Weijun Wang}, \bibinfo{person}{Tobias Weyand},
  \bibinfo{person}{Marco Andreetto}, {and} \bibinfo{person}{Hartwig Adam}.}
  \bibinfo{year}{2017}\natexlab{}.
\newblock \showarticletitle{Mobilenets: Efficient convolutional neural networks
  for mobile vision applications}.
\newblock \bibinfo{journal}{\emph{arXiv preprint arXiv:1704.04861}}
  (\bibinfo{year}{2017}).
\newblock


\bibitem[\protect\citeauthoryear{Hsieh, Si, and Dhillon}{Hsieh
  et~al\mbox{.}}{2014}]%
        {hsieh2014fast}
\bibfield{author}{\bibinfo{person}{Cho-Jui Hsieh}, \bibinfo{person}{Si Si},
  {and} \bibinfo{person}{Inderjit~S Dhillon}.} \bibinfo{year}{2014}\natexlab{}.
\newblock \showarticletitle{Fast prediction for large-scale kernel machines}.
  In \bibinfo{booktitle}{\emph{Advances in Neural Information Processing
  Systems}}. \bibinfo{pages}{3689--3697}.
\newblock


\bibitem[\protect\citeauthoryear{Hu, Tarakji, Raheja, Phillips, Wang, and
  Mohomed}{Hu et~al\mbox{.}}{2019}]%
        {hu2019deephome}
\bibfield{author}{\bibinfo{person}{Zhiming Hu}, \bibinfo{person}{Ahmad~Bisher
  Tarakji}, \bibinfo{person}{Vishal Raheja}, \bibinfo{person}{Caleb Phillips},
  \bibinfo{person}{Teng Wang}, {and} \bibinfo{person}{Iqbal Mohomed}.}
  \bibinfo{year}{2019}\natexlab{}.
\newblock \showarticletitle{Deephome: Distributed inference with heterogeneous
  devices in the edge}. In \bibinfo{booktitle}{\emph{The 3rd International
  Workshop on Deep Learning for Mobile Systems and Applications}}.
  \bibinfo{pages}{13--18}.
\newblock


\bibitem[\protect\citeauthoryear{Huang, Chen, Li, Wu, van~der Maaten, and
  Weinberger}{Huang et~al\mbox{.}}{2018a}]%
        {huang2017multi}
\bibfield{author}{\bibinfo{person}{Gao Huang}, \bibinfo{person}{Danlu Chen},
  \bibinfo{person}{Tianhong Li}, \bibinfo{person}{Felix Wu},
  \bibinfo{person}{Laurens van~der Maaten}, {and} \bibinfo{person}{Kilian~Q
  Weinberger}.} \bibinfo{year}{2018}\natexlab{a}.
\newblock \showarticletitle{Multi-scale dense networks for resource efficient
  image classification}. In \bibinfo{booktitle}{\emph{ICLR}}.
\newblock


\bibitem[\protect\citeauthoryear{Huang, Liu, Van~der Maaten, and
  Weinberger}{Huang et~al\mbox{.}}{2018b}]%
        {huang2018condensenet}
\bibfield{author}{\bibinfo{person}{Gao Huang}, \bibinfo{person}{Shichen Liu},
  \bibinfo{person}{Laurens Van~der Maaten}, {and} \bibinfo{person}{Kilian~Q
  Weinberger}.} \bibinfo{year}{2018}\natexlab{b}.
\newblock \showarticletitle{Condensenet: An efficient densenet using learned
  group convolutions}.
\newblock \bibinfo{journal}{\emph{CVPR}} \bibinfo{volume}{3},
  \bibinfo{number}{12} (\bibinfo{year}{2018}), \bibinfo{pages}{11}.
\newblock


\bibitem[\protect\citeauthoryear{Hubara, Courbariaux, Soudry, El-Yaniv, and
  Bengio}{Hubara et~al\mbox{.}}{2017}]%
        {hubara2017quantized}
\bibfield{author}{\bibinfo{person}{Itay Hubara}, \bibinfo{person}{Matthieu
  Courbariaux}, \bibinfo{person}{Daniel Soudry}, \bibinfo{person}{Ran
  El-Yaniv}, {and} \bibinfo{person}{Yoshua Bengio}.}
  \bibinfo{year}{2017}\natexlab{}.
\newblock \showarticletitle{Quantized neural networks: Training neural networks
  with low precision weights and activations}.
\newblock \bibinfo{journal}{\emph{The Journal of Machine Learning Research}}
  \bibinfo{volume}{18}, \bibinfo{number}{1} (\bibinfo{year}{2017}),
  \bibinfo{pages}{6869--6898}.
\newblock


\bibitem[\protect\citeauthoryear{Iandola, Han, Moskewicz, Ashraf, Dally, and
  Keutzer}{Iandola et~al\mbox{.}}{2016}]%
        {iandola2016squeezenet}
\bibfield{author}{\bibinfo{person}{Forrest~N Iandola}, \bibinfo{person}{Song
  Han}, \bibinfo{person}{Matthew~W Moskewicz}, \bibinfo{person}{Khalid Ashraf},
  \bibinfo{person}{William~J Dally}, {and} \bibinfo{person}{Kurt Keutzer}.}
  \bibinfo{year}{2016}\natexlab{}.
\newblock \showarticletitle{Squeezenet: Alexnet-level accuracy with 50x fewer
  parameters and< 0.5 mb model size}.
\newblock \bibinfo{journal}{\emph{arXiv preprint arXiv:1602.07360}}
  (\bibinfo{year}{2016}).
\newblock


\bibitem[\protect\citeauthoryear{Ignatov, Timofte, Szczepaniak, Chou, Wang, Wu,
  Hartley, and Van~Gool}{Ignatov et~al\mbox{.}}{2018}]%
        {ignatov2018ai}
\bibfield{author}{\bibinfo{person}{Andrey Ignatov}, \bibinfo{person}{Radu
  Timofte}, \bibinfo{person}{Przemyslaw Szczepaniak}, \bibinfo{person}{William
  Chou}, \bibinfo{person}{Ke Wang}, \bibinfo{person}{Max Wu},
  \bibinfo{person}{Tim Hartley}, {and} \bibinfo{person}{Luc Van~Gool}.}
  \bibinfo{year}{2018}\natexlab{}.
\newblock \showarticletitle{AI Benchmark: Running Deep Neural Networks on
  Android Smartphones}.
\newblock \bibinfo{journal}{\emph{arXiv preprint arXiv:1810.01109}}
  (\bibinfo{year}{2018}).
\newblock


\bibitem[\protect\citeauthoryear{Ito, Hatano, Sumita, Yabe, Fukunaga, Kakimura,
  and Kawarabayashi}{Ito et~al\mbox{.}}{2017}]%
        {ito2017efficient}
\bibfield{author}{\bibinfo{person}{Shinji Ito}, \bibinfo{person}{Daisuke
  Hatano}, \bibinfo{person}{Hanna Sumita}, \bibinfo{person}{Akihiro Yabe},
  \bibinfo{person}{Takuro Fukunaga}, \bibinfo{person}{Naonori Kakimura}, {and}
  \bibinfo{person}{Ken-Ichi Kawarabayashi}.} \bibinfo{year}{2017}\natexlab{}.
\newblock \showarticletitle{Efficient Sublinear-Regret Algorithms for Online
  Sparse Linear Regression with Limited Observation}. In
  \bibinfo{booktitle}{\emph{Advances in Neural Information Processing
  Systems}}. \bibinfo{pages}{4099--4108}.
\newblock


\bibitem[\protect\citeauthoryear{Ito, Hatano, Sumita, Yabe, Fukunaga, Kakimura,
  and Kawarabayashi}{Ito et~al\mbox{.}}{2018}]%
        {ito2018online}
\bibfield{author}{\bibinfo{person}{Shinji Ito}, \bibinfo{person}{Daisuke
  Hatano}, \bibinfo{person}{Hanna Sumita}, \bibinfo{person}{Akihiro Yabe},
  \bibinfo{person}{Takuro Fukunaga}, \bibinfo{person}{Naonori Kakimura}, {and}
  \bibinfo{person}{Ken-Ichi Kawarabayashi}.} \bibinfo{year}{2018}\natexlab{}.
\newblock \showarticletitle{Online Regression with Partial Information:
  Generalization and Linear Projection}. In
  \bibinfo{booktitle}{\emph{International Conference on Artificial Intelligence
  and Statistics}}. \bibinfo{pages}{1599--1607}.
\newblock


\bibitem[\protect\citeauthoryear{Jacob, Kligys, Chen, Zhu, Tang, Howard, Adam,
  and Kalenichenko}{Jacob et~al\mbox{.}}{2018}]%
        {jacob2018quantization}
\bibfield{author}{\bibinfo{person}{Benoit Jacob}, \bibinfo{person}{Skirmantas
  Kligys}, \bibinfo{person}{Bo Chen}, \bibinfo{person}{Menglong Zhu},
  \bibinfo{person}{Matthew Tang}, \bibinfo{person}{Andrew Howard},
  \bibinfo{person}{Hartwig Adam}, {and} \bibinfo{person}{Dmitry Kalenichenko}.}
  \bibinfo{year}{2018}\natexlab{}.
\newblock \showarticletitle{Quantization and training of neural networks for
  efficient integer-arithmetic-only inference}. In
  \bibinfo{booktitle}{\emph{Proceedings of the IEEE Conference on Computer
  Vision and Pattern Recognition}}. \bibinfo{pages}{2704--2713}.
\newblock


\bibitem[\protect\citeauthoryear{Jiao, Zhang, Liu, Yang, Li, Feng, and Qu}{Jiao
  et~al\mbox{.}}{2019}]%
        {jiao2019survey}
\bibfield{author}{\bibinfo{person}{Licheng Jiao}, \bibinfo{person}{Fan Zhang},
  \bibinfo{person}{Fang Liu}, \bibinfo{person}{Shuyuan Yang},
  \bibinfo{person}{Lingling Li}, \bibinfo{person}{Zhixi Feng}, {and}
  \bibinfo{person}{Rong Qu}.} \bibinfo{year}{2019}\natexlab{}.
\newblock \showarticletitle{A Survey of Deep Learning-Based Object Detection}.
\newblock \bibinfo{journal}{\emph{IEEE Access}}  \bibinfo{volume}{7}
  (\bibinfo{year}{2019}), \bibinfo{pages}{128837--128868}.
\newblock


\bibitem[\protect\citeauthoryear{Jose, Goyal, Aggrwal, and Varma}{Jose
  et~al\mbox{.}}{2013}]%
        {jose2013local}
\bibfield{author}{\bibinfo{person}{Cijo Jose}, \bibinfo{person}{Prasoon Goyal},
  \bibinfo{person}{Parv Aggrwal}, {and} \bibinfo{person}{Manik Varma}.}
  \bibinfo{year}{2013}\natexlab{}.
\newblock \showarticletitle{Local deep kernel learning for efficient non-linear
  svm prediction}. In \bibinfo{booktitle}{\emph{International conference on
  machine learning}}. \bibinfo{pages}{486--494}.
\newblock


\bibitem[\protect\citeauthoryear{Judd, Albericio, Hetherington, Aamodt, Jerger,
  Urtasun, and Moshovos}{Judd et~al\mbox{.}}{2015}]%
        {judd2015reduced}
\bibfield{author}{\bibinfo{person}{Patrick Judd}, \bibinfo{person}{Jorge
  Albericio}, \bibinfo{person}{Tayler Hetherington}, \bibinfo{person}{Tor
  Aamodt}, \bibinfo{person}{Natalie~Enright Jerger}, \bibinfo{person}{Raquel
  Urtasun}, {and} \bibinfo{person}{Andreas Moshovos}.}
  \bibinfo{year}{2015}\natexlab{}.
\newblock \showarticletitle{Reduced-precision strategies for bounded memory in
  deep neural nets}.
\newblock \bibinfo{journal}{\emph{arXiv preprint arXiv:1511.05236}}
  (\bibinfo{year}{2015}).
\newblock


\bibitem[\protect\citeauthoryear{Kearns}{Kearns}{1990}]%
        {kearns1990computational}
\bibfield{author}{\bibinfo{person}{Michael~J Kearns}.}
  \bibinfo{year}{1990}\natexlab{}.
\newblock \bibinfo{booktitle}{\emph{The computational complexity of machine
  learning}}.
\newblock \bibinfo{publisher}{MIT press}.
\newblock


\bibitem[\protect\citeauthoryear{Kim and Smaragdis}{Kim and Smaragdis}{2016}]%
        {kim2016bitwise}
\bibfield{author}{\bibinfo{person}{Minje Kim} {and} \bibinfo{person}{Paris
  Smaragdis}.} \bibinfo{year}{2016}\natexlab{}.
\newblock \showarticletitle{Bitwise neural networks}.
\newblock \bibinfo{journal}{\emph{arXiv preprint arXiv:1601.06071}}
  (\bibinfo{year}{2016}).
\newblock


\bibitem[\protect\citeauthoryear{Kollmann, Riemschneider, and Zeidler}{Kollmann
  et~al\mbox{.}}{1996}]%
        {kollmann1996chip}
\bibfield{author}{\bibinfo{person}{Kuno Kollmann}, \bibinfo{person}{K-R
  Riemschneider}, {and} \bibinfo{person}{Hans~Christoph Zeidler}.}
  \bibinfo{year}{1996}\natexlab{}.
\newblock \showarticletitle{On-chip backpropagation training using parallel
  stochastic bit streams}. In \bibinfo{booktitle}{\emph{Proceedings of Fifth
  International Conference on Microelectronics for Neural Networks}}. IEEE,
  \bibinfo{pages}{149--156}.
\newblock


\bibitem[\protect\citeauthoryear{Krizhevsky, Sutskever, and Hinton}{Krizhevsky
  et~al\mbox{.}}{2012}]%
        {krizhevsky2012imagenet}
\bibfield{author}{\bibinfo{person}{Alex Krizhevsky}, \bibinfo{person}{Ilya
  Sutskever}, {and} \bibinfo{person}{Geoffrey~E Hinton}.}
  \bibinfo{year}{2012}\natexlab{}.
\newblock \showarticletitle{Imagenet classification with deep convolutional
  neural networks}. In \bibinfo{booktitle}{\emph{Advances in neural information
  processing systems}}. \bibinfo{pages}{1097--1105}.
\newblock


\bibitem[\protect\citeauthoryear{Kulkarni and Sinha}{Kulkarni and
  Sinha}{2012}]%
        {kulkarni2012pruning}
\bibfield{author}{\bibinfo{person}{Vrushali~Y Kulkarni} {and}
  \bibinfo{person}{Pradeep~K Sinha}.} \bibinfo{year}{2012}\natexlab{}.
\newblock \showarticletitle{Pruning of random forest classifiers: A survey and
  future directions}. In \bibinfo{booktitle}{\emph{Data Science \& Engineering
  (ICDSE), 2012 International Conference on}}. IEEE, \bibinfo{pages}{64--68}.
\newblock


\bibitem[\protect\citeauthoryear{Kumar, Goyal, and Varma}{Kumar
  et~al\mbox{.}}{2017}]%
        {kumar2017resource}
\bibfield{author}{\bibinfo{person}{Ashish Kumar}, \bibinfo{person}{Saurabh
  Goyal}, {and} \bibinfo{person}{Manik Varma}.}
  \bibinfo{year}{2017}\natexlab{}.
\newblock \showarticletitle{Resource-efficient Machine Learning in 2 KB RAM for
  the Internet of Things}. In \bibinfo{booktitle}{\emph{International
  Conference on Machine Learning}}. \bibinfo{pages}{1935--1944}.
\newblock


\bibitem[\protect\citeauthoryear{Kusner, Tyree, Weinberger, and Agrawal}{Kusner
  et~al\mbox{.}}{2014}]%
        {kusner2014stochastic}
\bibfield{author}{\bibinfo{person}{Matt Kusner}, \bibinfo{person}{Stephen
  Tyree}, \bibinfo{person}{Kilian Weinberger}, {and} \bibinfo{person}{Kunal
  Agrawal}.} \bibinfo{year}{2014}\natexlab{}.
\newblock \showarticletitle{Stochastic neighbor compression}. In
  \bibinfo{booktitle}{\emph{International Conference on Machine Learning}}.
  \bibinfo{pages}{622--630}.
\newblock


\bibitem[\protect\citeauthoryear{Lakshmanan and Chandrasekaran}{Lakshmanan and
  Chandrasekaran}{1979}]%
        {lakshmanan1979compound}
\bibfield{author}{\bibinfo{person}{KB Lakshmanan} {and} \bibinfo{person}{B
  Chandrasekaran}.} \bibinfo{year}{1979}\natexlab{}.
\newblock \showarticletitle{Compound hypothesis testing with finite memory}.
\newblock \bibinfo{journal}{\emph{Information and Control}}
  \bibinfo{volume}{40}, \bibinfo{number}{2} (\bibinfo{year}{1979}),
  \bibinfo{pages}{223--233}.
\newblock


\bibitem[\protect\citeauthoryear{Langford, Li, and Zhang}{Langford
  et~al\mbox{.}}{2009}]%
        {langford2009sparse}
\bibfield{author}{\bibinfo{person}{John Langford}, \bibinfo{person}{Lihong Li},
  {and} \bibinfo{person}{Tong Zhang}.} \bibinfo{year}{2009}\natexlab{}.
\newblock \showarticletitle{Sparse online learning via truncated gradient}.
\newblock \bibinfo{journal}{\emph{Journal of Machine Learning Research}}
  \bibinfo{volume}{10}, \bibinfo{number}{Mar} (\bibinfo{year}{2009}),
  \bibinfo{pages}{777--801}.
\newblock


\bibitem[\protect\citeauthoryear{Le, Sarl{\'o}s, and Smola}{Le
  et~al\mbox{.}}{2013}]%
        {le2013fastfood}
\bibfield{author}{\bibinfo{person}{Quoc Le}, \bibinfo{person}{Tam{\'a}s
  Sarl{\'o}s}, {and} \bibinfo{person}{Alex Smola}.}
  \bibinfo{year}{2013}\natexlab{}.
\newblock \showarticletitle{Fastfood-approximating kernel expansions in
  loglinear time}. In \bibinfo{booktitle}{\emph{Proceedings of the
  international conference on machine learning}}, Vol.~\bibinfo{volume}{85}.
\newblock


\bibitem[\protect\citeauthoryear{Learn}{Learn}{[n.d.]}]%
        {treescikit}
\bibfield{author}{\bibinfo{person}{Scikit Learn}.}
  \bibinfo{year}{[n.d.]}\natexlab{}.
\newblock \bibinfo{booktitle}{\emph{Decision Tree}}.
\newblock
\urldef\tempurl%
\url{http://scikit-learn.org/stable/modules/tree.html#complexity}
\showURL{%
\tempurl}


\bibitem[\protect\citeauthoryear{Lebedev and Lempitsky}{Lebedev and
  Lempitsky}{2016}]%
        {lebedev2016fast}
\bibfield{author}{\bibinfo{person}{Vadim Lebedev} {and} \bibinfo{person}{Victor
  Lempitsky}.} \bibinfo{year}{2016}\natexlab{}.
\newblock \showarticletitle{Fast convnets using group-wise brain damage}. In
  \bibinfo{booktitle}{\emph{Proceedings of the IEEE Conference on Computer
  Vision and Pattern Recognition}}. \bibinfo{pages}{2554--2564}.
\newblock


\bibitem[\protect\citeauthoryear{Lee, Stanley, Spanias, and
  Tepedelenlioglu}{Lee et~al\mbox{.}}{2016}]%
        {lee2016integrating}
\bibfield{author}{\bibinfo{person}{Jongmin Lee}, \bibinfo{person}{Michael
  Stanley}, \bibinfo{person}{Andreas Spanias}, {and} \bibinfo{person}{Cihan
  Tepedelenlioglu}.} \bibinfo{year}{2016}\natexlab{}.
\newblock \showarticletitle{Integrating machine learning in embedded sensor
  systems for internet-of-things applications}. In
  \bibinfo{booktitle}{\emph{Signal Processing and Information Technology
  (ISSPIT), 2016 IEEE International Symposium on}}. IEEE,
  \bibinfo{pages}{290--294}.
\newblock


\bibitem[\protect\citeauthoryear{Lee and Nirjon}{Lee and Nirjon}{2019}]%
        {lee2019neuro}
\bibfield{author}{\bibinfo{person}{Seulki Lee} {and} \bibinfo{person}{Shahriar
  Nirjon}.} \bibinfo{year}{2019}\natexlab{}.
\newblock \showarticletitle{Neuro. ZERO: a zero-energy neural network
  accelerator for embedded sensing and inference systems}. In
  \bibinfo{booktitle}{\emph{Proceedings of the 17th Conference on Embedded
  Networked Sensor Systems}}. \bibinfo{pages}{138--152}.
\newblock


\bibitem[\protect\citeauthoryear{Leighton and Rivest}{Leighton and
  Rivest}{1986}]%
        {leighton1986estimating}
\bibfield{author}{\bibinfo{person}{F Leighton} {and} \bibinfo{person}{Ronald
  Rivest}.} \bibinfo{year}{1986}\natexlab{}.
\newblock \showarticletitle{Estimating a probability using finite memory}.
\newblock \bibinfo{journal}{\emph{IEEE Transactions on Information Theory}}
  \bibinfo{volume}{32}, \bibinfo{number}{6} (\bibinfo{year}{1986}),
  \bibinfo{pages}{733--742}.
\newblock


\bibitem[\protect\citeauthoryear{Li, Lin, and Lu}{Li et~al\mbox{.}}{2016a}]%
        {li2016rivalry}
\bibfield{author}{\bibinfo{person}{Chun-Liang Li}, \bibinfo{person}{Hsuan-Tien
  Lin}, {and} \bibinfo{person}{Chi-Jen Lu}.} \bibinfo{year}{2016}\natexlab{a}.
\newblock \showarticletitle{Rivalry of two families of algorithms for
  memory-restricted streaming pca}. In \bibinfo{booktitle}{\emph{Artificial
  Intelligence and Statistics}}. \bibinfo{pages}{473--481}.
\newblock


\bibitem[\protect\citeauthoryear{Li, De, Xu, Studer, Samet, and Goldstein}{Li
  et~al\mbox{.}}{2017}]%
        {li2017training}
\bibfield{author}{\bibinfo{person}{Hao Li}, \bibinfo{person}{Soham De},
  \bibinfo{person}{Zheng Xu}, \bibinfo{person}{Christoph Studer},
  \bibinfo{person}{Hanan Samet}, {and} \bibinfo{person}{Tom Goldstein}.}
  \bibinfo{year}{2017}\natexlab{}.
\newblock \showarticletitle{Training quantized nets: A deeper understanding}.
  In \bibinfo{booktitle}{\emph{Advances in Neural Information Processing
  Systems}}. \bibinfo{pages}{5811--5821}.
\newblock


\bibitem[\protect\citeauthoryear{Li, Sahu, Talwalkar, and Smith}{Li
  et~al\mbox{.}}{2020}]%
        {li2020federated}
\bibfield{author}{\bibinfo{person}{Tian Li}, \bibinfo{person}{Anit~Kumar Sahu},
  \bibinfo{person}{Ameet Talwalkar}, {and} \bibinfo{person}{Virginia Smith}.}
  \bibinfo{year}{2020}\natexlab{}.
\newblock \showarticletitle{Federated learning: Challenges, methods, and future
  directions}.
\newblock \bibinfo{journal}{\emph{IEEE Signal Processing Magazine}}
  \bibinfo{volume}{37}, \bibinfo{number}{3} (\bibinfo{year}{2020}),
  \bibinfo{pages}{50--60}.
\newblock


\bibitem[\protect\citeauthoryear{Li, Qin, Yang, and Liu}{Li
  et~al\mbox{.}}{2016b}]%
        {li2016lightrnn}
\bibfield{author}{\bibinfo{person}{Xiang Li}, \bibinfo{person}{Tao Qin},
  \bibinfo{person}{Jian Yang}, {and} \bibinfo{person}{Tieyan Liu}.}
  \bibinfo{year}{2016}\natexlab{b}.
\newblock \showarticletitle{LightRNN: Memory and computation-efficient
  recurrent neural networks}. In \bibinfo{booktitle}{\emph{Advances in Neural
  Information Processing Systems}}. \bibinfo{pages}{4385--4393}.
\newblock


\bibitem[\protect\citeauthoryear{Li and De~Sa}{Li and De~Sa}{2019}]%
        {li2019dimension}
\bibfield{author}{\bibinfo{person}{Zheng Li} {and}
  \bibinfo{person}{Christopher~M De~Sa}.} \bibinfo{year}{2019}\natexlab{}.
\newblock \showarticletitle{Dimension-Free Bounds for Low-Precision Training}.
  In \bibinfo{booktitle}{\emph{Advances in Neural Information Processing
  Systems}}. \bibinfo{pages}{11728--11738}.
\newblock


\bibitem[\protect\citeauthoryear{Lim, Loh, and Shih}{Lim et~al\mbox{.}}{2000}]%
        {lim2000comparison}
\bibfield{author}{\bibinfo{person}{Tjen-Sien Lim}, \bibinfo{person}{Wei-Yin
  Loh}, {and} \bibinfo{person}{Yu-Shan Shih}.} \bibinfo{year}{2000}\natexlab{}.
\newblock \showarticletitle{A comparison of prediction accuracy, complexity,
  and training time of thirty-three old and new classification algorithms}.
\newblock \bibinfo{journal}{\emph{Machine learning}} \bibinfo{volume}{40},
  \bibinfo{number}{3} (\bibinfo{year}{2000}), \bibinfo{pages}{203--228}.
\newblock


\bibitem[\protect\citeauthoryear{Lim, Luong, Hoang, Jiao, Liang, Yang, Niyato,
  and Miao}{Lim et~al\mbox{.}}{2020}]%
        {lim2020federated}
\bibfield{author}{\bibinfo{person}{Wei Yang~Bryan Lim},
  \bibinfo{person}{Nguyen~Cong Luong}, \bibinfo{person}{Dinh~Thai Hoang},
  \bibinfo{person}{Yutao Jiao}, \bibinfo{person}{Ying-Chang Liang},
  \bibinfo{person}{Qiang Yang}, \bibinfo{person}{Dusit Niyato}, {and}
  \bibinfo{person}{Chunyan Miao}.} \bibinfo{year}{2020}\natexlab{}.
\newblock \showarticletitle{Federated learning in mobile edge networks: A
  comprehensive survey}.
\newblock \bibinfo{journal}{\emph{IEEE Communications Surveys \& Tutorials}}
  (\bibinfo{year}{2020}).
\newblock


\bibitem[\protect\citeauthoryear{Lin, Han, Mao, Wang, and Dally}{Lin
  et~al\mbox{.}}{2017}]%
        {lin2017deep}
\bibfield{author}{\bibinfo{person}{Yujun Lin}, \bibinfo{person}{Song Han},
  \bibinfo{person}{Huizi Mao}, \bibinfo{person}{Yu Wang}, {and}
  \bibinfo{person}{William~J Dally}.} \bibinfo{year}{2017}\natexlab{}.
\newblock \showarticletitle{Deep gradient compression: Reducing the
  communication bandwidth for distributed training}.
\newblock \bibinfo{journal}{\emph{arXiv preprint arXiv:1712.01887}}
  (\bibinfo{year}{2017}).
\newblock


\bibitem[\protect\citeauthoryear{Lin, Courbariaux, Memisevic, and Bengio}{Lin
  et~al\mbox{.}}{2015}]%
        {lin2015neural}
\bibfield{author}{\bibinfo{person}{Zhouhan Lin}, \bibinfo{person}{Matthieu
  Courbariaux}, \bibinfo{person}{Roland Memisevic}, {and}
  \bibinfo{person}{Yoshua Bengio}.} \bibinfo{year}{2015}\natexlab{}.
\newblock \showarticletitle{Neural networks with few multiplications}.
\newblock \bibinfo{journal}{\emph{arXiv preprint arXiv:1510.03009}}
  (\bibinfo{year}{2015}).
\newblock


\bibitem[\protect\citeauthoryear{Lin, Gu, and Chakraborty}{Lin
  et~al\mbox{.}}{2010}]%
        {lin2010tuning}
\bibfield{author}{\bibinfo{person}{Ziheng Lin}, \bibinfo{person}{Yan Gu}, {and}
  \bibinfo{person}{Samarjit Chakraborty}.} \bibinfo{year}{2010}\natexlab{}.
\newblock \showarticletitle{Tuning Machine-Learning Algorithms for
  Battery-Operated Portable Devices}. In \bibinfo{booktitle}{\emph{Asia
  Information Retrieval Symposium}}. Springer, \bibinfo{pages}{502--513}.
\newblock


\bibitem[\protect\citeauthoryear{Liu, Tripathi, Kurup, and Shah}{Liu
  et~al\mbox{.}}{2020}]%
        {liu2020pruning}
\bibfield{author}{\bibinfo{person}{Jiayi Liu}, \bibinfo{person}{Samarth
  Tripathi}, \bibinfo{person}{Unmesh Kurup}, {and} \bibinfo{person}{Mohak
  Shah}.} \bibinfo{year}{2020}\natexlab{}.
\newblock \showarticletitle{Pruning Algorithms to Accelerate Convolutional
  Neural Networks for Edge Applications: A Survey}.
\newblock \bibinfo{journal}{\emph{arXiv preprint arXiv:2005.04275}}
  (\bibinfo{year}{2020}).
\newblock


\bibitem[\protect\citeauthoryear{Lu, Rallapalli, Chan, and La~Porta}{Lu
  et~al\mbox{.}}{2017}]%
        {lu2017modeling}
\bibfield{author}{\bibinfo{person}{Zongqing Lu}, \bibinfo{person}{Swati
  Rallapalli}, \bibinfo{person}{Kevin Chan}, {and} \bibinfo{person}{Thomas
  La~Porta}.} \bibinfo{year}{2017}\natexlab{}.
\newblock \showarticletitle{Modeling the resource requirements of convolutional
  neural networks on mobile devices}. In \bibinfo{booktitle}{\emph{Proceedings
  of the 2017 ACM on Multimedia Conference}}. ACM, \bibinfo{pages}{1663--1671}.
\newblock


\bibitem[\protect\citeauthoryear{Lucic}{Lucic}{2017}]%
        {lucic2017computational}
\bibfield{author}{\bibinfo{person}{Mario Lucic}.}
  \bibinfo{year}{2017}\natexlab{}.
\newblock \emph{\bibinfo{title}{Computational and Statistical Tradeoffs via
  Data Summarization}}.
\newblock \bibinfo{thesistype}{Ph.D. Dissertation}. \bibinfo{school}{ETH
  Zurich}.
\newblock


\bibitem[\protect\citeauthoryear{Magoulas, Vrahatis, and Androulakis}{Magoulas
  et~al\mbox{.}}{1996}]%
        {magoulas1996new}
\bibfield{author}{\bibinfo{person}{GD Magoulas}, \bibinfo{person}{MN Vrahatis},
  {and} \bibinfo{person}{GS Androulakis}.} \bibinfo{year}{1996}\natexlab{}.
\newblock \showarticletitle{A new method in neural network supervised training
  with imprecision}. In \bibinfo{booktitle}{\emph{Proceedings of Third
  International Conference on Electronics, Circuits, and Systems}},
  Vol.~\bibinfo{volume}{1}. IEEE, \bibinfo{pages}{287--290}.
\newblock


\bibitem[\protect\citeauthoryear{Marco, Taylor, Wang, and Elkhatib}{Marco
  et~al\mbox{.}}{2020}]%
        {marco2020optimizing}
\bibfield{author}{\bibinfo{person}{Vicent~Sanz Marco}, \bibinfo{person}{Ben
  Taylor}, \bibinfo{person}{Zheng Wang}, {and} \bibinfo{person}{Yehia
  Elkhatib}.} \bibinfo{year}{2020}\natexlab{}.
\newblock \showarticletitle{Optimizing deep learning inference on embedded
  systems through adaptive model selection}.
\newblock \bibinfo{journal}{\emph{ACM Transactions on Embedded Computing
  Systems (TECS)}} \bibinfo{volume}{19}, \bibinfo{number}{1}
  (\bibinfo{year}{2020}), \bibinfo{pages}{1--28}.
\newblock


\bibitem[\protect\citeauthoryear{Marculescu, Stamoulis, and Cai}{Marculescu
  et~al\mbox{.}}{2018}]%
        {marculescu2018hardware}
\bibfield{author}{\bibinfo{person}{Diana Marculescu},
  \bibinfo{person}{Dimitrios Stamoulis}, {and} \bibinfo{person}{Ermao Cai}.}
  \bibinfo{year}{2018}\natexlab{}.
\newblock \showarticletitle{Hardware-Aware Machine Learning: Modeling and
  Optimization}.
\newblock \bibinfo{journal}{\emph{arXiv preprint arXiv:1809.05476}}
  (\bibinfo{year}{2018}).
\newblock


\bibitem[\protect\citeauthoryear{Mayekar and Tyagi}{Mayekar and Tyagi}{2019}]%
        {mayekar2019ratq}
\bibfield{author}{\bibinfo{person}{Prathamesh Mayekar} {and}
  \bibinfo{person}{Himanshu Tyagi}.} \bibinfo{year}{2019}\natexlab{}.
\newblock \bibinfo{title}{RATQ: A Universal Fixed-Length Quantizer for
  Stochastic Optimization}.
\newblock
\newblock
\showeprint[arxiv]{cs.LG/1908.08200}


\bibitem[\protect\citeauthoryear{McInerney, Constantinides, and
  Kerrigan}{McInerney et~al\mbox{.}}{2018}]%
        {mcinerney2018survey}
\bibfield{author}{\bibinfo{person}{Ian McInerney}, \bibinfo{person}{George~A
  Constantinides}, {and} \bibinfo{person}{Eric~C Kerrigan}.}
  \bibinfo{year}{2018}\natexlab{}.
\newblock \showarticletitle{A survey of the implementation of linear model
  predictive control on fpgas}.
\newblock \bibinfo{journal}{\emph{IFAC-PapersOnLine}} \bibinfo{volume}{51},
  \bibinfo{number}{20} (\bibinfo{year}{2018}), \bibinfo{pages}{381--387}.
\newblock


\bibitem[\protect\citeauthoryear{McMahan, Moore, Ramage, Hampson, and
  y~Arcas}{McMahan et~al\mbox{.}}{2017}]%
        {mcmahan2017communication}
\bibfield{author}{\bibinfo{person}{Brendan McMahan}, \bibinfo{person}{Eider
  Moore}, \bibinfo{person}{Daniel Ramage}, \bibinfo{person}{Seth Hampson},
  {and} \bibinfo{person}{Blaise~Aguera y Arcas}.}
  \bibinfo{year}{2017}\natexlab{}.
\newblock \showarticletitle{Communication-Efficient Learning of Deep Networks
  from Decentralized Data}. In \bibinfo{booktitle}{\emph{Artificial
  Intelligence and Statistics}}. \bibinfo{pages}{1273--1282}.
\newblock


\bibitem[\protect\citeauthoryear{McMahan, Moore, Ramage, and y~Arcas}{McMahan
  et~al\mbox{.}}{2016}]%
        {McMahanMRA16}
\bibfield{author}{\bibinfo{person}{H.~Brendan McMahan}, \bibinfo{person}{Eider
  Moore}, \bibinfo{person}{Daniel Ramage}, {and}
  \bibinfo{person}{Blaise~Ag{\"{u}}era y Arcas}.}
  \bibinfo{year}{2016}\natexlab{}.
\newblock \showarticletitle{Federated Learning of Deep Networks using Model
  Averaging}.
\newblock \bibinfo{journal}{\emph{CoRR}}  \bibinfo{volume}{abs/1602.05629}
  (\bibinfo{year}{2016}).
\newblock
\showeprint[arxiv]{1602.05629}
\urldef\tempurl%
\url{http://arxiv.org/abs/1602.05629}
\showURL{%
\tempurl}


\bibitem[\protect\citeauthoryear{Micikevicius, Narang, Alben, Diamos, Elsen,
  Garcia, Ginsburg, Houston, Kuchaev, Venkatesh, et~al\mbox{.}}{Micikevicius
  et~al\mbox{.}}{2017}]%
        {micikevicius2017mixed}
\bibfield{author}{\bibinfo{person}{Paulius Micikevicius},
  \bibinfo{person}{Sharan Narang}, \bibinfo{person}{Jonah Alben},
  \bibinfo{person}{Gregory Diamos}, \bibinfo{person}{Erich Elsen},
  \bibinfo{person}{David Garcia}, \bibinfo{person}{Boris Ginsburg},
  \bibinfo{person}{Michael Houston}, \bibinfo{person}{Oleksii Kuchaev},
  \bibinfo{person}{Ganesh Venkatesh}, {et~al\mbox{.}}}
  \bibinfo{year}{2017}\natexlab{}.
\newblock \showarticletitle{Mixed precision training}.
\newblock \bibinfo{journal}{\emph{arXiv preprint arXiv:1710.03740}}
  (\bibinfo{year}{2017}).
\newblock


\bibitem[\protect\citeauthoryear{Minaee, Boykov, Porikli, Plaza, Kehtarnavaz,
  and Terzopoulos}{Minaee et~al\mbox{.}}{2020}]%
        {minaee2020image}
\bibfield{author}{\bibinfo{person}{Shervin Minaee}, \bibinfo{person}{Yuri
  Boykov}, \bibinfo{person}{Fatih Porikli}, \bibinfo{person}{Antonio Plaza},
  \bibinfo{person}{Nasser Kehtarnavaz}, {and} \bibinfo{person}{Demetri
  Terzopoulos}.} \bibinfo{year}{2020}\natexlab{}.
\newblock \showarticletitle{Image Segmentation Using Deep Learning: A Survey}.
\newblock \bibinfo{journal}{\emph{arXiv preprint arXiv:2001.05566}}
  (\bibinfo{year}{2020}).
\newblock


\bibitem[\protect\citeauthoryear{Mitliagkas, Caramanis, and Jain}{Mitliagkas
  et~al\mbox{.}}{2013}]%
        {mitliagkas2013memory}
\bibfield{author}{\bibinfo{person}{Ioannis Mitliagkas},
  \bibinfo{person}{Constantine Caramanis}, {and} \bibinfo{person}{Prateek
  Jain}.} \bibinfo{year}{2013}\natexlab{}.
\newblock \showarticletitle{Memory limited, streaming PCA}. In
  \bibinfo{booktitle}{\emph{Advances in Neural Information Processing
  Systems}}. \bibinfo{pages}{2886--2894}.
\newblock


\bibitem[\protect\citeauthoryear{Mohri, Rostamizadeh, and Talwalkar}{Mohri
  et~al\mbox{.}}{2012}]%
        {mohri2012foundations}
\bibfield{author}{\bibinfo{person}{Mehryar Mohri}, \bibinfo{person}{Afshin
  Rostamizadeh}, {and} \bibinfo{person}{Ameet Talwalkar}.}
  \bibinfo{year}{2012}\natexlab{}.
\newblock \bibinfo{booktitle}{\emph{Foundations of machine learning}}.
\newblock \bibinfo{publisher}{MIT press}.
\newblock


\bibitem[\protect\citeauthoryear{Morris}{Morris}{1978}]%
        {morris1978counting}
\bibfield{author}{\bibinfo{person}{Robert Morris}.}
  \bibinfo{year}{1978}\natexlab{}.
\newblock \showarticletitle{Counting large numbers of events in small
  registers}.
\newblock \bibinfo{journal}{\emph{Commun. ACM}} \bibinfo{volume}{21},
  \bibinfo{number}{10} (\bibinfo{year}{1978}), \bibinfo{pages}{840--842}.
\newblock


\bibitem[\protect\citeauthoryear{Moshkovitz and Moshkovitz}{Moshkovitz and
  Moshkovitz}{2017}]%
        {moshkovitz2017mixing}
\bibfield{author}{\bibinfo{person}{Dana Moshkovitz} {and}
  \bibinfo{person}{Michal Moshkovitz}.} \bibinfo{year}{2017}\natexlab{}.
\newblock \showarticletitle{Mixing implies lower bounds for space bounded
  learning}. In \bibinfo{booktitle}{\emph{Conference on Learning Theory}}.
  \bibinfo{pages}{1516--1566}.
\newblock


\bibitem[\protect\citeauthoryear{Moshkovitz and Moshkovitz}{Moshkovitz and
  Moshkovitz}{2018}]%
        {moshkovitz2018entropy}
\bibfield{author}{\bibinfo{person}{Dana Moshkovitz} {and}
  \bibinfo{person}{Michal Moshkovitz}.} \bibinfo{year}{2018}\natexlab{}.
\newblock \showarticletitle{Entropy Samplers and Strong Generic Lower Bounds
  For Space Bounded Learning}. In \bibinfo{booktitle}{\emph{LIPIcs-Leibniz
  International Proceedings in Informatics}}, Vol.~\bibinfo{volume}{94}.
  Schloss Dagstuhl-Leibniz-Zentrum fuer Informatik.
\newblock


\bibitem[\protect\citeauthoryear{Moshkovitz and Tishby}{Moshkovitz and
  Tishby}{2017}]%
        {moshkovitz2017general}
\bibfield{author}{\bibinfo{person}{Michal Moshkovitz} {and}
  \bibinfo{person}{Naftali Tishby}.} \bibinfo{year}{2017}\natexlab{}.
\newblock \showarticletitle{A General Memory-Bounded Learning Algorithm}.
\newblock \bibinfo{journal}{\emph{arXiv preprint arXiv:1712.03524}}
  (\bibinfo{year}{2017}).
\newblock


\bibitem[\protect\citeauthoryear{Murata and Suzuki}{Murata and Suzuki}{2018}]%
        {murata2018sample}
\bibfield{author}{\bibinfo{person}{Tomoya Murata} {and} \bibinfo{person}{Taiji
  Suzuki}.} \bibinfo{year}{2018}\natexlab{}.
\newblock \showarticletitle{Sample Efficient Stochastic Gradient Iterative Hard
  Thresholding Method for Stochastic Sparse Linear Regression with Limited
  Attribute Observation}. In \bibinfo{booktitle}{\emph{Advances in Neural
  Information Processing Systems}}. \bibinfo{pages}{5317--5326}.
\newblock


\bibitem[\protect\citeauthoryear{Murshed, Murphy, Hou, Khan, Ananthanarayanan,
  and Hussain}{Murshed et~al\mbox{.}}{2019}]%
        {murshed2019machine}
\bibfield{author}{\bibinfo{person}{MG Murshed}, \bibinfo{person}{Christopher
  Murphy}, \bibinfo{person}{Daqing Hou}, \bibinfo{person}{Nazar Khan},
  \bibinfo{person}{Ganesh Ananthanarayanan}, {and} \bibinfo{person}{Faraz
  Hussain}.} \bibinfo{year}{2019}\natexlab{}.
\newblock \showarticletitle{Machine learning at the network edge: A survey}.
\newblock \bibinfo{journal}{\emph{arXiv preprint arXiv:1908.00080}}
  (\bibinfo{year}{2019}).
\newblock


\bibitem[\protect\citeauthoryear{MXNet}{MXNet}{[n.d.]}]%
        {MXNetmemory}
\bibfield{author}{\bibinfo{person}{Apache MXNet}.}
  \bibinfo{year}{[n.d.]}\natexlab{}.
\newblock \bibinfo{booktitle}{\emph{Memory Cost of Deep Nets under Different
  Allocations}}.
\newblock
\urldef\tempurl%
\url{https://github.com/apache/incubator-mxnet/tree/master/example/memcost#memory-cost-of-deep-nets-under-different-allocations}
\showURL{%
\tempurl}


\bibitem[\protect\citeauthoryear{Nesterov}{Nesterov}{2013}]%
        {nesterov2013introductory}
\bibfield{author}{\bibinfo{person}{Yurii Nesterov}.}
  \bibinfo{year}{2013}\natexlab{}.
\newblock \bibinfo{booktitle}{\emph{Introductory lectures on convex
  optimization: A basic course}}. Vol.~\bibinfo{volume}{87}.
\newblock \bibinfo{publisher}{Springer Science \& Business Media}.
\newblock


\bibitem[\protect\citeauthoryear{Nielsen}{Nielsen}{1993}]%
        {nielsen1993usability}
\bibfield{author}{\bibinfo{person}{Jakob Nielsen}.}
  \bibinfo{year}{1993}\natexlab{}.
\newblock \showarticletitle{{Usability Heuristics}}.
\newblock In \bibinfo{booktitle}{\emph{Usability Engineering}},
  \bibfield{editor}{\bibinfo{person}{JAKOB NIELSEN}} (Ed.).
  \bibinfo{publisher}{Morgan Kaufmann}, \bibinfo{address}{San Diego},
  \bibinfo{pages}{115--163}.
\newblock
\showISBNx{978-0-12-518406-9}
\showISSN{1548-5552}
\urldef\tempurl%
\url{https://doi.org/10.1016/B978-0-08-052029-2.50008-5}
\showDOI{\tempurl}


\bibitem[\protect\citeauthoryear{Qi, Sparks, and Talwalkar}{Qi
  et~al\mbox{.}}{2016}]%
        {qi2016paleo}
\bibfield{author}{\bibinfo{person}{Hang Qi}, \bibinfo{person}{Evan~R Sparks},
  {and} \bibinfo{person}{Ameet Talwalkar}.} \bibinfo{year}{2016}\natexlab{}.
\newblock \showarticletitle{Paleo: A performance model for deep neural
  networks}.
\newblock  (\bibinfo{year}{2016}).
\newblock


\bibitem[\protect\citeauthoryear{Radosavovic, Kosaraju, Girshick, He, and
  Doll{\'a}r}{Radosavovic et~al\mbox{.}}{2020}]%
        {radosavovic2020designing}
\bibfield{author}{\bibinfo{person}{Ilija Radosavovic},
  \bibinfo{person}{Raj~Prateek Kosaraju}, \bibinfo{person}{Ross Girshick},
  \bibinfo{person}{Kaiming He}, {and} \bibinfo{person}{Piotr Doll{\'a}r}.}
  \bibinfo{year}{2020}\natexlab{}.
\newblock \showarticletitle{Designing Network Design Spaces}.
\newblock \bibinfo{journal}{\emph{arXiv preprint arXiv:2003.13678}}
  (\bibinfo{year}{2020}).
\newblock


\bibitem[\protect\citeauthoryear{Rastegari, Ordonez, Redmon, and
  Farhadi}{Rastegari et~al\mbox{.}}{2016}]%
        {rastegari2016xnor}
\bibfield{author}{\bibinfo{person}{Mohammad Rastegari},
  \bibinfo{person}{Vicente Ordonez}, \bibinfo{person}{Joseph Redmon}, {and}
  \bibinfo{person}{Ali Farhadi}.} \bibinfo{year}{2016}\natexlab{}.
\newblock \showarticletitle{Xnor-net: Imagenet classification using binary
  convolutional neural networks}. In \bibinfo{booktitle}{\emph{European
  conference on computer vision}}. Springer, \bibinfo{pages}{525--542}.
\newblock


\bibitem[\protect\citeauthoryear{Ravi}{Ravi}{2017}]%
        {ravi2017projectionnet}
\bibfield{author}{\bibinfo{person}{Sujith Ravi}.}
  \bibinfo{year}{2017}\natexlab{}.
\newblock \showarticletitle{Projectionnet: Learning efficient on-device deep
  networks using neural projections}.
\newblock \bibinfo{journal}{\emph{arXiv preprint arXiv:1708.00630}}
  (\bibinfo{year}{2017}).
\newblock


\bibitem[\protect\citeauthoryear{Raz}{Raz}{2017}]%
        {raz2017time}
\bibfield{author}{\bibinfo{person}{Ran Raz}.} \bibinfo{year}{2017}\natexlab{}.
\newblock \showarticletitle{A time-space lower bound for a large class of
  learning problems}. In \bibinfo{booktitle}{\emph{Foundations of Computer
  Science (FOCS), 2017 IEEE 58th Annual Symposium on}}. IEEE,
  \bibinfo{pages}{732--742}.
\newblock


\bibitem[\protect\citeauthoryear{Research}{Research}{[n.d.]}]%
        {deepbench}
\bibfield{author}{\bibinfo{person}{Baidu Research}.}
  \bibinfo{year}{[n.d.]}\natexlab{}.
\newblock \bibinfo{booktitle}{\emph{Benchmarking Deep Learning Operations on
  Different Hardwares}}.
\newblock
\urldef\tempurl%
\url{https://github.com/baidu-research/DeepBench}
\showURL{%
\tempurl}


\bibitem[\protect\citeauthoryear{Rhu, Gimelshein, Clemons, Zulfiqar, and
  Keckler}{Rhu et~al\mbox{.}}{2016}]%
        {rhu2016vdnn}
\bibfield{author}{\bibinfo{person}{Minsoo Rhu}, \bibinfo{person}{Natalia
  Gimelshein}, \bibinfo{person}{Jason Clemons}, \bibinfo{person}{Arslan
  Zulfiqar}, {and} \bibinfo{person}{Stephen~W Keckler}.}
  \bibinfo{year}{2016}\natexlab{}.
\newblock \showarticletitle{vDNN: Virtualized deep neural networks for
  scalable, memory-efficient neural network design}. In
  \bibinfo{booktitle}{\emph{The 49th Annual IEEE/ACM International Symposium on
  Microarchitecture}}. IEEE Press, \bibinfo{pages}{18}.
\newblock


\bibitem[\protect\citeauthoryear{Rodrigues, Riley, and Luj{\'a}n}{Rodrigues
  et~al\mbox{.}}{[n.d.]}]%
        {rodriguesfine}
\bibfield{author}{\bibinfo{person}{Crefeda~Faviola Rodrigues},
  \bibinfo{person}{Graham Riley}, {and} \bibinfo{person}{Mikel Luj{\'a}n}.}
  \bibinfo{year}{[n.d.]}\natexlab{}.
\newblock \showarticletitle{Fine-Grained Energy and Performance Profiling
  framework for Deep Convolutional Neural Networks}.
\newblock  (\bibinfo{year}{[n.\,d.]}).
\newblock


\bibitem[\protect\citeauthoryear{Rouhani, Mirhoseini, and Koushanfar}{Rouhani
  et~al\mbox{.}}{2016}]%
        {rouhani2016delight}
\bibfield{author}{\bibinfo{person}{Bita~Darvish Rouhani},
  \bibinfo{person}{Azalia Mirhoseini}, {and} \bibinfo{person}{Farinaz
  Koushanfar}.} \bibinfo{year}{2016}\natexlab{}.
\newblock \showarticletitle{Delight: Adding energy dimension to deep neural
  networks}. In \bibinfo{booktitle}{\emph{Proceedings of the 2016 International
  Symposium on Low Power Electronics and Design}}. ACM,
  \bibinfo{pages}{112--117}.
\newblock


\bibitem[\protect\citeauthoryear{Rouhani, Mirhoseini, and Koushanfar}{Rouhani
  et~al\mbox{.}}{2017}]%
        {rouhani2017tinydl}
\bibfield{author}{\bibinfo{person}{Bita~Darvish Rouhani},
  \bibinfo{person}{Azalia Mirhoseini}, {and} \bibinfo{person}{Farinaz
  Koushanfar}.} \bibinfo{year}{2017}\natexlab{}.
\newblock \showarticletitle{TinyDL: Just-in-time deep learning solution for
  constrained embedded systems}. In \bibinfo{booktitle}{\emph{IEEE
  International Symposium on Circuits and Systems (ISCAS)}}.
  \bibinfo{pages}{1--4}.
\newblock


\bibitem[\protect\citeauthoryear{Ryabko}{Ryabko}{2007}]%
        {ryabko2007sample}
\bibfield{author}{\bibinfo{person}{Daniil Ryabko}.}
  \bibinfo{year}{2007}\natexlab{}.
\newblock \showarticletitle{Sample complexity for computational classification
  problems}.
\newblock \bibinfo{journal}{\emph{Algorithmica}} \bibinfo{volume}{49},
  \bibinfo{number}{1} (\bibinfo{year}{2007}), \bibinfo{pages}{69--77}.
\newblock


\bibitem[\protect\citeauthoryear{Sandler, Howard, Zhu, Zhmoginov, and
  Chen}{Sandler et~al\mbox{.}}{2018}]%
        {sandler2018mobilenetv2}
\bibfield{author}{\bibinfo{person}{Mark Sandler}, \bibinfo{person}{Andrew
  Howard}, \bibinfo{person}{Menglong Zhu}, \bibinfo{person}{Andrey Zhmoginov},
  {and} \bibinfo{person}{Liang-Chieh Chen}.} \bibinfo{year}{2018}\natexlab{}.
\newblock \showarticletitle{MobileNetV2: Inverted Residuals and Linear
  Bottlenecks}. In \bibinfo{booktitle}{\emph{Proceedings of the IEEE Conference
  on Computer Vision and Pattern Recognition}}. \bibinfo{pages}{4510--4520}.
\newblock


\bibitem[\protect\citeauthoryear{Servedio}{Servedio}{2000}]%
        {servedio2000computational}
\bibfield{author}{\bibinfo{person}{Rocco~A Servedio}.}
  \bibinfo{year}{2000}\natexlab{}.
\newblock \showarticletitle{Computational sample complexity and
  attribute-efficient learning}.
\newblock \bibinfo{journal}{\emph{J. Comput. System Sci.}}
  \bibinfo{volume}{60}, \bibinfo{number}{1} (\bibinfo{year}{2000}),
  \bibinfo{pages}{161--178}.
\newblock


\bibitem[\protect\citeauthoryear{Shalev-Shwartz and Ben-David}{Shalev-Shwartz
  and Ben-David}{2014}]%
        {shalev2014understanding}
\bibfield{author}{\bibinfo{person}{Shai Shalev-Shwartz} {and}
  \bibinfo{person}{Shai Ben-David}.} \bibinfo{year}{2014}\natexlab{}.
\newblock \bibinfo{booktitle}{\emph{Understanding machine learning: From theory
  to algorithms}}.
\newblock \bibinfo{publisher}{Cambridge university press}.
\newblock


\bibitem[\protect\citeauthoryear{Shalev-Shwartz, Shamir, and
  Tromer}{Shalev-Shwartz et~al\mbox{.}}{2012}]%
        {shalev2012using}
\bibfield{author}{\bibinfo{person}{Shai Shalev-Shwartz}, \bibinfo{person}{Ohad
  Shamir}, {and} \bibinfo{person}{Eran Tromer}.}
  \bibinfo{year}{2012}\natexlab{}.
\newblock \showarticletitle{Using more data to speed-up training time}. In
  \bibinfo{booktitle}{\emph{Artificial Intelligence and Statistics}}.
  \bibinfo{pages}{1019--1027}.
\newblock


\bibitem[\protect\citeauthoryear{Shamir}{Shamir}{2014}]%
        {shamir2014fundamental}
\bibfield{author}{\bibinfo{person}{Ohad Shamir}.}
  \bibinfo{year}{2014}\natexlab{}.
\newblock \showarticletitle{Fundamental limits of online and distributed
  algorithms for statistical learning and estimation}. In
  \bibinfo{booktitle}{\emph{Advances in Neural Information Processing
  Systems}}. \bibinfo{pages}{163--171}.
\newblock


\bibitem[\protect\citeauthoryear{Simonyan and Zisserman}{Simonyan and
  Zisserman}{2014}]%
        {simonyan2014very}
\bibfield{author}{\bibinfo{person}{Karen Simonyan} {and}
  \bibinfo{person}{Andrew Zisserman}.} \bibinfo{year}{2014}\natexlab{}.
\newblock \showarticletitle{Very deep convolutional networks for large-scale
  image recognition}.
\newblock \bibinfo{journal}{\emph{arXiv preprint arXiv:1409.1556}}
  (\bibinfo{year}{2014}).
\newblock


\bibitem[\protect\citeauthoryear{Soudry, Hubara, and Meir}{Soudry
  et~al\mbox{.}}{2014}]%
        {soudry2014expectation}
\bibfield{author}{\bibinfo{person}{Daniel Soudry}, \bibinfo{person}{Itay
  Hubara}, {and} \bibinfo{person}{Ron Meir}.} \bibinfo{year}{2014}\natexlab{}.
\newblock \showarticletitle{Expectation backpropagation: Parameter-free
  training of multilayer neural networks with continuous or discrete weights}.
  In \bibinfo{booktitle}{\emph{Advances in Neural Information Processing
  Systems}}. \bibinfo{pages}{963--971}.
\newblock


\bibitem[\protect\citeauthoryear{Srebro and Sridharan}{Srebro and
  Sridharan}{2011}]%
        {srebro2011theoretical}
\bibfield{author}{\bibinfo{person}{Nathan Srebro} {and}
  \bibinfo{person}{Karthik Sridharan}.} \bibinfo{year}{2011}\natexlab{}.
\newblock \showarticletitle{Theoretical basis for “more data less work”}.
  In \bibinfo{booktitle}{\emph{NIPS Workshop on Computataional Trade-offs in
  Statistical Learning}}.
\newblock


\bibitem[\protect\citeauthoryear{Stamoulis, Cai, Juan, and
  Marculescu}{Stamoulis et~al\mbox{.}}{2018}]%
        {stamoulis2018hyperpower}
\bibfield{author}{\bibinfo{person}{Dimitrios Stamoulis}, \bibinfo{person}{Ermao
  Cai}, \bibinfo{person}{Da-Cheng Juan}, {and} \bibinfo{person}{Diana
  Marculescu}.} \bibinfo{year}{2018}\natexlab{}.
\newblock \showarticletitle{HyperPower: Power-and memory-constrained
  hyper-parameter optimization for neural networks}. In
  \bibinfo{booktitle}{\emph{Design, Automation \& Test in Europe Conference \&
  Exhibition (DATE), 2018}}. IEEE, \bibinfo{pages}{19--24}.
\newblock


\bibitem[\protect\citeauthoryear{Stamoulis, Ding, Wang, Lymberopoulos,
  Priyantha, Liu, and Marculescu}{Stamoulis et~al\mbox{.}}{2019}]%
        {stamoulis2019single}
\bibfield{author}{\bibinfo{person}{Dimitrios Stamoulis},
  \bibinfo{person}{Ruizhou Ding}, \bibinfo{person}{Di Wang},
  \bibinfo{person}{Dimitrios Lymberopoulos}, \bibinfo{person}{Bodhi Priyantha},
  \bibinfo{person}{Jie Liu}, {and} \bibinfo{person}{Diana Marculescu}.}
  \bibinfo{year}{2019}\natexlab{}.
\newblock \showarticletitle{Single-path nas: Designing hardware-efficient
  convnets in less than 4 hours}.
\newblock \bibinfo{journal}{\emph{arXiv preprint arXiv:1904.02877}}
  (\bibinfo{year}{2019}).
\newblock


\bibitem[\protect\citeauthoryear{Steinhardt and Duchi}{Steinhardt and
  Duchi}{2015}]%
        {steinhardt2015minimax}
\bibfield{author}{\bibinfo{person}{Jacob Steinhardt} {and}
  \bibinfo{person}{John Duchi}.} \bibinfo{year}{2015}\natexlab{}.
\newblock \showarticletitle{Minimax rates for memory-bounded sparse linear
  regression}. In \bibinfo{booktitle}{\emph{Conference on Learning Theory}}.
  \bibinfo{pages}{1564--1587}.
\newblock


\bibitem[\protect\citeauthoryear{Steinhardt, Valiant, and Wager}{Steinhardt
  et~al\mbox{.}}{2016}]%
        {steinhardt2016memory}
\bibfield{author}{\bibinfo{person}{Jacob Steinhardt}, \bibinfo{person}{Gregory
  Valiant}, {and} \bibinfo{person}{Stefan Wager}.}
  \bibinfo{year}{2016}\natexlab{}.
\newblock \showarticletitle{Memory, communication, and statistical queries}. In
  \bibinfo{booktitle}{\emph{Conference on Learning Theory}}.
  \bibinfo{pages}{1490--1516}.
\newblock


\bibitem[\protect\citeauthoryear{Stich, Cordonnier, and Jaggi}{Stich
  et~al\mbox{.}}{2018}]%
        {stich2018sparsified}
\bibfield{author}{\bibinfo{person}{Sebastian~U Stich},
  \bibinfo{person}{Jean-Baptiste Cordonnier}, {and} \bibinfo{person}{Martin
  Jaggi}.} \bibinfo{year}{2018}\natexlab{}.
\newblock \showarticletitle{Sparsified SGD with memory}. In
  \bibinfo{booktitle}{\emph{Advances in Neural Information Processing
  Systems}}. \bibinfo{pages}{4447--4458}.
\newblock


\bibitem[\protect\citeauthoryear{Strom}{Strom}{2015}]%
        {strom2015scalable}
\bibfield{author}{\bibinfo{person}{Nikko Strom}.}
  \bibinfo{year}{2015}\natexlab{}.
\newblock \showarticletitle{Scalable distributed DNN training using commodity
  GPU cloud computing}. In \bibinfo{booktitle}{\emph{Sixteenth Annual
  Conference of the International Speech Communication Association}}.
\newblock


\bibitem[\protect\citeauthoryear{Sze, Chen, Yang, and Emer}{Sze
  et~al\mbox{.}}{2017}]%
        {sze2017efficient}
\bibfield{author}{\bibinfo{person}{Vivienne Sze}, \bibinfo{person}{Yu-Hsin
  Chen}, \bibinfo{person}{Tien-Ju Yang}, {and} \bibinfo{person}{Joel~S Emer}.}
  \bibinfo{year}{2017}\natexlab{}.
\newblock \showarticletitle{Efficient processing of deep neural networks: A
  tutorial and survey}.
\newblock \bibinfo{journal}{\emph{Proc. IEEE}} \bibinfo{volume}{105},
  \bibinfo{number}{12} (\bibinfo{year}{2017}), \bibinfo{pages}{2295--2329}.
\newblock


\bibitem[\protect\citeauthoryear{Szegedy, Liu, Jia, Sermanet, Reed, Anguelov,
  Erhan, Vanhoucke, and Rabinovich}{Szegedy et~al\mbox{.}}{2015}]%
        {szegedy2015going}
\bibfield{author}{\bibinfo{person}{Christian Szegedy}, \bibinfo{person}{Wei
  Liu}, \bibinfo{person}{Yangqing Jia}, \bibinfo{person}{Pierre Sermanet},
  \bibinfo{person}{Scott Reed}, \bibinfo{person}{Dragomir Anguelov},
  \bibinfo{person}{Dumitru Erhan}, \bibinfo{person}{Vincent Vanhoucke}, {and}
  \bibinfo{person}{Andrew Rabinovich}.} \bibinfo{year}{2015}\natexlab{}.
\newblock \showarticletitle{Going deeper with convolutions}. In
  \bibinfo{booktitle}{\emph{Proceedings of the IEEE conference on computer
  vision and pattern recognition}}. \bibinfo{pages}{1--9}.
\newblock


\bibitem[\protect\citeauthoryear{Szegedy, Vanhoucke, Ioffe, Shlens, and
  Wojna}{Szegedy et~al\mbox{.}}{2016}]%
        {szegedy2016rethinking}
\bibfield{author}{\bibinfo{person}{Christian Szegedy}, \bibinfo{person}{Vincent
  Vanhoucke}, \bibinfo{person}{Sergey Ioffe}, \bibinfo{person}{Jon Shlens},
  {and} \bibinfo{person}{Zbigniew Wojna}.} \bibinfo{year}{2016}\natexlab{}.
\newblock \showarticletitle{Rethinking the inception architecture for computer
  vision}. In \bibinfo{booktitle}{\emph{Proceedings of the IEEE conference on
  computer vision and pattern recognition}}. \bibinfo{pages}{2818--2826}.
\newblock


\bibitem[\protect\citeauthoryear{Tai, Sharan, Bailis, and Valiant}{Tai
  et~al\mbox{.}}{2018}]%
        {tai2018sketching}
\bibfield{author}{\bibinfo{person}{Kai~Sheng Tai}, \bibinfo{person}{Vatsal
  Sharan}, \bibinfo{person}{Peter Bailis}, {and} \bibinfo{person}{Gregory
  Valiant}.} \bibinfo{year}{2018}\natexlab{}.
\newblock \showarticletitle{Sketching Linear Classifiers over Data Streams}. In
  \bibinfo{booktitle}{\emph{Proceedings of the 2018 International Conference on
  Management of Data}}. ACM, \bibinfo{pages}{757--772}.
\newblock


\bibitem[\protect\citeauthoryear{Tan, Chen, Pang, Vasudevan, and Le}{Tan
  et~al\mbox{.}}{2018}]%
        {tan2018mnasnet}
\bibfield{author}{\bibinfo{person}{Mingxing Tan}, \bibinfo{person}{Bo Chen},
  \bibinfo{person}{Ruoming Pang}, \bibinfo{person}{Vijay Vasudevan}, {and}
  \bibinfo{person}{Quoc~V Le}.} \bibinfo{year}{2018}\natexlab{}.
\newblock \showarticletitle{MnasNet: Platform-Aware Neural Architecture Search
  for Mobile}.
\newblock \bibinfo{journal}{\emph{arXiv preprint arXiv:1807.11626}}
  (\bibinfo{year}{2018}).
\newblock


\bibitem[\protect\citeauthoryear{Tan and Le}{Tan and Le}{2019a}]%
        {tan2019efficientnet}
\bibfield{author}{\bibinfo{person}{Mingxing Tan} {and} \bibinfo{person}{Quoc~V
  Le}.} \bibinfo{year}{2019}\natexlab{a}.
\newblock \showarticletitle{Efficientnet: Rethinking model scaling for
  convolutional neural networks}.
\newblock \bibinfo{journal}{\emph{arXiv preprint arXiv:1905.11946}}
  (\bibinfo{year}{2019}).
\newblock


\bibitem[\protect\citeauthoryear{Tan and Le}{Tan and Le}{2019b}]%
        {tan2019mixconv}
\bibfield{author}{\bibinfo{person}{Mingxing Tan} {and} \bibinfo{person}{Quoc~V
  Le}.} \bibinfo{year}{2019}\natexlab{b}.
\newblock \showarticletitle{Mixconv: Mixed depthwise convolutional kernels}. In
  \bibinfo{booktitle}{\emph{British Machine Vision Conference}}.
\newblock


\bibitem[\protect\citeauthoryear{Tang, Gan, Zhang, Zhang, and Liu}{Tang
  et~al\mbox{.}}{2018}]%
        {tang2018communication}
\bibfield{author}{\bibinfo{person}{Hanlin Tang}, \bibinfo{person}{Shaoduo Gan},
  \bibinfo{person}{Ce Zhang}, \bibinfo{person}{Tong Zhang}, {and}
  \bibinfo{person}{Ji Liu}.} \bibinfo{year}{2018}\natexlab{}.
\newblock \showarticletitle{Communication compression for decentralized
  training}. In \bibinfo{booktitle}{\emph{Advances in Neural Information
  Processing Systems}}. \bibinfo{pages}{7652--7662}.
\newblock


\bibitem[\protect\citeauthoryear{Tang, Shi, Chu, Wang, and Li}{Tang
  et~al\mbox{.}}{2020}]%
        {tang2020communicationefficient}
\bibfield{author}{\bibinfo{person}{Zhenheng Tang}, \bibinfo{person}{Shaohuai
  Shi}, \bibinfo{person}{Xiaowen Chu}, \bibinfo{person}{Wei Wang}, {and}
  \bibinfo{person}{Bo Li}.} \bibinfo{year}{2020}\natexlab{}.
\newblock \bibinfo{title}{Communication-Efficient Distributed Deep Learning: A
  Comprehensive Survey}.
\newblock
\newblock
\showeprint[arxiv]{cs.DC/2003.06307}


\bibitem[\protect\citeauthoryear{Tsianos, Lawlor, and Rabbat}{Tsianos
  et~al\mbox{.}}{2012}]%
        {tsianos2012consensus}
\bibfield{author}{\bibinfo{person}{Konstantinos~I Tsianos},
  \bibinfo{person}{Sean Lawlor}, {and} \bibinfo{person}{Michael~G Rabbat}.}
  \bibinfo{year}{2012}\natexlab{}.
\newblock \showarticletitle{Consensus-based distributed optimization: Practical
  issues and applications in large-scale machine learning}. In
  \bibinfo{booktitle}{\emph{50th Annual Allerton Conference on Communication,
  Control, and Computing (Allerton)}}. IEEE, \bibinfo{pages}{1543--1550}.
\newblock


\bibitem[\protect\citeauthoryear{Valiant}{Valiant}{1984}]%
        {valiant1984theory}
\bibfield{author}{\bibinfo{person}{Leslie~G Valiant}.}
  \bibinfo{year}{1984}\natexlab{}.
\newblock \showarticletitle{A theory of the learnable}.
\newblock \bibinfo{journal}{\emph{Commun. ACM}} \bibinfo{volume}{27},
  \bibinfo{number}{11} (\bibinfo{year}{1984}), \bibinfo{pages}{1134--1142}.
\newblock


\bibitem[\protect\citeauthoryear{van~den Berg, Ramabhadran, and
  Picheny}{van~den Berg et~al\mbox{.}}{2017}]%
        {van2017training}
\bibfield{author}{\bibinfo{person}{Ewout van~den Berg},
  \bibinfo{person}{Bhuvana Ramabhadran}, {and} \bibinfo{person}{Michael
  Picheny}.} \bibinfo{year}{2017}\natexlab{}.
\newblock \showarticletitle{Training variance and performance evaluation of
  neural networks in speech}. In \bibinfo{booktitle}{\emph{2017 IEEE
  International Conference on Acoustics, Speech and Signal Processing
  (ICASSP)}}. IEEE, \bibinfo{pages}{2287--2291}.
\newblock


\bibitem[\protect\citeauthoryear{Van~Engelen and Hoos}{Van~Engelen and
  Hoos}{2020}]%
        {van2020survey}
\bibfield{author}{\bibinfo{person}{Jesper~E Van~Engelen} {and}
  \bibinfo{person}{Holger~H Hoos}.} \bibinfo{year}{2020}\natexlab{}.
\newblock \showarticletitle{A survey on semi-supervised learning}.
\newblock \bibinfo{journal}{\emph{Machine Learning}} \bibinfo{volume}{109},
  \bibinfo{number}{2} (\bibinfo{year}{2020}), \bibinfo{pages}{373--440}.
\newblock


\bibitem[\protect\citeauthoryear{Vapnik}{Vapnik}{2006}]%
        {vapnik2006estimation}
\bibfield{author}{\bibinfo{person}{Vladimir Vapnik}.}
  \bibinfo{year}{2006}\natexlab{}.
\newblock \bibinfo{booktitle}{\emph{Estimation of dependences based on
  empirical data}}.
\newblock \bibinfo{publisher}{Springer Science \& Business Media}.
\newblock


\bibitem[\protect\citeauthoryear{Vapnik and Izmailov}{Vapnik and
  Izmailov}{2019}]%
        {vapnik2019rethinking}
\bibfield{author}{\bibinfo{person}{Vladimir Vapnik} {and} \bibinfo{person}{Rauf
  Izmailov}.} \bibinfo{year}{2019}\natexlab{}.
\newblock \showarticletitle{Rethinking statistical learning theory: learning
  using statistical invariants}.
\newblock \bibinfo{journal}{\emph{Machine Learning}} \bibinfo{volume}{108},
  \bibinfo{number}{3} (\bibinfo{year}{2019}), \bibinfo{pages}{381--423}.
\newblock


\bibitem[\protect\citeauthoryear{Vapnik and Chervonenkis}{Vapnik and
  Chervonenkis}{1971}]%
        {vapnik71uniform}
\bibfield{author}{\bibinfo{person}{V.~N. Vapnik} {and} \bibinfo{person}{A.~Ya.
  Chervonenkis}.} \bibinfo{year}{1971}\natexlab{}.
\newblock \showarticletitle{On the Uniform Convergence of Relative Frequencies
  of Events to their Probabilities}.
\newblock \bibinfo{journal}{\emph{Theory of Probability and its Applications}}
  \bibinfo{volume}{16}, \bibinfo{number}{2} (\bibinfo{year}{1971}),
  \bibinfo{pages}{264--280}.
\newblock


\bibitem[\protect\citeauthoryear{Velasco-Montero, Fern{\'{a}}ndez-Berni,
  Carmona-Gal{\'{a}}n, and Rodr{\'{i}}guez-V{\'{a}}zquez}{Velasco-Montero
  et~al\mbox{.}}{2018}]%
        {velasco2018performance}
\bibfield{author}{\bibinfo{person}{Delia Velasco-Montero},
  \bibinfo{person}{Jorge Fern{\'{a}}ndez-Berni}, \bibinfo{person}{Ricardo
  Carmona-Gal{\'{a}}n}, {and} \bibinfo{person}{Angel
  Rodr{\'{i}}guez-V{\'{a}}zquez}.} \bibinfo{year}{2018}\natexlab{}.
\newblock \showarticletitle{{Performance Analysis of Real-Time DNN Inference on
  Raspberry Pi}}. In \bibinfo{booktitle}{\emph{Proc. SPIE 10670, Real-Time
  Image and Video Processing}}. \bibinfo{address}{Orlando, Florida, United
  States}, \bibinfo{pages}{10670 -- 10670 -- 9}.
\newblock


\bibitem[\protect\citeauthoryear{Venieris, Kouris, and Bouganis}{Venieris
  et~al\mbox{.}}{2018}]%
        {venieris2018deploying}
\bibfield{author}{\bibinfo{person}{Stylianos~I. Venieris},
  \bibinfo{person}{Alexandros Kouris}, {and} \bibinfo{person}{Christos-Savvas
  Bouganis}.} \bibinfo{year}{2018}\natexlab{}.
\newblock \showarticletitle{{Deploying Deep Neural Networks in the Embedded
  Space}}. In \bibinfo{booktitle}{\emph{2nd International Workshop on Embedded
  and Mobile Deep Learning}}. \bibinfo{address}{Munich, Germany}.
\newblock


\bibitem[\protect\citeauthoryear{Wan, Dai, Zhang, He, Tian, Xie, Wu, Yu, Xu,
  Chen, et~al\mbox{.}}{Wan et~al\mbox{.}}{2020}]%
        {wan2020fbnetv2}
\bibfield{author}{\bibinfo{person}{Alvin Wan}, \bibinfo{person}{Xiaoliang Dai},
  \bibinfo{person}{Peizhao Zhang}, \bibinfo{person}{Zijian He},
  \bibinfo{person}{Yuandong Tian}, \bibinfo{person}{Saining Xie},
  \bibinfo{person}{Bichen Wu}, \bibinfo{person}{Matthew Yu},
  \bibinfo{person}{Tao Xu}, \bibinfo{person}{Kan Chen}, {et~al\mbox{.}}}
  \bibinfo{year}{2020}\natexlab{}.
\newblock \showarticletitle{FBNetV2: Differentiable Neural Architecture Search
  for Spatial and Channel Dimensions}.
\newblock \bibinfo{journal}{\emph{arXiv preprint arXiv:2004.05565}}
  (\bibinfo{year}{2020}).
\newblock


\bibitem[\protect\citeauthoryear{Wang, Liu, Lin, Lin, and Han}{Wang
  et~al\mbox{.}}{2018b}]%
        {wang2018haq}
\bibfield{author}{\bibinfo{person}{Kuan Wang}, \bibinfo{person}{Zhijian Liu},
  \bibinfo{person}{Yujun Lin}, \bibinfo{person}{Ji Lin}, {and}
  \bibinfo{person}{Song Han}.} \bibinfo{year}{2018}\natexlab{b}.
\newblock \showarticletitle{HAQ: Hardware-Aware Automated Quantization}.
\newblock \bibinfo{journal}{\emph{arXiv preprint arXiv:1811.08886}}
  (\bibinfo{year}{2018}).
\newblock


\bibitem[\protect\citeauthoryear{Wang, Choi, Brand, Chen, and
  Gopalakrishnan}{Wang et~al\mbox{.}}{2018a}]%
        {wang2018training}
\bibfield{author}{\bibinfo{person}{Naigang Wang}, \bibinfo{person}{Jungwook
  Choi}, \bibinfo{person}{Daniel Brand}, \bibinfo{person}{Chia-Yu Chen}, {and}
  \bibinfo{person}{Kailash Gopalakrishnan}.} \bibinfo{year}{2018}\natexlab{a}.
\newblock \showarticletitle{Training Deep Neural Networks with 8-bit Floating
  Point Numbers}. In \bibinfo{booktitle}{\emph{Advances in Neural Information
  Processing Systems}}. \bibinfo{pages}{7685--7694}.
\newblock


\bibitem[\protect\citeauthoryear{Wang, Chen, Chen, Rai, and Carin}{Wang
  et~al\mbox{.}}{2016}]%
        {wang2016deep}
\bibfield{author}{\bibinfo{person}{Wenlin Wang}, \bibinfo{person}{Changyou
  Chen}, \bibinfo{person}{Wenlin Chen}, \bibinfo{person}{Piyush Rai}, {and}
  \bibinfo{person}{Lawrence Carin}.} \bibinfo{year}{2016}\natexlab{}.
\newblock \showarticletitle{Deep distance metric learning with data
  summarization}. In \bibinfo{booktitle}{\emph{ECML PKDD}}.
\newblock


\bibitem[\protect\citeauthoryear{Wang, Han, Leung, Niyato, Yan, and Chen}{Wang
  et~al\mbox{.}}{2020b}]%
        {wang2020convergence}
\bibfield{author}{\bibinfo{person}{Xiaofei Wang}, \bibinfo{person}{Yiwen Han},
  \bibinfo{person}{Victor~CM Leung}, \bibinfo{person}{Dusit Niyato},
  \bibinfo{person}{Xueqiang Yan}, {and} \bibinfo{person}{Xu Chen}.}
  \bibinfo{year}{2020}\natexlab{b}.
\newblock \showarticletitle{Convergence of edge computing and deep learning: A
  comprehensive survey}.
\newblock \bibinfo{journal}{\emph{IEEE Communications Surveys \& Tutorials}}
  (\bibinfo{year}{2020}).
\newblock


\bibitem[\protect\citeauthoryear{Wang, Yu, Dou, and Gonzalez}{Wang
  et~al\mbox{.}}{2017}]%
        {wang2017skipnet}
\bibfield{author}{\bibinfo{person}{Xin Wang}, \bibinfo{person}{Fisher Yu},
  \bibinfo{person}{Zi-Yi Dou}, {and} \bibinfo{person}{Joseph~E Gonzalez}.}
  \bibinfo{year}{2017}\natexlab{}.
\newblock \showarticletitle{Skipnet: Learning dynamic routing in convolutional
  networks}.
\newblock \bibinfo{journal}{\emph{arXiv preprint arXiv:1711.09485}}
  (\bibinfo{year}{2017}).
\newblock


\bibitem[\protect\citeauthoryear{Wang, Chen, and Hoi}{Wang
  et~al\mbox{.}}{2020a}]%
        {wang2020deep}
\bibfield{author}{\bibinfo{person}{Zhihao Wang}, \bibinfo{person}{Jian Chen},
  {and} \bibinfo{person}{Steven~CH Hoi}.} \bibinfo{year}{2020}\natexlab{a}.
\newblock \showarticletitle{Deep learning for image super-resolution: A
  survey}.
\newblock \bibinfo{journal}{\emph{IEEE Transactions on Pattern Analysis and
  Machine Intelligence}} (\bibinfo{year}{2020}).
\newblock


\bibitem[\protect\citeauthoryear{Wangni, Wang, Liu, and Zhang}{Wangni
  et~al\mbox{.}}{2018}]%
        {wangni2018gradient}
\bibfield{author}{\bibinfo{person}{Jianqiao Wangni}, \bibinfo{person}{Jialei
  Wang}, \bibinfo{person}{Ji Liu}, {and} \bibinfo{person}{Tong Zhang}.}
  \bibinfo{year}{2018}\natexlab{}.
\newblock \showarticletitle{Gradient sparsification for communication-efficient
  distributed optimization}. In \bibinfo{booktitle}{\emph{Advances in Neural
  Information Processing Systems}}. \bibinfo{pages}{1299--1309}.
\newblock


\bibitem[\protect\citeauthoryear{Wen, Wu, Wang, Chen, and Li}{Wen
  et~al\mbox{.}}{2016}]%
        {wen2016learning}
\bibfield{author}{\bibinfo{person}{Wei Wen}, \bibinfo{person}{Chunpeng Wu},
  \bibinfo{person}{Yandan Wang}, \bibinfo{person}{Yiran Chen}, {and}
  \bibinfo{person}{Hai Li}.} \bibinfo{year}{2016}\natexlab{}.
\newblock \showarticletitle{Learning structured sparsity in deep neural
  networks}. In \bibinfo{booktitle}{\emph{Advances in Neural Information
  Processing Systems}}. \bibinfo{pages}{2074--2082}.
\newblock


\bibitem[\protect\citeauthoryear{Wen, Xu, Yan, Wu, Wang, Chen, and Li}{Wen
  et~al\mbox{.}}{2017}]%
        {wen2017terngrad}
\bibfield{author}{\bibinfo{person}{Wei Wen}, \bibinfo{person}{Cong Xu},
  \bibinfo{person}{Feng Yan}, \bibinfo{person}{Chunpeng Wu},
  \bibinfo{person}{Yandan Wang}, \bibinfo{person}{Yiran Chen}, {and}
  \bibinfo{person}{Hai Li}.} \bibinfo{year}{2017}\natexlab{}.
\newblock \showarticletitle{Terngrad: Ternary gradients to reduce communication
  in distributed deep learning}. In \bibinfo{booktitle}{\emph{Advances in
  neural information processing systems}}. \bibinfo{pages}{1509--1519}.
\newblock


\bibitem[\protect\citeauthoryear{Wiedemann, Mehari, Kepp, and Samek}{Wiedemann
  et~al\mbox{.}}{2020}]%
        {wiedemann2020dithered}
\bibfield{author}{\bibinfo{person}{Simon Wiedemann}, \bibinfo{person}{Temesgen
  Mehari}, \bibinfo{person}{Kevin Kepp}, {and} \bibinfo{person}{Wojciech
  Samek}.} \bibinfo{year}{2020}\natexlab{}.
\newblock \showarticletitle{Dithered backprop: A sparse and quantized
  backpropagation algorithm for more efficient deep neural network training}.
\newblock \bibinfo{journal}{\emph{arXiv preprint arXiv:2004.04729}}
  (\bibinfo{year}{2020}).
\newblock


\bibitem[\protect\citeauthoryear{Wistuba, Rawat, and Pedapati}{Wistuba
  et~al\mbox{.}}{2019}]%
        {wistuba2019survey}
\bibfield{author}{\bibinfo{person}{Martin Wistuba}, \bibinfo{person}{Ambrish
  Rawat}, {and} \bibinfo{person}{Tejaswini Pedapati}.}
  \bibinfo{year}{2019}\natexlab{}.
\newblock \showarticletitle{A survey on neural architecture search}.
\newblock \bibinfo{journal}{\emph{arXiv preprint arXiv:1905.01392}}
  (\bibinfo{year}{2019}).
\newblock


\bibitem[\protect\citeauthoryear{Wolf}{Wolf}{2018}]%
        {wolf2018mathematical}
\bibfield{author}{\bibinfo{person}{Michael~M Wolf}.}
  \bibinfo{year}{2018}\natexlab{}.
\newblock \showarticletitle{Mathematical Foundations of Supervised Learning}.
\newblock  (\bibinfo{year}{2018}).
\newblock


\bibitem[\protect\citeauthoryear{Wong}{Wong}{2018}]%
        {wong2018netscore}
\bibfield{author}{\bibinfo{person}{Alexander Wong}.}
  \bibinfo{year}{2018}\natexlab{}.
\newblock \showarticletitle{NetScore: Towards Universal Metrics for Large-scale
  Performance Analysis of Deep Neural Networks for Practical On-Device Edge
  Usage}.
\newblock  (\bibinfo{year}{2018}).
\newblock


\bibitem[\protect\citeauthoryear{Wu, Dai, Zhang, Wang, Sun, Wu, Tian, Vajda,
  Jia, and Keutzer}{Wu et~al\mbox{.}}{2019}]%
        {wu2019fbnet}
\bibfield{author}{\bibinfo{person}{Bichen Wu}, \bibinfo{person}{Xiaoliang Dai},
  \bibinfo{person}{Peizhao Zhang}, \bibinfo{person}{Yanghan Wang},
  \bibinfo{person}{Fei Sun}, \bibinfo{person}{Yiming Wu},
  \bibinfo{person}{Yuandong Tian}, \bibinfo{person}{Peter Vajda},
  \bibinfo{person}{Yangqing Jia}, {and} \bibinfo{person}{Kurt Keutzer}.}
  \bibinfo{year}{2019}\natexlab{}.
\newblock \showarticletitle{Fbnet: Hardware-aware efficient convnet design via
  differentiable neural architecture search}. In
  \bibinfo{booktitle}{\emph{Proceedings of the IEEE Conference on Computer
  Vision and Pattern Recognition}}. \bibinfo{pages}{10734--10742}.
\newblock


\bibitem[\protect\citeauthoryear{Wu, Huang, Huang, and Zhang}{Wu
  et~al\mbox{.}}{2018}]%
        {wu2018error}
\bibfield{author}{\bibinfo{person}{Jiaxiang Wu}, \bibinfo{person}{Weidong
  Huang}, \bibinfo{person}{Junzhou Huang}, {and} \bibinfo{person}{Tong Zhang}.}
  \bibinfo{year}{2018}\natexlab{}.
\newblock \showarticletitle{Error compensated quantized SGD and its
  applications to large-scale distributed optimization}.
\newblock \bibinfo{journal}{\emph{arXiv preprint arXiv:1806.08054}}
  (\bibinfo{year}{2018}).
\newblock


\bibitem[\protect\citeauthoryear{Xie and Jabri}{Xie and Jabri}{1992}]%
        {xie1992analysis}
\bibfield{author}{\bibinfo{person}{Yun Xie} {and} \bibinfo{person}{Marwan~A
  Jabri}.} \bibinfo{year}{1992}\natexlab{}.
\newblock \showarticletitle{Analysis of the effects of quantization in
  multilayer neural networks using a statistical model}.
\newblock \bibinfo{journal}{\emph{IEEE Transactions on Neural Networks}}
  \bibinfo{volume}{3}, \bibinfo{number}{2} (\bibinfo{year}{1992}),
  \bibinfo{pages}{334--338}.
\newblock


\bibitem[\protect\citeauthoryear{Yang, Zhang, Kirichenko, Bai, Wilson, and
  De~Sa}{Yang et~al\mbox{.}}{2019}]%
        {yang2019swalp}
\bibfield{author}{\bibinfo{person}{Guandao Yang}, \bibinfo{person}{Tianyi
  Zhang}, \bibinfo{person}{Polina Kirichenko}, \bibinfo{person}{Junwen Bai},
  \bibinfo{person}{Andrew~Gordon Wilson}, {and} \bibinfo{person}{Christopher
  De~Sa}.} \bibinfo{year}{2019}\natexlab{}.
\newblock \showarticletitle{Swalp: Stochastic weight averaging in low-precision
  training}.
\newblock \bibinfo{journal}{\emph{arXiv preprint arXiv:1904.11943}}
  (\bibinfo{year}{2019}).
\newblock


\bibitem[\protect\citeauthoryear{Yang, Chen, and Sze}{Yang
  et~al\mbox{.}}{2017}]%
        {yang2017designing}
\bibfield{author}{\bibinfo{person}{Tien-Ju Yang}, \bibinfo{person}{Yu-Hsin
  Chen}, {and} \bibinfo{person}{Vivienne Sze}.}
  \bibinfo{year}{2017}\natexlab{}.
\newblock \showarticletitle{Designing Energy-Efficient Convolutional Neural
  Networks Using Energy-Aware Pruning}. In \bibinfo{booktitle}{\emph{2017 IEEE
  Conference on Computer Vision and Pattern Recognition (CVPR)}}. IEEE,
  \bibinfo{pages}{6071--6079}.
\newblock


\bibitem[\protect\citeauthoryear{Yang and Xu}{Yang and Xu}{2015}]%
        {yang2015streaming}
\bibfield{author}{\bibinfo{person}{Wenzhuo Yang} {and} \bibinfo{person}{Huan
  Xu}.} \bibinfo{year}{2015}\natexlab{}.
\newblock \showarticletitle{Streaming sparse principal component analysis}. In
  \bibinfo{booktitle}{\emph{International Conference on Machine Learning}}.
  \bibinfo{pages}{494--503}.
\newblock


\bibitem[\protect\citeauthoryear{Yang, Deng, Wu, Yan, Xie, and Li}{Yang
  et~al\mbox{.}}{2020}]%
        {yang2020training}
\bibfield{author}{\bibinfo{person}{Yukuan Yang}, \bibinfo{person}{Lei Deng},
  \bibinfo{person}{Shuang Wu}, \bibinfo{person}{Tianyi Yan},
  \bibinfo{person}{Yuan Xie}, {and} \bibinfo{person}{Guoqi Li}.}
  \bibinfo{year}{2020}\natexlab{}.
\newblock \showarticletitle{Training high-performance and large-scale deep
  neural networks with full 8-bit integers}.
\newblock \bibinfo{journal}{\emph{Neural Networks}} (\bibinfo{year}{2020}).
\newblock


\bibitem[\protect\citeauthoryear{Yao, Xiao, Wang, Viswanath, Zheng, and
  Zhao}{Yao et~al\mbox{.}}{2017}]%
        {yao2017complexity}
\bibfield{author}{\bibinfo{person}{Yuanshun Yao}, \bibinfo{person}{Zhujun
  Xiao}, \bibinfo{person}{Bolun Wang}, \bibinfo{person}{Bimal Viswanath},
  \bibinfo{person}{Haitao Zheng}, {and} \bibinfo{person}{Ben~Y Zhao}.}
  \bibinfo{year}{2017}\natexlab{}.
\newblock \showarticletitle{Complexity vs. performance: empirical analysis of
  machine learning as a service}. In \bibinfo{booktitle}{\emph{Proceedings of
  the 2017 Internet Measurement Conference}}. ACM, \bibinfo{pages}{384--397}.
\newblock


\bibitem[\protect\citeauthoryear{Zen, Agiomyrgiannakis, Egberts, Henderson, and
  Szczepaniak}{Zen et~al\mbox{.}}{2016}]%
        {zen2016fast}
\bibfield{author}{\bibinfo{person}{Heiga Zen}, \bibinfo{person}{Yannis
  Agiomyrgiannakis}, \bibinfo{person}{Niels Egberts}, \bibinfo{person}{Fergus
  Henderson}, {and} \bibinfo{person}{Przemys{\l}aw Szczepaniak}.}
  \bibinfo{year}{2016}\natexlab{}.
\newblock \showarticletitle{Fast, compact, and high quality LSTM-RNN based
  statistical parametric speech synthesizers for mobile devices}.
\newblock \bibinfo{journal}{\emph{arXiv preprint arXiv:1606.06061}}
  (\bibinfo{year}{2016}).
\newblock


\bibitem[\protect\citeauthoryear{Zhang, Liu, Zhang, and Almpanidis}{Zhang
  et~al\mbox{.}}{2017a}]%
        {zhang2017up}
\bibfield{author}{\bibinfo{person}{Chongsheng Zhang},
  \bibinfo{person}{Changchang Liu}, \bibinfo{person}{Xiangliang Zhang}, {and}
  \bibinfo{person}{George Almpanidis}.} \bibinfo{year}{2017}\natexlab{a}.
\newblock \showarticletitle{An up-to-date comparison of state-of-the-art
  classification algorithms}.
\newblock \bibinfo{journal}{\emph{Expert Systems with Applications}}
  \bibinfo{volume}{82} (\bibinfo{year}{2017}), \bibinfo{pages}{128--150}.
\newblock


\bibitem[\protect\citeauthoryear{Zhang, Li, Kara, Alistarh, Liu, and
  Zhang}{Zhang et~al\mbox{.}}{2016}]%
        {zhang2016zipml}
\bibfield{author}{\bibinfo{person}{Hantian Zhang}, \bibinfo{person}{Jerry Li},
  \bibinfo{person}{Kaan Kara}, \bibinfo{person}{Dan Alistarh},
  \bibinfo{person}{Ji Liu}, {and} \bibinfo{person}{Ce Zhang}.}
  \bibinfo{year}{2016}\natexlab{}.
\newblock \showarticletitle{The zipml framework for training models with
  end-to-end low precision: The cans, the cannots, and a little bit of deep
  learning}.
\newblock \bibinfo{journal}{\emph{arXiv preprint arXiv:1611.05402}}
  (\bibinfo{year}{2016}).
\newblock


\bibitem[\protect\citeauthoryear{Zhang, Liu, Zhang, Xiong, Xing, and Ye}{Zhang
  et~al\mbox{.}}{2017b}]%
        {zhang2017randomization}
\bibfield{author}{\bibinfo{person}{Kai Zhang}, \bibinfo{person}{Chuanren Liu},
  \bibinfo{person}{Jie Zhang}, \bibinfo{person}{Hui Xiong},
  \bibinfo{person}{Eric Xing}, {and} \bibinfo{person}{Jieping Ye}.}
  \bibinfo{year}{2017}\natexlab{b}.
\newblock \showarticletitle{Randomization or Condensation?: Linear-Cost Matrix
  Sketching Via Cascaded Compression Sampling}. In
  \bibinfo{booktitle}{\emph{Proceedings of the 23rd ACM SIGKDD International
  Conference on Knowledge Discovery and Data Mining}}. ACM,
  \bibinfo{pages}{615--623}.
\newblock


\bibitem[\protect\citeauthoryear{Zhang, Zhou, Lin, and Sun}{Zhang
  et~al\mbox{.}}{[n.d.]}]%
        {zhang1707shufflenet}
\bibfield{author}{\bibinfo{person}{X Zhang}, \bibinfo{person}{X Zhou},
  \bibinfo{person}{M Lin}, {and} \bibinfo{person}{J Sun}.}
  \bibinfo{year}{[n.d.]}\natexlab{}.
\newblock \showarticletitle{ShuffleNet: An Extremely Efficient Convolutional
  Neural Network for Mobile Devices. arXiv 2017}.
\newblock \bibinfo{journal}{\emph{arXiv preprint arXiv:1707.01083}}
  (\bibinfo{year}{[n.\,d.]}).
\newblock


\bibitem[\protect\citeauthoryear{Zhou, Wu, Ni, Zhou, Wen, and Zou}{Zhou
  et~al\mbox{.}}{2016}]%
        {zhou2016dorefa}
\bibfield{author}{\bibinfo{person}{Shuchang Zhou}, \bibinfo{person}{Yuxin Wu},
  \bibinfo{person}{Zekun Ni}, \bibinfo{person}{Xinyu Zhou}, \bibinfo{person}{He
  Wen}, {and} \bibinfo{person}{Yuheng Zou}.} \bibinfo{year}{2016}\natexlab{}.
\newblock \showarticletitle{Dorefa-net: Training low bitwidth convolutional
  neural networks with low bitwidth gradients}.
\newblock \bibinfo{journal}{\emph{arXiv preprint arXiv:1606.06160}}
  (\bibinfo{year}{2016}).
\newblock


\bibitem[\protect\citeauthoryear{Zhou, Chen, Li, Zeng, Luo, and Zhang}{Zhou
  et~al\mbox{.}}{2019}]%
        {zhou2019edge}
\bibfield{author}{\bibinfo{person}{Zhi Zhou}, \bibinfo{person}{Xu Chen},
  \bibinfo{person}{En Li}, \bibinfo{person}{Liekang Zeng}, \bibinfo{person}{Ke
  Luo}, {and} \bibinfo{person}{Junshan Zhang}.}
  \bibinfo{year}{2019}\natexlab{}.
\newblock \showarticletitle{Edge intelligence: Paving the last mile of
  artificial intelligence with edge computing}.
\newblock \bibinfo{journal}{\emph{Proc. IEEE}} \bibinfo{volume}{107},
  \bibinfo{number}{8} (\bibinfo{year}{2019}), \bibinfo{pages}{1738--1762}.
\newblock


\bibitem[\protect\citeauthoryear{Zhu, Akrout, Zheng, Pelegris, Phanishayee,
  Schroeder, and Pekhimenko}{Zhu et~al\mbox{.}}{2018}]%
        {zhu2018tbd}
\bibfield{author}{\bibinfo{person}{Hongyu Zhu}, \bibinfo{person}{Mohamed
  Akrout}, \bibinfo{person}{Bojian Zheng}, \bibinfo{person}{Andrew Pelegris},
  \bibinfo{person}{Amar Phanishayee}, \bibinfo{person}{Bianca Schroeder}, {and}
  \bibinfo{person}{Gennady Pekhimenko}.} \bibinfo{year}{2018}\natexlab{}.
\newblock \showarticletitle{TBD: Benchmarking and Analyzing Deep Neural Network
  Training}.
\newblock \bibinfo{journal}{\emph{arXiv preprint arXiv:1803.06905}}
  (\bibinfo{year}{2018}).
\newblock


\bibitem[\protect\citeauthoryear{Zhu}{Zhu}{2005}]%
        {zhu2005semi}
\bibfield{author}{\bibinfo{person}{Xiaojin~Jerry Zhu}.}
  \bibinfo{year}{2005}\natexlab{}.
\newblock \bibinfo{booktitle}{\emph{Semi-supervised learning literature
  survey}}.
\newblock \bibinfo{type}{{T}echnical {R}eport}.
  \bibinfo{institution}{University of Wisconsin-Madison Department of Computer
  Sciences}.
\newblock


\end{thebibliography}
\setcitestyle{numbers,sort&compress}

\end{document}